\documentclass{article}

\usepackage[letterpaper,margin=0.95in]{geometry}
\usepackage[authoryear,round]{natbib}

\usepackage{amsmath,amssymb,amsthm,mathtools}
\usepackage{bm}
\usepackage{mathrsfs}
\usepackage{enumitem}
\usepackage{graphicx}
\usepackage{wrapfig}
\usepackage{xcolor}
\usepackage{hyperref}
\hypersetup{
  colorlinks=true,
  linkcolor=blue,
  citecolor=blue,
  urlcolor=blue,
}

\newtheorem{theorem}{Theorem}

\newtheorem{corollary}[theorem]{Corollary}
\newtheorem{lemma}[theorem]{Lemma}

\theoremstyle{definition}
\newtheorem{definition}[theorem]{Definition}
\newtheorem{assumption}[theorem]{Assumption}
\theoremstyle{remark}
\newtheorem{remark}[theorem]{Remark}

\newcommand{\E}{\mathbb{E}}
\newcommand{\R}{\mathbb{R}}
\newcommand{\N}{\mathbb{N}}

\newcommand{\supp}{\mathrm{supp}}

\title{
 ``Deep lifts'': Neural networks can generalize even with the dumbest optimizer
}

\author{
  Sunil Madhow\\
  \small \textit{UC San Diego}
  \and
  Yu-Xiang Wang\\
  \small \textit{UC San Diego}
}

\date{}

\begin{document}
\maketitle
\begin{abstract}
Given a labeled dataset $\{(x_i, y_i)\}_{i = 1}^n$, consider drawing a parameter vector $\hat{\theta}$ uniformly at random from those that interpolate the data within a width-$W$ depth-$D$ ReLU architecture with continuous weights bounded within $[-M, M]$. We study the generalization properties of $\hat{\theta}$ in the case where the dataset can, in fact, be explained by a smaller $w$-by-$d$ ReLU network. Despite the obliviousness of the selection rule for $\hat{\theta}$, it adaptively generalizes at the rate $\tilde O\left(\frac{wdWD}{n}\right)$, rather than being governed by the VC dimension of the full $W$-by-$D$ hypothesis class. The reason is due to a particular inductive bias in ReLU networks: the $w$-by-$d$ parameter \textbf{witnesses} an exponential volume of $W$-by-$D$ interpolating parameters, which dwarfs the volume of poorly generalizing parameters (conditioned on interpolation). We explicitly construct this set of near-equivalent $W$-by-$D$ parameters by a proof technique called \textbf{deep lifting}, and then tie the achievable volume to stability properties of the $w$-by-$d$ witness.  Examples show that these techniques allow us to convert approximation-theoretic results into generalization guarantees for nonparametric classes.  We further show that generalization of uniform-random interpolators can survive label corruptions. 
\end{abstract}

\section{Introduction}

\begin{wrapfigure}{r}{0.35\textwidth}
  \centering
  \includegraphics[width=0.35\textwidth]{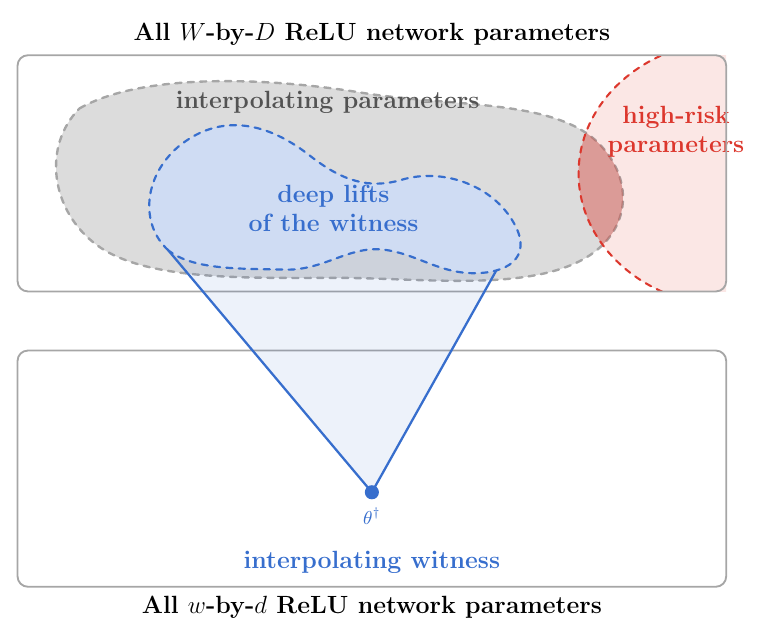}
  \caption{A small interpolating network ``witnesses'' many large interpolating networks. Blue volume dwarfs the intersection of the gray and red volumes once $n \gg wWD$.}
  \label{fig:teaser}
\end{wrapfigure}

Modern deep learning generalization poses a problem for classical statistical learning theory (SLT). SLT typically formulates generalization bounds in terms of the complexity of the underlying hypothesis class, but the class of functions realized by overparameterized neural networks is complex enough to
fit even randomly assigned labels \citep{zhang2017understanding}.

It is broadly believed that the explanation for this apparent contradiction is that modern learning 
pipelines implicitly constrain learned deep neural networks (DNNs) to be ``simple'' \citep{vallePerez2019deep, berchenko2024simplicity, wilson2025position}.  Much recent work searches for this \emph{simplicity bias} in the optimizers used to train DNNs.  For example, gradient-based methods paired with specific loss functions have been shown to preferentially select low-norm/low-rank \citep{gunasekar2017implicit, gunasekar2018implicit, arora2019implicit}, max-margin \citep{soudry2018implicit, lyu2020gradient, chizat2020implicit}, or flat/stable \citep{mulayoff2020unique, arora2022understanding, qiao2024stable, liang2025stable} solutions, even when the surrounding hypothesis class contains far more complex alternatives.

At the same time,  DNN generalization occurs robustly across many choices of loss function and optimizer, and it is unclear whether \emph{optimizer-level} analyses are overindexing on properties of specific optimizer/loss-function pairs \citep{chiang2023loss}. We are interested in the question: 
\begin{center}\textbf{do optimizers need to be secret geniuses to produce generalization?}
\end{center}

Consider as an alternate object of study the ``learning algorithm'' that chooses a width-$W$-depth-$D$ DNN's parameters uniformly at random from those that interpolate the training labels. Although such a learner cannot be feasibly implemented, it represents the ``dumbest possible'' learning rule, because it chooses interpolating parameters without any regard for simplicity. \textbf{We study the generalization properties of this dumb learner, in order to isolate those aspects of generalization that are independent of optimizer bias.} 

How can uniform random interpolating parameters be expected to generalize? Consider the scenario where there exist $w \ll W$ and $d \ll D$ such that some $w$-by-$d$ network approximates the data generating mechanism. In the case where parameter weights are forced to belong to a discrete set, \cite{pmlr-v235-buzaglo24a} show that uniform random interpolating parameters generalize faster than expected. The reason is architectural: any width-$w$ depth-$d$ DNN can be used to construct an exponential number of width-$W$ depth-$D$ networks that compute the same function, so that a $w$-by-$d$ interpolating network \textbf{witnesses} a large number of $W$-by-$D$ interpolators. A counting argument therefore establishes that a uniform random interpolating parameter for a $W$-by-$D$ network generalizes at the rate $\tilde O(\frac{wWD}{n})$, where $n$ is the number of data points, rather than the naive $\tilde O(\frac{W^2D}{n})$ one would expect from uniform convergence. The proof technique of ``lifting'' a small interpolating witness to furnish a vast set of larger interpolators has been used in several recent generalization studies \citep{pmlr-v235-buzaglo24a, harel2024provable, dziugaite2025size}, and we call it \textbf{deep lifting}.

However, existing work in this direction is limited in at least two ways. The first limitation is that parameters are assumed to be drawn from a discrete set of values; this assumption is crucial in prior work, because the set of equivalent ``lifted'' parameters generally has zero probability under a continuous measure. The second limitation is that data is assumed to be noiseless and realizable ---  a necessary expedient for there to exist a small DNN that simultaneously interpolates and generalizes. Although results that address these limitations exist, they rely on further restrictions to the architecture or data model (e.g. shallow or boolean DNNs \citep{pmlr-v235-buzaglo24a, harel2024provable}). 

We extend the technique of deep lifting to (i) cover arbitrarily deep
DNNs with continuous parameters drawn uniformly from a bounded interval, and (ii) handle label noise and unrealizability. 

Suppose we have data $x_1 \dots x_n \in \R^p \overset{i.i.d.}{\sim} \mu_X$ and (multi-dimensional) labels $y_1 \dots y_n \in \R^s$. Let $\mathcal C_M(w,d)$ denote the class of parameter vectors for width-$w$, depth-$d$ DNN with weights inside $[-M, M]$, and write $g_\theta^{w,d}$ for the function realized by $\theta\in\mathcal C_M(w,d)$.  Given training data
$\{(x_i,y_i)\}_{i=1}^n$, let
\begin{equation}\label{body eq:interpolation definitions}
\textstyle S_{\tau, n}^{W, D, M} := \left\{
\phi\in\mathcal C_M(W,D):
\max_{i\in[n]}
\|g_\phi^{W,D}(x_i)-y_i\|_\infty\le\tau
\right\} \qquad 
U_{\tau,n}^{W,D,M}
:=
\operatorname{Unif}\left(S_{\tau, n}^{W, D, M}\right).
\end{equation}
define the uniform distribution over parameters that interpolate up to precision $\tau$. See Appendix \ref{apdx sec:uniform dist def} for a formal definition of $U^{W, D, M}_{\tau, n}$. 
For a target $f$, define the ``interpolation risk at precision $\tau$'' as
\begin{equation}\label{body eq:delta def}
\Delta_\tau(\theta;f)
:=
\Pr_{x\sim\mu_X}
\left[
\|g_\theta(x)-f(x)\|_\infty>\tau
\right].
\end{equation}

We study $\Delta_\tau(\hat \theta; f)$ under $\hat \theta \sim U^{W, D, M}_{\tau, n}$, under various data-generating mechanisms. Some of our main findings are the following. 

 \noindent \textbf{Noiseless/realizable extension from continuous deep lifting.} Assume $y_i = f(x_i)$. Furthermore, assume that $f$ is realizable (up to precision $\tau/2$) by some witness $\theta \in \mathcal C_M(w, d)$, i.e.  $\sup_{x \in [-X, X]^p}\|f(x) - g_\theta^{w, d}(x)\|_\infty \leq \tau/2$. Then Theorem \ref{body thm:continuous noiseless thm} shows
    \[\textstyle \Pr_{x^n, \hat{\theta}}\left[\Delta_\tau(\hat{\theta}; f) \geq \tilde O\left(\frac{wdWD}{n}\right)\right] \leq \delta,\] 
    where $\tilde O(\cdot)$ hides a linear contribution from $p$ and $s$, as well as log terms. In general, the factor of $d$ can be replaced by a quantity depending on fine-grained ``stability'' properties of $\theta$, which is often polylogarithmic (see examples in Section \ref{subsec:noiseless realizable} and Lemma \ref{body lem:gen from structured stab}). 
    
 \noindent\textbf{Noisy/unrealizable settings} Assume that, of the $n$ training points, $m$ labels are (possibly adversarially) corrupted by a  bounded perturbation, while all other labels are from the ground truth $f$. Then, under the same witness assumption as above, $\hat{\theta}$  still generalizes with respect to $f$ at the rate:
    \[\textstyle \Pr_{x^n, \hat{\theta}}\left[\Delta_\tau(\hat{\theta}; f) \geq \tilde O\left(\frac{W(wD + m)d}{n}\right)\right] \leq \delta\]
    even though we force it to interpolate adversarial noise (Theorem \ref{body thm:noisy informal}).

     By this token, when $m \asymp \epsilon n$ grows as a constant fraction of $n$, we have a tempered overfitting type rate. Corollary \ref{body cor:tempered regression} shows that $n$ can be chosen so that, with high probability
    \[\textstyle \Delta_\tau(\hat\theta; f) = \tilde O(W\epsilon).\]
    
    when depth $D_n$ grows with $n$. Similarly, we are able to certify generalization when $f$ is not uniformly approximable by any DNN for any $\tau$ (Theorem \ref{body thm:unrealizable thm informal}).


    

\noindent\textbf{Classification}
We derive analogous rates for classification. Here, the $s$ network outputs are interpreted as logits for $s$ classes, and a network parameter is said to interpolate the data if its argmax logit is always the correct class. As a corollary of these results, we are able to extend generalization results to $\hat{\theta}$ sampled from the Boltzmann distribution built on cross entropy loss and conditioned on interpolation. Due to space constraints, we report these results in Appendix \ref{apdx:classification}.


\noindent\textbf{Key techniques}
The paper's proofs are based on a central technique that we call \textbf{deep lifts}. Heuristically: a ``witness'' $\theta \in \mathcal C_M(w, d)$ can be ``lifted'' into an equivalent parameter in $\mathcal C_M(W, D)$ in many ways. Thus, the existence of a single ``good'' width-$w$ depth-$d$ implies a large amount of probability mass on ``good'' width-$W$ depth-$D$ parameters (conditioned on interpolation) \citep{pmlr-v235-buzaglo24a, harel2024provable, dziugaite2025size}.

The basic difficulty introduced by the continuous setting is producing a set of ``deep lifts'' with positive measure. While in the discrete case, some subset of the witness network's parameters could be frozen while maintaining exponential lifted mass, the continuous case requires that each weight vary in a neighborhood of the corresponding weight in the witness. For the witness behavior to survive such perturbations in the weights, we identify a sufficient condition that we call ``structured stability,'' which we also tie to the witness' ability to reject small perturbations injected at each layer.

Another crucial difficulty of the analysis is dealing with label noise in the interpolation setting. By requiring interpolation, we require that the network $g_{\hat{\theta}}$ explicitly disagree with the ground truth on the training set; but we proceed to evaluate $g_{\hat{\theta}}$ based on its population agreement with $f$. Here, our proof technique is to design a memorization gadget that learns localized spikes in order to fit corruptions. This gadget can be ``bolted on'' to a witness $\theta \in \mathcal C_M(w, d)$ that realizes $f$, generating an interpolating parameter $\theta^\dagger \in \mathcal C_M(w', D)$, upon which we may apply our analysis, whenever $W > w'$. Importantly, the memorization gadget appears only in the proof (to furnish a small interpolating witness), and it is not assumed or expected that $\hat{\theta}$ explicitly implements the gadget. 

\vspace{-0.5em}
\subsection{Related work}
\vspace{-0.5em}

\cite{vallePerez2019deep, depalma2019random} provide early evidence showing that uniform distributions over parameters push forward onto non-uniform distributions over implemented functions; \cite{mingard2021sgd, mingard2025occam} correspondingly analyzes SGD from the perspective of Bayesian sampling. \cite{berchenko2024simplicity} shows that random constructions can exhibit simplicity bias  in overparameterized models. Unlike most of these works, we specifically consider uniformity \emph{conditioned on interpolation}. Most relevant to our work are \cite{pmlr-v235-buzaglo24a}, who show that randomly sampled DNNs also generalize in the case of noiseless data from a realizable ground truth; \cite{harel2024provable}, who show that this analysis can be extended to noisy binary classification, but considers only DNNs with binary weights; and \cite{dziugaite2025size}, who reformulate the proof technique of \cite{pmlr-v235-buzaglo24a}, and consider DNNs drawn from loss-based Boltzmann distributions rather than only interpolating DNNs. To our knowledge, prior work does not cover deep networks with continuous weights, or label noise in a general setting. 
Extended discussion of related work is in Appendix \ref{apdx:related work}.

\vspace{-0.5em}
\section{Preliminaries}
\vspace{-1em}

We are interested in approximating functions mapping $\R^p \rightarrow \R^s$ with ReLU Deep Neural Networks. Given $p$ and $s$ of interest\footnote{treated as constants in this paper}, we define a width-$w$ depth-$d$ neural network to be given by a set of parameters, $\theta$. We consider $\theta$ as organizing into a parameter vector the weight matrices
\[W^\theta_1 \in \R^{w\times p}, W^\theta_k \in \R^{w\times w} \text{ for $k = 2 \dots d$,}\]
as well as biases $b^\theta_k \in \R^{w} \text{ for $k = 1 \dots d$}$
and a readout matrix $R^\theta \in \R^{s\times w}$. As such, $\theta$ is a vector with $N_{w, d} = w^2(d - 1) + wd +wp + ws$ entries. 

The neural network defined by $\theta$ processes inputs $x \in \R^p$ according to $x^\theta_k =\sigma(W^\theta_k x^\theta_{k - 1} + b^\theta_k)$
with $x^\theta_0 := x$ and $\sigma(z) = \max\{z, 0\}$ defined to be an elementwise application of ReLU. We call $x^\theta_k$ the layer-$k$ activations. The function output defined by the neural network is $g^{w, d}_\theta (x) = R^\theta x^\theta_d(x)\in \R^s$.

For a real number, $M > 0$, we define $\mathcal C_M(w, d)$ to be the set of all parameter vectors for width-$w$ depth-$d$ neural networks with real-valued elements in $[-M, M]$. Thus, $\mathcal C_M(w, d)$ may be identified with $[-M, M]^{N_{w, d}}$. For a sequence $M_n \to \infty$, we define $\mathcal C(w, d) = \bigcup_n \mathcal C_{M_n}(w, d)$, which may likewise be identified with $\R^{N_{w, d}}$. 

\section{Regression setting}\label{sec:regression}

In the (multi-valued) regression setting, we receive data $\{(x_i, y_i)\}_{i = 1}^n \subset \R^p \times \R^s$. We assume that the covariates $x_i$ are drawn i.i.d. from a measure $\mu_X$ on $p$ dimensions supported within $[-X, X]^p$ for some bound $X > 0$. For now, labels are left as general vectors. 

We generally assume that there exists some ground truth $f: \R^p \rightarrow \R^s$ against which we measure the performance of a parameter $\theta \in \mathcal C(w, d)$. In the paper body, this performance is measured  by \textbf{interpolation risk} at precision $\tau > 0$, or simply \textbf{risk} at $\tau$, notated as $\Delta_\tau(\theta; f)$ and defined in Equation \eqref{body eq:delta def}.  Because $f$ will always be clear from context in this paper, we often write only $\Delta_\tau(\theta)$. Given data $\{(x_i, y_i)\}_{i =1 }^n$, target width $W$ and depth $D$, and $\tau > 0$, we recall the definition of  $S^{W, D, M}_{\tau, n}$ as the set of ``$\tau$-interpolating parameters,'' according to \eqref{body eq:interpolation definitions}, and $U^{W, D, M}_{\tau, n}$ as the Lebesgue-induced uniform distribution over $S^{W, D, M}_{\tau, n}$ (technical definition in Appendix \ref{apdx sec:uniform dist def}). In this paper, our goal is to study the generalization properties of $\hat{\theta} \sim U_{\tau, n}^{W, D, M}$, under various assumptions on the mechanism generating the labels $\{y_i\}$.

\begin{remark}[Interpretation of $\tau$]
$\tau$ should be interpreted as a numerical precision. In practice, it is common to say a model ``interpolates'' if its predictions are within $\tau = 10^{-6}$ of the true values. 
\end{remark}

\begin{remark}[Why study $\Delta_\tau(\hat{\theta}; f)$?]\label{body remark:L1 conversion}

$\Delta_\tau(\hat{\theta}; f)$ has a clear interpretation as the population probability of making an incorrect continuous prediction, at the tolerance $\tau$. Moreover, in the case where $s = 1$,   $|f(x)| \leq F$, and we define $\hat{f}$ by clamping $g_{\hat{\theta}}^{W, D}$ between $[-F, F]$, then a bound on $\Delta_\tau(\hat{\theta}; f)$ can be converted into an $L_\alpha$-norm error bound for an exponent $\alpha \geq 1$:
\[\|\hat{f} - f\|_{L_\alpha(\mu_X)}^\alpha := \E_{x \sim \mu_X}\left[\left|\hat{f}(x) - f(x)\right|^\alpha\right]\leq \tau^\alpha  + (2F)^\alpha\Delta_{\tau}(\hat{\theta}; f).\]
In this case, depending on the behavior of $\Delta_\tau(\hat{\theta}; f)$ as a function of $n$ and $\tau$, we may choose $\tau$ \emph{as a function of $n$ and $\alpha$} to derive $L_\alpha$ rates.
\end{remark}
\subsection{Regression with noiseless and realizable data}\label{subsec:noiseless realizable}

We now state our first result, which is an extension of \cite{pmlr-v235-buzaglo24a} to continuous weights. Theorem \ref{body thm:continuous noiseless thm} says that, if we receive noiseless observations from a function $f$ that is uniformly approximated by a small neural network $\theta \in \mathcal C_M(w, d)$, then we generalize at the rate $\tilde O(\frac{wdWD}{n})$.

\begin{theorem}[Continuous noiseless realizable setting]\label{body thm:continuous noiseless thm}
Let $\tau>0$ and $\delta\in(0,1)$. In the dataset $\{(x_i, y_i)\}_{i = 1}^n$ assume that $y_i = f(x_i)$ for all $i = 1 \dots n$.

Let $M \geq 1$. Assume that, for some $w$ and $d$, there exists $\theta \in \mathcal C_M(w, d)$ so that $\|f(x) - g_{\theta}^{w, d}(x)\|_\infty \leq \frac{\tau}{2}$ for all $x \in \supp (\mu_X)$. Let $D \geq d$ and $W > w$. Then for $\hat{\theta} \sim U^{W, D, M}_{\tau, n}$
\[\textstyle \Pr_{x^n, \hat{\theta}}\left[\Delta_\tau(\hat{\theta}) \geq  t_n\right] \leq \delta, \qquad t_n := \frac{C_0wWD\log q_{\mathrm{eff}} + \log\frac{1}{\delta}}{n}.\]
Here, $C_0$ is a constant depending linearly on $p$ and $s$, and $q_{\mathrm{eff}}$ is a value satisfying 
\[\textstyle \log q_{\mathrm{eff}} \leq (d+2)\log \left(16MwD(pX+2)\max\{1,\tau^{-1}\}\right).\]
\end{theorem}

We emphasize a key feature of Theorem \ref{body thm:continuous noiseless thm}: since the bound holds for every witness that
(i) fits inside the ambient $W$-by-$D$ architecture and (ii) explains the data-generating mechanism, the result may be optimized over all such witnesses; in this sense, the theorem is an adaptivity result, showing that the generalization rate adapts to the size of the \emph{smallest} explanation available within the architecture. Though
the size of the smallest network representing the ground truth is unknown, the cost of choosing an overlarge architecture is relatively mild, which motivates overparameterization. 

In Theorem \ref{body thm:continuous noiseless thm}, the quantity $q_{\mathrm{eff}}$ plays a role analogous to the alphabet size in \cite{pmlr-v235-buzaglo24a} (i.e. the cardinality of the discrete set from which weights are drawn). Unlike in the discrete setting of \cite{pmlr-v235-buzaglo24a}, however, our generic bound incurs a linear dependence on the witness depth $d$ through $\log q_{\mathrm{eff}}$.\footnote{This dependence may be considered mild, as $w$ and $d$ are viewed as small constants relative to $W$, $D$ \citep{pmlr-v235-buzaglo24a}.} The source of this dependence is that, in the continuous setting, a witness must continue to implement nearly the same function when its parameters are perturbed within a small neighborhood. For generic networks,  control of this ``stability'' can deteriorate rapidly (exponentially) with $d$. However, specific deep witnesses may in fact have stronger stability properties. Theorem \ref{thm:disturbance stable rate}, explicitly gives the rate in terms of the ``stability'' of the witness (as defined in Section \ref{sec:proof sketch}), and the second example below shows that doing so can lead to sharper statistical rates when a deep but ``stable'' witness can be procured.

\noindent\textbf{Example: H\"older functions}
Suppose $\supp(\mu_X) \subseteq [0, 1]^p$ and let $f: [0, 1]^p \rightarrow \R$ satisfy H\"older smoothness with index $\beta > 0$ and radius $K> 0$, meaning  it has bounded partial derivatives up to order $\lfloor \beta \rfloor$, and its $\lfloor \beta \rfloor$-order partial derivative are $\beta - \lfloor \beta\rfloor$ H\"older continuous \citep{schmidt2020nonparametric}. Using the approximation theoretic results from \cite{ohn2019smooth}, we find that $\Delta_\tau(\hat{\theta}) = \tilde O\left(\frac{WD\tau^{-p/\beta}}{n}\right)$ with high probability, as long as $W$ and $D$ are large enough. If $f$ is bounded by $F$, then following Remark \ref{body remark:L1 conversion} and tuning $\tau$ leads to the high probability $L_1$ bound
\vspace{-1mm}
\[\textstyle \left\|\hat{f} - f\right\|_{L^1(\mu_X)} = \tilde O\left(n^{-\frac{\beta}{\beta + 2p}}\right).\]
For reference, without additional regularity assumptions on $\mu_X$, even infinitely smooth functions have a lower bound of $\Omega\left(\frac{1}{n}\right)$ \citep{kohler2013optimal}; the exponent in our rate approaches this barrier as $\beta$ grows with $p$ fixed. Further details for examples can be found in Appendix \ref{apdx:examples}.

\noindent\textbf{Example: Barron functions} Now assume $\supp(\mu_X) \subseteq B_p$, where $B_p$ is the Euclidean unit ball in $p$ dimensions.  ReLU-Barron functions $f: B_p \rightarrow \R$ are functions realizable as ``infinite-width'' ReLU neural networks:
\begin{equation}\label{eq:barron func def}\textstyle f(x) = \E_{a, b, c \sim \rho}\left[a \sigma(b^Tx + c)\right].\end{equation}
The ReLU-Barron norm $\|f\|_\mathcal{B}$ is defined as $\inf_{\rho}\E\left[|a|(\|b\|_1 + |c|)\right]$, where the infimum is taken over measures $\rho$ that realize \eqref{eq:barron func def}.  Let $f$ be ground truth and $Q := \|f\|_\mathcal B$. In Appendix \ref{apdx sec:Barron theory}, we find that a witness of depth $d_\tau\asymp (Q/\tau)^{\frac{2p}{p+3}}$ and width-$(2p+3)$, approximates  $f$ uniformly on $B_p$ to precision $\tau/2$. Moreover, this witness's ``structured stability'' (to be defined in Section \ref{sec:proof sketch}) deteriorates only linearly in $d_\tau$. Taking $D\asymp d_\tau$, Corollary \ref{cor:structured parameter rate} gives $\Delta_\tau(\hat\theta)
\leq
\tilde O(
(W/n)(Q/\tau)^{\frac{2p}{p+3}}
)$ w.h.p whenever $W>2p+3$.  $f$ is bounded by $Q$, and tuning $\tau$ with Remark \ref{body remark:L1 conversion} gives the rate
\vspace{-1mm}
\[\textstyle 
\left\|\hat f - f\right\|_{L_2(\mu_X)}^2=
\tilde O\left(
Q^2\left(\frac{W}{n}\right)^{\frac{p+3}{2p + 3}}
\right).
\]
For fixed $p,Q$ and $W>2p+3$, this rate matches, up to logarithmic factors, the squared-$L_2$ minimax lower bound for Gaussian-noise regression \citep[Remark 4.3]{yangzhou2025optimal} and the best known upper-bound exponent for noiseless i.i.d. sampling \citep[Equation (39), $k=1$]{parhi2025upper}.

\subsection{Regression with corrupted labels}

We now extend the results of the previous section to cover data with noise. Specifically, we cover the case where an adversary is allowed to apply a bounded perturbation to any subset of the labels before $\hat{\theta}$ is drawn. 

\begin{assumption}[$(m, \Gamma)$ adversarial corruptions]\label{body ass:adversarial corruptions}
Suppose the ground truth function is $f$. The dataset $(x_i, y_i)$ is $(m, \Gamma)$ adversarially corrupted if, after seeing $x_1 \dots x_n \sim \mu_X$, an adversary chooses any subset $I \subseteq [n]$ with size $|I| \leq m$, and $\{\delta_i\}_{i \in I} \subset [-\Gamma, \Gamma]^s$, and reveals to the learner the labels
\[\textstyle y_i = f(x_i) + \delta_i \ (i \in I),  \qquad y_i = f(x_i) \  (i \notin I).\]
\end{assumption}

Our result requires the following mild regularity condition on the distribution $\mu_X$. 

\begin{assumption}[$\Lambda$ marginal boundedness of $\mu$]\label{body ass:lambda marginal boundedness}
For $\Lambda>0$, we say that $\mu_X$ is $\Lambda$-marginally bounded if the marginal of the first coordinate of $x\sim\mu_X$ satisfies $\Pr[x[1]\in I]\leq\Lambda|I|$ for every interval $I\subseteq[-X,X]$.\footnote{We take the first coordinate only for simplicity. Indeed, we can derive similar results under the assumption that there exists some fixed vector $a \in \R^p$ so that $x^Ta$ has bounded density.}
\end{assumption}

We now have a similar result to Theorem \ref{body thm:continuous noiseless thm}, which says that, even in the face of adversarial corruptions, if the ground truth $f$ is approximated to precision $\tau/2$ by some $\theta \in \mathcal C_M(w, d)$, we may bound the generalization error of $\hat{\theta} \sim U_{\tau, n}^{W, D, M}$ with high probability. A formal version of Theorem \ref{body thm:noisy informal} is stated as Theorem \ref{apdx thm:continuous noisy main thm} in the Appendix. 

\begin{theorem}[Regression with corrupted labels, informal]
\label{body thm:noisy informal}
Let $\tau > 0$, $\delta \in (0, 1)$, and $n\geq2$. Suppose that our data-generating mechanism satisfies Assumptions \ref{body ass:adversarial corruptions} and \ref{body ass:lambda marginal boundedness}, and assume that $f$ is realizable up to precision $\tau/2$ by some witness
$\theta\in\mathcal C_M(w,d)$.

Assume moreover that $M\geq \mathrm{poly}(n,\Lambda,\Gamma,X,\delta^{-1})$, where $\mathrm{poly}$ is a fixed polynomial stated in the formal version.  Let $D\geq d$ and $W\gtrsim w + s + \frac{m}{D}$. Then, for $\hat{\theta} \sim U_{\tau, n}^{W, D, M}$
\[\textstyle 
\Pr_{x^n, \hat{\theta}}\left[\Delta_\tau(\hat\theta;f)
\geq
 C_0
\frac{W(wD+m)\log q_{\mathrm{eff}, n}}{n}
\right] \leq \delta,
\]
where $\log q_{\mathrm{eff}, n} \leq 2 \log \left(\mathrm{poly}(n,m, W, w, d,   \Gamma, \Lambda, X, p, s, \delta)DM\max\{1,\tau^{-1}\}\right)$ and $C_0$ is a constant depending only on $p$ and $s$.
\end{theorem}

Let us interpret Theorem \ref{body thm:noisy informal}. The crucial witness assumption has remained unchanged: $f$ is approximately realized by a small $\theta \in \mathcal C_M(w, d)$. The key conclusion is also the same; we get a generalization bound for $\hat{\theta} \sim U_{\tau, n}^{W, D, M}$. However, since it is no longer assumed that all $n$ labels come from $f$, we require additional assumptions on $M$ and $W$. Since $M$ enters the bound only under a $\log$, choosing it to satisfy the required polynomial dependence is mild. There is additionally a new requirement on $W$. While in the noiseless setting, we needed only $W > w$, we now need $W > w + w_+$, for some slack $w_+$ that increases according to the ratio $m/D$. As we will see, this is because $\theta \in \mathcal C_M(w, d)$ does not itself interpolate the noisy points, and we use the extra width to memorize $m/D$ points per layer. The upshot is that the noisy rate of Theorem \ref{body thm:noisy informal} incurs an extra $\frac{Wmq_{\mathrm{eff}, n}}{n}$ compared to the noiseless rate of Theorem \ref{body thm:continuous noiseless thm}. 

If $m = o_n(n)$, then $\Delta_\tau(\hat{\theta}) = o_n(1)$ in Theorem \ref{body thm:noisy informal}. However, a key case of interest is when $m$ scales as $m \asymp \epsilon n$. The following corollary finds that the assumptions of Theorem \ref{body thm:noisy informal} are satisfied for $W > w +s+  O(1)$ and $D = \Theta(n)$. Plugging in $m \asymp \epsilon n$ together with these values gives a tempered-overfitting type risk floor at $\tilde O(W \epsilon)$ when $W \gg w$. The formal statement is Theorem \ref{cor: continuous epsilon theorem}.

\begin{corollary}[Tempered-type overfitting under constant label noise, informal]\label{body cor:tempered regression}

In the setting of the previous theorem, if $m \sim \mathrm{Binom}(n, \epsilon)$, $W \gtrsim w + s + B$ for some $B \in \N$, then there is some depth scaling $D = \Theta(n)$ so that the following holds. 

At $n = \max\left\{2,\left\lceil \frac{\log(2/\delta)}{\epsilon} \vee \frac{Bd-\log(2/\delta)}{\epsilon}\right\rceil\right\}$, we have
\[\textstyle \Pr_{x^n \sim \mu_X^, \hat{\theta} \sim U^{W, D, M}_{\tau, n}}\left[\Delta_\tau(\hat{\theta}; f) > 
\tilde O\left(Wd\left(1+\frac{w}{B}\right)\epsilon 
\right)\right] \leq \delta.\]
\end{corollary}
 
Empirical work finds that interpolating solutions typically exhibit ``tempered overfitting,'' rather than converging to the optimal risk \citep{mallinar2022benign} in noisy settings, so it is not surprising that we observe a nonzero risk floor. We also point out a weakness. Corollary \ref{body cor:tempered regression} is not an asymptotic result, but rather holds at a fixed value of $n$. Because of the $\log n$ dependence in Theorem \ref{body thm:noisy informal}, there are a wide range of $n$ that give the $\tilde O(W\epsilon)$ rate, but eventually the guaranteed risk diverges as $\Theta(\log n)$. 


\subsection{Regression with unrealizable ground truth}\label{subsec:unrealizable}

We now come to results that cover the case where the ground truth, $f$, cannot be realized (i.e. uniformly approximated) by any ReLU neural network. Instead of assuming realizability, we assume that there is a sequence of witnesses $\theta_{n} \in \mathcal C_{M_n}(w_n, d_m)$ that approximate $f$ to precision $\tau$ on some set of measure at least $1 - \epsilon_n$. Then, if our $W_n$-by-$D_n$ architecture scales so that these witnesses remain admissible, we may generalize at a rate defined by $\epsilon_n$.

\begin{theorem}[Unrealizable ground truths, informal]\label{body thm:unrealizable thm informal}
Let $\delta \in (0, 1)$ and $\tau > 0$. Assume the noiseless setting, i.e. $y_i = f(x_i)$ for all $i \in [n]$. Assume $\mu_X$ satisfies Assumption \ref{body ass:lambda marginal boundedness}. 

Consider sequences $\epsilon_n\in(0,1]$, $w_n$, $d_n$, $D_n$, $A_n$ with the requirements that $\epsilon_n = \Omega(\frac{1}{n})$ and $D_n\geq d_n$. Suppose that, for all $n\in\mathbb N_{\geq2}$, there is $\theta_{n} \in \mathcal C_{A_n}(w_n, d_n)$ such that
\[\textstyle 
\mu_X\left(
\left\{
x:
\|g_{\theta_{n}}^{w_n,d_n}(x)-f(x)\|_\infty
>
\frac{\tau}{2}
\right\}
\right)
\le \epsilon_n, 
\qquad 
\sup_{x\in[-X,X]^p}
\|g_{\theta_{n}}^{w_n,d_n}(x)-f(x)\|_\infty
\le \Gamma.\]

Assume moreover that $M_n\geq \mathrm{poly}(n,\Lambda,\Gamma,X,\delta^{-1})\vee A_n$, where $\mathrm{poly}$ is a fixed polynomial stated in the formal version, and $W_n >w_n + w_+$, where $w_+  = \tilde O\left(1 + \frac{\epsilon_n n}{D_n}\right)$. Consider $\hat \theta \sim U^{W_n, D_n, M_n}_{\tau, n}$. Then 
\[\textstyle \Pr_{x^n, \hat{\theta} }\left[\Delta_\tau(\hat{\theta}) \geq
\tilde O\left(W_n\left(\frac{w_nD_n + \log \frac{1}{\delta}}{n} +\epsilon_n\right)\log q_\mathrm{eff, n}\right) \right] \leq \delta\]

where $\log q_\mathrm{eff} = d_n \log \mathrm{poly}(n, W_n, D_n, w_n, d_n, \Lambda, \Gamma, M_n, \tau^{-1}, \delta^{-1})$. 
\end{theorem}

Technically, Theorem \ref{body thm:unrealizable thm informal} bears a resemblance to Corollary \ref{body cor:tempered regression}, due to the additive $\tilde O(W_n \epsilon_n)$ term. However, in the noisy case, $\epsilon$ is a constant rate parameter, whereas in the unrealizable case, we may have $\epsilon_n = o_n(1)$, allowing us to achieve $\Delta_\tau(\hat{\theta}; f) = o_n(1)$.

\noindent\textbf{Example: Learning half spaces} Consider the space of functions indicating $p$-dimensional half spaces:
\[
\mathrm{Halfspaces}(R)
=
\left\{
f_{a,b}(x)=\mathbf{1}\{a^Tx-b>0\}
:
a\in\mathbb S^{p-1},\ |b|\leq R
\right\}.
\]
Notice that for any $f \in \mathrm{Halfspaces}(R)$, the $\tau$-realizability condition of Theorem \ref{body thm:continuous noiseless thm} is unsatisfiable for any $\tau < 1$. However, under a mild regularity condition on $\mu_X$, it is straightforward to define a sequence of $\theta_n \in \mathcal C_{A_n}(2, 1)$ with $A_n = \Theta(n)$ so that the hypotheses of Theorem \ref{body thm:unrealizable thm informal} hold with $\epsilon_n = \frac{1}{n}$ and $\Gamma = 1/2$. These are the $\theta_n$ implementing ramp-type approximations to $f$. Details for this construction are in Appendix \ref{apdx:examples}.

Choose $M_n$ polynomial in $n$ and the problem parameters, large enough to satisfy the theorem statement, and take $D_n\geq1$. For $\delta\in(0,1)$, set \[\textstyle W_n>6+4\left\lceil\frac{2 + \log \frac{4}{\delta}}{D_n}\right\rceil.\]
Then, for $\hat{\theta}\sim U^{W_n,D_n,M_n}_{\tau,n}$,
we have $\Pr_{x^n,\hat{\theta}}
[
\Delta_\tau(\hat{\theta})
\geq
\tilde O(
\frac{W_nD_n}{n}
)
]\leq\delta$ when $\tau\asymp1/n$.
According to Remark \ref{body remark:L1 conversion}, the corresponding $L_1$ rate is 
\[\textstyle \|\hat f - f\|_{L_1(\mu_X)}\leq\tau+\Delta_\tau(\hat\theta).\]
Consequently, for $\tau\asymp1/n$, with probability at least $1-\delta$,
\[\textstyle \|\hat f-f\|_{L_1(\mu_X)}=\tilde O\left(\frac{W_nD_n}{n}\right).\]

\section{Proof Sketch}\label{sec:proof sketch}

We now sketch the proofs for Theorems \ref{body thm:continuous noiseless thm} and \ref{body thm:noisy informal}. Formal proofs are deferred to the Appendix 

\noindent\textbf{Structured Stability of Parameters} We begin by introducing the notion of ``structured stability,'' which quantifies how similarly a parameter  $\theta \in \mathcal C_M(w, D)$ behaves to its nearby neighbors.

\begin{definition}[Structured stability]
A parameter $\theta\in\mathcal C_M(u,D)$ has $(H,L,C)$, with $H,L,C>0$,
\emph{structured stability} if (i) $\max_{k \in [D]}\sup_{x\in\mathcal X}
\|x_k^\theta(x)\|_\infty\le H$
and (ii) for every $\tilde\theta\in\mathcal C_M(u,D)$ and $\rho\le C^{-1}$ satisfying $\|\tilde\theta-\theta\|_\infty\leq\rho$, we have 
\[
\textstyle 
\sup_{\substack{x\in\mathcal X\\ k\in[D]}}
\|x_k^{\tilde\theta}(x)\|_\infty\le H+1 \quad \text{and} \quad 
\sup_{x\in\mathcal X}
\|g_{\tilde\theta}^{u,D}(x)-g_\theta^{u,D}(x)\|_\infty
\le L\rho.
\]
\end{definition}
In other words, $\theta$ has bounded activations and all of its $C^{-1}$-close neighbors (i) enjoy a similar activation bound (ii) give similar outputs for all input in the support of $\mu_X$. In the next section, we show that $(H, L, C)$ govern the volume of a set of  parameters $\phi \in \mathcal C_M(W, D)$ constructed to ``behave like'' $\theta$, and therefore generalization in the larger architecture. 

\noindent\textbf{Deep Lifts for Stable Parameters}
\begin{figure}[!t]
\vspace{-30pt}
    \centering
\includegraphics[width=0.8\linewidth]{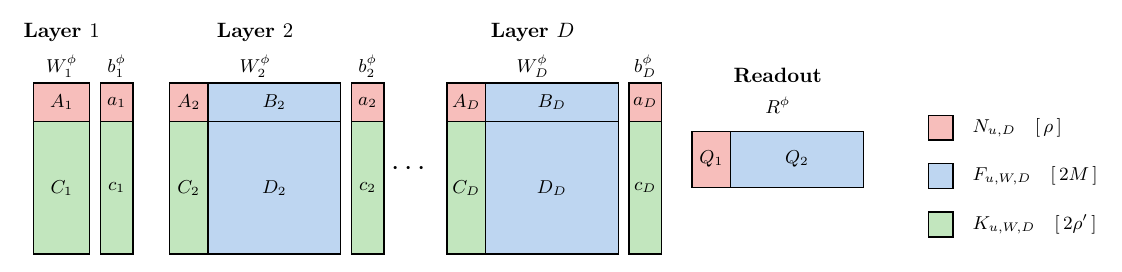}
    \caption{A visualization of the structure of a deep lift$, \phi \in \mathcal C_M(W, D)$,  of $\theta \in \mathcal C_M(u, D)$. Most of $\phi$'s weights (in blue), may be chosen arbitrarily from $[-M, M]$. Those in green are chosen from an interval of size $2\rho'$. Those in red are chosen from an interval of size $\rho$ (built around $\theta$'s weights). The result is a high-volume set of nearly equivalent lifts of $\theta$. }
    \label{fig:placeholder}
\end{figure}
Given a parameter $\theta\in\mathcal C_M(u,D)$, our proof
technique relies on producing a large volume of parameters
$\phi\in\mathcal C_M(W,D)$ that compute similar functions. We now construct
such a set and write its volume in terms of the structured stability
of $\theta$.

Let $\theta\in\mathcal C_M(u,D)$ have $(H,L,C)$ structured stability. Write
the parameters of $\phi$ in block form
\[ \textstyle 
W_1^\phi=
\begin{pmatrix}
A_1\\ C_1
\end{pmatrix},
\qquad
W_{k\geq 2}^\phi=
\begin{pmatrix}
A_k&B_k\\
C_k&D_k
\end{pmatrix},
\qquad
b_k^\phi=
\begin{pmatrix}
a_k\\ c_k
\end{pmatrix},
\qquad
R^\phi=(Q_1\;Q_2),
\]
where the block sizes follow from dividing the activation vector into a ``carried'' state of dimension $u$, and ``added'' state of dimension $W - u$. First, we explain that it is possible to arrange for the carried activations to emulate $\theta$, and for the added activations to remain identically zero:
\begin{equation}
\label{eq:pf sketch activation desideratum}
\textstyle 
x^\phi_k[i]
=
\begin{cases}
x^\theta_k[i], & i=1,\ldots,u,\\
0, & i=u+1,\ldots,W.
\end{cases}
\end{equation}
The proof is by induction. By structured stability, we have $\max_k \|x^\theta_k(x)\|_\infty \leq H$. If \eqref{eq:pf sketch activation desideratum} holds at $k$, then $\max_{j \geq u + 1}\left|W^\phi_k x^\phi_{k -1}[j]\right| \leq Hu\rho'$. Therefore, If $\rho'$ is chosen small enough, we can choose the biases $c_k \in [-M, -M + 2\rho']$ so that the added pre-activations are negative, and the ReLU nonlinearity zeros them out. At the same time, under \eqref{eq:pf sketch activation desideratum}, the choice $A_k = W^\theta_k$ and $a_k = b^\theta_k$ suffices to establish $x^\phi_{k + 1}[i \leq u] = x^\theta_{k + 1}$. Crucially, $B_k$ and $D_k$ have been left to choose their elements arbitrarily from $[-M, M]$. At readout, we may fix $Q_2$ arbitrarily.


This already gives many parameters $\phi$ implementing the same function as $\theta$, but the construction has zero Lebesgue measure as a subset of $\R^{N_{W, D}}$, because the carried parameters $A_k,a_k, Q_1$ are fixed. The solution is to allow each carried weight to vary over an interval of length $\rho > 0$ around its corresponding
value in $\theta$.  We call the constructed set of parameters $\phi \in \mathcal C_M(W, D)$  the set of ``deep lifts of $\theta$ from $u$ to $W$,'' notated as $\mathscr L_{u\rightarrow W}^{\rho,\rho'}(\theta)$. For suitably small radii $\rho$ and $\rho'$, Lemma \ref{lem:lift behaves like structured parameter} ensures that every $\phi \in \mathscr L^{\rho,\rho'}_{u\rightarrow W}(\theta)$ satisfies
\[\textstyle \sup_{x \in [-X, X]^p} \|g_\theta^{u, D}(x) - g_\phi^{W, D}(x)\|_\infty \leq \frac{\tau}{2}.\]

\vspace{-1.5mm}
By construction, the Lebesgue measure of this set decomposes according the product of the intervals from which each individual weight is chosen, i.e.
\begin{equation}
\label{body eq:continuous lift volume}
\textstyle \lambda\!\left(
\mathscr L_{u\rightarrow W}^{\rho,\rho'}(\theta)
\right)
=
\underbrace{(2M)^{F_{u,W,D}}}_{\text{from $B_k$, $D_k$}}
\underbrace{\rho^{N_{u,D}}}_{\text{from $A_k$, $a_k$}}
\underbrace{(2\rho')^{K_{u,W,D}}}_{\text{from $C_k$, $c_k$}},
\end{equation}
where $F_{u,W,D}$ counts the freely chosen parameters, $N_{u,D}$ the
carried parameters chosen from intervals of length $\rho$, and
$K_{u,W,D}$ the parameters chosen from intervals of length $2\rho'$. Notice $\rho$ and $\rho'$ both shrink as the structured stability parameters increase, so smaller stability parameters correspond to more lift volume. We summarize with the key upshot. 

\begin{lemma}[Informal]\label{body lem:lift set interpolates}
Let $\theta \in \mathcal C_M(u, D)$ have $(H, L, C)$-structured stability and suppose it $\tau/2$-interpolates realized data $\{(x_i, y_i)\}_{i = 1}^n$. Then, for $\rho = 1/C \wedge \tau/(2L)$ and $\rho' \approx 1/H$, the set $\mathscr L_{u\rightarrow W}^{\rho,\rho'}(\theta) \subset \mathcal C_M(W, D)$ has measure given in Equation \eqref{body eq:continuous lift volume} and each $\phi \in \mathscr L_{u\rightarrow W}^{\rho,\rho'}(\theta)$ $\tau$-interpolates data. See Lemma \ref{lem:lift behaves like structured parameter} in the Appendix for the formal statement and precise radii.
\end{lemma}

\noindent\textbf{Generalization from Stable Interpolators} We now state the main technical result that powers all of our results and sketch its proof. 

\begin{lemma}[Generalization from Structured Stability, Corollary \ref{cor:adaptive structured parameter rate} in Appendix]\label{body lem:gen from structured stab}
Assume $y_i=f(x_i)$ for all labels except at most $m$ (possibly chosen adversarially). Fix $u<W$ and $D,M,H,L,C$ independently of the realized sample, with $M\geq1$, $\tau>0$, $m\leq uD$, and $0\leq\eta<\delta<1$. Suppose there is a training-data event $\mathcal G$ with probability at least $1-\eta$ such that, on $\mathcal G$, there exists $\theta^\dagger\in\mathcal C_M(u,D)$ with $(H,L,C)$ structured stability and
\[\textstyle 
\max_{i\in[n]}
\|g_{\theta^\dagger}^{u,D}(x_i)-y_i\|_\infty
\leq\frac\tau2.
\]
Then, for any $W > u$, we have that, for  $\hat{\theta} \sim U^{W, D, M}_{\tau, n}$:
\[\textstyle 
\Pr_{\hat{\theta}, x^n}[\Delta_\tau(\hat\theta)\geq t_n]\leq\delta, \qquad 
t_n
:=
C_0\frac{uWD\log \bar q_{\mathrm{eff}}+\log\frac1{\delta-\eta}}n
,\]
where $C_0$ depends only on $p$ and $s$, and $\log \bar q_{\mathrm{eff}} \lesssim \log 2M\left(n \vee C \vee 2L/\tau \vee uH\right)$. This bound on $\log \bar q_\mathrm{eff}$ is somewhat simplified, and its precise value can be found in Theorem \ref{thm:stable witness bound} and Corollary \ref{cor:adaptive structured parameter rate}.
\end{lemma}

Lemma \ref{body lem:gen from structured stab} tells us that if, with high probability over the draw of the dataset, there exists a (possibly data-dependent) parameter $\theta^\dagger \in \mathcal C_M(u, D)$ that (i) has structured stability with polynomial parameters (ii) interpolates the data, then $\hat{\theta}$ generalizes at the rate $\tilde O(uWD/n)$. We now outline the proof of Lemma. 

\noindent\textbf{Numerator/denominator decomposition}
Let $\mathcal G$ be the good training-data event in Lemma \ref{body lem:gen from structured stab}. On $\mathcal G$, the lift construction gives positive interpolation volume. Thus, for $\hat\theta\sim U_{\tau,n}^{W,D,M}$,
\begin{equation}
\Pr_{x^n,\hat\theta}[\Delta_\tau(\hat\theta;f)\geq t]
\leq\eta+\Pr(\mathcal G)\,
\mathbb E\!\left[\left.
\frac{\lambda(B_t)}{\lambda(S_{\tau,n}^{W,D,M})}
\,\right|\,\mathcal G\right],
\end{equation}

where $\lambda$ is the Lebesgue measure defined on $\R^{N_{W, D}}$, and $B_t$ is the set of interpolating parameters with risk larger than $t$:
\[\textstyle B_t := S_{\tau, n}^{W, D, M} \cap \left\{\theta \in \mathcal C_M(W, D) \ : \ \Delta_\tau(\theta) \geq t\right\}.\]

We now consider the numerator terms and denominator terms separately. For the denominator, we focus on the event where there is a $\theta^\dagger \in \mathcal C_M(u, D)$ that $\tau/2$ interpolates the data. Under Lemma \ref{body lem:lift set interpolates}, we have $\mathscr L_{u\rightarrow W}^{\rho,\rho'}(\theta^\dagger) \subseteq S_{\tau, n}^{W, D, M}$, so that $\lambda \left(S^{W, D, M}_{\tau, n}\right) \geq \lambda \left(\mathscr L_{u\rightarrow W}^{\rho,\rho'}(\theta^\dagger)\right)$, and we know the exact volume on the RHS.  This bound holds deterministically on $\mathcal G$, so we may pull it out of the expectation. Then we're left with the expectation of the numerator, and we interchange integration to express it as $\int_{\theta \ : \Delta_\tau(\theta) \geq t} \Pr_{x^n \sim \mu_X}\{\theta \in B_t\}d\lambda $. Lemma \ref{lemma:adaptive continuous num_bound} show that the probability in the integrand can be shown to decay like $\binom{n}{m}e^{-t(n - m)}$, so that the whole numerator is bounded roughly as $(2M)^{N_{W,D}}\binom{n}{m}e^{-t(n-m)}$. Combining these bounds and simplifying, under the definition $q_\mathrm{eff} = \left(\frac{2M}{\rho}\right)\vee\left(\frac{M}{\rho'}\right)$ gives Lemma \ref{body lem:gen from structured stab}.

\noindent\textbf {Existence of parameters with structured stability.} In light of Lemma \ref{body lem:gen from structured stab}, each result from the paper follows from showing that a $\tau/2$-interpolator $ \theta \in \mathcal C_M(u, D)$ exists with high probability over the draw of the data, $\{(x_i, y_i)\}_{i = 1}^n$,  and quantifying its stability for each data model proposed. In Theorem \ref{body thm:continuous noiseless thm}, we assumed existence of $\theta \in \mathcal C_M(w, d)$ that $\tau/2$-interpolates every realization of the data, and we show (Lemma \ref{lem:direct padded structured stability}) that simply padding this witness to depth $D$ results in an interpolating witness with stability parameters $\log (H\vee L \vee C) = \tilde O(d)$, which suffices to give the result. For Theorem \ref{body thm:noisy informal}, we explicitly construct a data-dependent network, or ``memorization gadget,'' $\theta_\mathrm{mem} \in \mathcal C_M(w_+, D)$ so that $g_{\theta_\mathrm{mem}}^{w_+, D}$ is $\delta_i$ on all corrupted points, and $0$ on all uncorrupted points. In order to memorize $m$ points, the memorization gadget uses trapezoidal bumps to localize $m/D$ points per layer, as constructed in Lemma \ref{lem:real valued memorization}. Assumption \ref{body ass:lambda marginal boundedness} allows us to choose these bumps to be sharp enough to successfully isolate only corrupted points with high probability (Lemma \ref{lem:probabilistic memorization}). Then we define $\theta^\dagger \in \mathcal C_M(w + w_+, D)$ to simply add these corrections into the function output (after padding the witness from depth $d$ to $D$), and prove that the resulting $\theta^\dagger$ still satisfies structured stability such that $\log H, \log L, \log C = \tilde O(d)$ (Lemma \ref{lemma:structured stability}). Notice that the data-dependence of $\theta^\dagger$ is not a problem, according to Lemma \ref{body lem:gen from structured stab}.  An intermediate notion that we call ``disturbance stability'' ---  defined in Appendix \ref{apdx subsec:existence of strucftured interpolators} --- relates to the DNN's ability to reject bounded perturbations applied to its computation at each layer, and makes proving structured stability more convenient.

\section{Conclusion}
We extended the techniques of \cite{pmlr-v235-buzaglo24a} to handle continuous weights and label noise, drawing connections between overparameterized generalization and the existence of narrow, \emph{stable} witnesses. The principal limitation of this work is that we analyze an unrealistic learning rule; however, analyzing uniform random interpolating parameters allows us to isolate those aspects of generalization that are unequivocally due to deep neural architectures and not optimizer bias. A key technical takeaway of this work is that approximation-theoretic guarantees can be converted into generalization guarantees, which may depend on finegrained stability properties of the approximator. 

\subsection*{AI use statement}

In this work, we used generative AI tools for assistance with information retrieval, proofs and exposition. We have not used generative AI for any other purposes. The proofs of Theorems \ref{body thm:continuous noiseless thm} and \ref{body thm:noisy informal} were first written by authors, but proofs were substantially improved with the help of AI. For example, the proof that a depth-$d$ witness, when padded to depth-$D$, has structured stability $H, L, C \approx e^d$ rather than $e^D$ was initially proven separately for (i) a generic witness (ii) a generic witness plus a memorization gadget, by means of an ugly induction argument. Generative AI suggested the introduction of disturbance stability as a stronger (Lemma \ref{lem:disturbance implies structured}) but more modular (Lemma \ref{lem:parallel disturbance stability}) type of stability, streamlining our proofs. AI was also used in order to speed up formalization; for example, the memorization gadget (Lemma \ref{lem:real valued memorization}) was formalized by AI at a request to build a narrow gadget using trapezoidal functions to localize points at each layer. We searched for prior work containing this memorization gadget, but found no exact matches, so we cite \cite{zhang2017understanding, bubeck2020network, Yun2018SmallRN}, which all share substantial DNA with the construction. At each stage, model output was verified by authors, and all textual content of the paper was manually written by humans. Figures \ref{fig:teaser} and \ref{fig:placeholder} were generated by AI according to author specifications and with extensive back-and-forth interactions. AI was used to retrieve related work, which was then read by authors. This includes the approximation results used in Examples 1 and 2, such as \cite{schmidt2020nonparametric,e2022barron, ohn2019smooth}. Finally, AI was used in gathering feedback on writing quality and polishing prose. We take responsibility for the final content of this work, including text, claims or artifacts produced with the aid of generative AI.

\bibliographystyle{plainnat}
\bibliography{citations}

@InProceedings{pmlr-v235-buzaglo24a,
  title = 	 {How Uniform Random Weights Induce Non-uniform Bias: Typical Interpolating Neural Networks Generalize with Narrow Teachers},
  author =       {Buzaglo, Gon and Harel, Itamar and Nacson, Mor Shpigel and Brutzkus, Alon and Srebro, Nathan and Soudry, Daniel},
  booktitle = 	 {Proceedings of the 41st International Conference on Machine Learning},
  year = 	 {2024}
}

@inproceedings{zhang2017understanding,
  title={Understanding deep learning requires rethinking generalization},
  author={Zhang, Chiyuan and Bengio, Samy and Hardt, Moritz and Recht, Benjamin and Vinyals, Oriol},
  booktitle={International Conference on Learning Representations},
  year={2017}
}

@inproceedings{theisen2021good,
  title     = {Good Classifiers are Abundant in the Interpolating Regime},
  author    = {Theisen, Ryan and Klusowski, Jason and Mahoney, Michael},
  booktitle = {Proceedings of the 24th International Conference on Artificial Intelligence and Statistics},
  year      = {2021}
}

@inproceedings{vallePerez2019deep,
  title     = {Deep Learning Generalizes Because the Parameter-Function Map is Biased Towards Simple Functions},
  author    = {Valle-P{\'e}rez, Guillermo and Camargo, Chico Q. and Louis, Ard A.},
  booktitle = {International Conference on Learning Representations},
  year      = {2019}
}

@inproceedings{dePalma2019random,
  title     = {Random Deep Neural Networks are Biased Towards Simple Functions},
  author    = {De Palma, Giacomo and Kiani, Bobak and Lloyd, Seth},
  booktitle = {Advances in Neural Information Processing Systems},
  year      = {2019}
}

@inproceedings{
chiang2023loss,
title={Loss Landscapes are All You Need: Neural Network Generalization Can Be Explained Without the Implicit Bias of Gradient Descent},
author={Chiang, Ping-yeh  and Ni,Renkun  and Miller, David Yu  and Bansal,Arpit  and  Geiping,Jonas and Goldblum,Micah  and  Goldstein,Tom},
booktitle={International Conference on Learning Representations },
year={2023}
}

@article{mingard2021sgd,
  title   = {Is {SGD} a Bayesian Sampler? Well, Almost},
  author  = {Mingard, Chris and Valle-P{\'e}rez, Guillermo and Skalse, Joar and Louis, Ard A.},
  journal = {Journal of Machine Learning Research},
  volume  = {22},
  year    = {2021}
}

@article{mingard2025occam,
  title   = {Deep Neural Networks Have an Inbuilt Occam's Razor},
  author  = {Mingard, Chris and Rees, Henry and Valle-P{\'e}rez, Guillermo and Louis, Ard A.},
  journal = {Nature Communications},
  volume  = {16},
  pages   = {220},
  year    = {2025}
}

@inproceedings{gunasekar2017implicit,
author = {Gunasekar, Suriya and Woodworth, Blake and Bhojanapalli, Srinadh and Neyshabur, Behnam and Srebro, Nathan},
title = {Implicit regularization in matrix factorization},
year = {2017},
booktitle = {Proceedings of the 31st International Conference on Neural Information Processing Systems}
}

@inproceedings{gunasekar2018implicit,
 author = {Gunasekar, Suriya and Lee, Jason D and Soudry, Daniel and Srebro, Nati},
 booktitle = {Advances in Neural Information Processing Systems},
 title = {Implicit Bias of Gradient Descent on Linear Convolutional Networks},
 year = {2018}
}

@inproceedings{arora2019implicit,
author = {Arora, Sanjeev and Cohen, Nadav and Hu, Wei and Luo, Yuping},
title = {Implicit regularization in deep matrix factorization},
year = {2019},
booktitle = {Proceedings of the 33rd International Conference on Neural Information Processing Systems}
}

@article{soudry2018implicit,
  title   = {The Implicit Bias of Gradient Descent on Separable Data},
  author  = {Soudry, Daniel and Hoffer, Elad and Nacson, Mor Shpigel and
             Gunasekar, Suriya and Srebro, Nathan},
  journal = {Journal of Machine Learning Research},
  number  = {70},
  year    = {2018}
}

@inproceedings{
lyu2020Gradient,
title={Gradient Descent Maximizes the Margin of Homogeneous Neural Networks},
author={Kaifeng Lyu and Jian Li},
booktitle={International Conference on Learning Representations},
year={2020}
}

@InProceedings{chizat2020implicit,
  title = 	 {Implicit Bias of Gradient Descent for Wide Two-layer Neural Networks Trained with the Logistic Loss},
  author =       {Chizat, L\'ena\"ic  and Bach, Francis},
  booktitle = 	 {Proceedings of Thirty Third Conference on Learning Theory},
  year = 	 {2020}
}

@inproceedings{mulayoff2020unique,
author = {Mulayoff, Rotem and Michaeli, Tomer},
title = {Unique properties of flat minima in deep networks},
year = {2020},
booktitle = {Proceedings of the 37th International Conference on Machine Learning}
}

@inproceedings{arora2022understanding,
  title     = {Understanding Gradient Descent on the Edge of Stability in Deep Learning},
  author    = {Arora, Sanjeev and Li, Zhiyuan and Panigrahi, Abhishek},
  booktitle = {Proceedings of the 39th International Conference on Machine Learning},
  year      = {2022}
}

@inproceedings{
madry2018towards,
title={Towards Deep Learning Models Resistant to Adversarial Attacks},
author={Aleksander Madry and Aleksandar Makelov and Ludwig Schmidt and Dimitris Tsipras and Adrian Vladu},
booktitle={International Conference on Learning Representations},
year={2018}
}

@article{kohler2013optimal,
  title   = {Optimal Global Rates of Convergence for Interpolation Problems with Random Design},
  author  = {Kohler, Michael and Krzy{\.z}ak, Adam},
  journal = {Statistics \& Probability Letters},
  volume  = {83},
  number  = {8},
  pages   = {1871--1879},
  year    = {2013},
  doi     = {10.1016/j.spl.2013.04.018}
}

@article{yangzhou2025optimal,
  title   = {Optimal Rates of Approximation by Shallow {ReLU}$^k$ Neural Networks and Applications to Nonparametric Regression},
  author  = {Yang, Yunfei and Zhou, Ding-Xuan},
  journal = {Constructive Approximation},
  volume  = {62},
  pages   = {329--360},
  year    = {2025},
  doi     = {10.1007/s00365-024-09679-z}
}

@article{e2022barron,
  title   = {The Barron Space and the Flow-Induced Function Spaces for Neural Network Models},
  author  = {E, Weinan and Ma, Chao and Wu, Lei},
  journal = {Constructive Approximation},
  volume  = {55},
  number  = {1},
  pages   = {369--406},
  year    = {2022},
  doi     = {10.1007/s00365-021-09549-y}
}

@misc{voigtlaender2022sampling,
  title         = {{$L^p$} Sampling Numbers for the {Fourier}-Analytic {Barron} Space},
  author        = {Voigtlaender, Felix},
  year          = {2022},
  howpublished  = {arXiv:2208.07605},
  url           = {https://arxiv.org/abs/2208.07605}
}

@inproceedings{peleg2024bias,
  title     = {Bias of Stochastic Gradient Descent or the Architecture:
               Disentangling the Effects of Overparameterization of Neural Networks},
  author    = {Peleg, Amit and Hein, Matthias},
  booktitle = {Proceedings of the 41st International Conference on Machine Learning},
  volume    = {235},
  pages     = {40154--40184},
  year      = {2024}
}

@inproceedings{berchenko2024simplicity,
  title     = {Simplicity Bias in Overparameterized Machine Learning},
  author    = {Berchenko, Yakir},
  booktitle = {Proceedings of the AAAI Conference on Artificial Intelligence},
  year      = {2024}
}

@inproceedings{welling2011bayesian,
author = {Welling, Max and Teh, Yee Whye},
title = {Bayesian learning via stochastic gradient langevin dynamics},
year = {2011},
booktitle = {Proceedings of the 28th International Conference on International Conference on Machine Learning},
series = {ICML'11}
}

@article{chen2022stationary,
author = {Chen, Zaiwei and Mou, Shancong and Maguluri, Siva Theja},
title = {Stationary Behavior of Constant Stepsize SGD Type Algorithms: An Asymptotic Characterization},
year = {2022},
issue_date = {March 2022},
publisher = {Association for Computing Machinery},
volume = {6},
number = {1},
journal = {Proc. ACM Meas. Anal. Comput. Syst.}
}

@inproceedings{harel2024provable,
  title     = {Provable Tempered Overfitting of Minimal Nets and Typical Nets},
  author    = {Harel, Itamar and Hoza, William M. and Vardi, Gal and Evron, Itay and Srebro, Nathan and Soudry, Daniel},
  booktitle = {Advances in Neural Information Processing Systems},
  year      = {2024}
}

@inproceedings{alexander2025neural,
  title     = {Do Neural Networks Need Gradient Descent to Generalize? A Theoretical Study},
  author    = {Alexander, Yotam and Slutzky, Yonatan and Ran-Milo, Yuval and Cohen, Nadav},
  booktitle = {Advances in Neural Information Processing Systems},
  year      = {2025}
}

@inproceedings{liang2025stable,
  title     = {Stable Minima of {ReLU} Neural Networks Suffer from the Curse of Dimensionality: The Neural Shattering Phenomenon},
  author    = {Liang, Tongtong and Qiao, Dan and Wang, Yu-Xiang and Parhi, Rahul},
  booktitle = {Advances in Neural Information Processing Systems},
  year      = {2025}
}

@misc{ohn2021nonconvexsparseregularizationdeep,
      title={Nonconvex sparse regularization for deep neural networks and its optimality}, 
      author={Ilsang Ohn and Yongdai Kim},
      year={2021},
      eprint={2003.11769},
      archivePrefix={arXiv},
      primaryClass={math.ST}
}

@inproceedings{liang2026generalization,
  title     = {Generalization Below the Edge of Stability: The Role of Data Geometry},
  author    = {Liang, Tongtong and Cloninger, Alexander and Parhi, Rahul and Wang, Yu-Xiang},
  booktitle = {International Conference on Learning Representations},
  year      = {2026}
}

@inproceedings{qiao2024stable,
  title     = {Stable Minima Cannot Overfit in Univariate {ReLU} Networks: Generalization by Large Step Sizes},
  author    = {Qiao, Dan and Zhang, Kaiqi and Singh, Esha and Soudry, Daniel and Wang, Yu-Xiang},
  booktitle = {Advances in Neural Information Processing Systems},
  year      = {2024}
}

@article{alquier2024user,
  title   = {User-Friendly Introduction to {PAC-Bayes} Bounds},
  author  = {Alquier, Pierre},
  journal = {Foundations and Trends in Machine Learning},
  volume  = {17},
  number  = {2},
  pages   = {174--303},
  year    = {2024}
}

@inproceedings{clerico2025generalisation,
  title     = {Generalisation under Gradient Descent via Deterministic {PAC-Bayes}},
  author    = {Clerico, Eugenio and Farghly, Tyler and Deligiannidis, George
               and Guedj, Benjamin and Doucet, Arnaud},
  booktitle = {Proceedings of the 36th International Conference on Algorithmic Learning Theory},
  series    = {Proceedings of Machine Learning Research},
  volume    = {272},
  pages     = {349--389},
  year      = {2025},
  publisher = {PMLR}
}

@Article{viallard2024general,
author={Viallard, Paul
and Germain, Pascal
and Habrard, Amaury
and Morvant, Emilie},
title={A general framework for the practical disintegration of PAC-Bayesian bounds},
journal={Machine Learning},
year={2024},
month={Feb},
day={01},
volume={113},
number={2},
pages={519-604},
issn={1573-0565},
doi={10.1007/s10994-023-06391-0}
}

@article{bartlett2020benign,
author = {Peter L. Bartlett  and Philip M. Long  and Gábor Lugosi  and Alexander Tsigler },
title = {Benign overfitting in linear regression},
journal = {Proceedings of the National Academy of Sciences},
volume = {117},
number = {48},
pages = {30063-30070},
year = {2020},
}

@inproceedings{dziugaite2025size,
  title     = {The Size of Teachers as a Measure of Data Complexity: {PAC-Bayes} Excess Risk Bounds and Scaling Laws},
  author    = {Dziugaite, Gintare Karolina and Roy, Daniel M.},
  booktitle = {Proceedings of the 28th International Conference on Artificial Intelligence and Statistics},
  year      = {2025}
}

@inproceedings{lotfi2024nonvacuous,
  title     = {Non-Vacuous Generalization Bounds for Large Language Models},
  author    = {Lotfi, Sanae and Finzi, Marc and Kuang, Yilun and Rudner, Tim G. J. and Goldblum, Micah and Wilson, Andrew Gordon},
  booktitle = {Proceedings of the 41st International Conference on Machine Learning},
  year      = {2024}
}

@inproceedings{wilson2025position,
  title     = {Position: Deep Learning is Not So Mysterious or Different},
  author    = {Wilson, Andrew Gordon},
  booktitle = {Proceedings of the 42nd International Conference on Machine Learning},
  year      = {2025}
}

@inproceedings{dinh2017sharp,
  title     = {Sharp Minima Can Generalize For Deep Nets},
  author    = {Dinh, Laurent and Pascanu, Razvan and Bengio, Samy and Bengio, Yoshua},
  booktitle = {Proceedings of the 34th International Conference on Machine Learning},
  year      = {2017}
}

@inproceedings{li2018visualizing,
  title     = {Visualizing the Loss Landscape of Neural Nets},
  author    = {Li, Hao and Xu, Zheng and Taylor, Gavin and Studer, Christoph and Goldstein, Tom},
  booktitle = {Advances in Neural Information Processing Systems},
  year      = {2018}
}

@inproceedings{garipov2018loss,
  title     = {Loss Surfaces, Mode Connectivity, and Fast Ensembling of {DNNs}},
  author    = {Garipov, Timur and Izmailov, Pavel and Podoprikhin, Dmitrii and Vetrov, Dmitry and Wilson, Andrew Gordon},
  booktitle = {Advances in Neural Information Processing Systems},
  year      = {2018}
}

@inproceedings{
foret2021sharpnessaware,
title={Sharpness-aware Minimization for Efficiently Improving Generalization},
author={Pierre Foret and Ariel Kleiner and Hossein Mobahi and Behnam Neyshabur},
booktitle={International Conference on Learning Representations},
year={2021}
}

@inproceedings{power2021grokking,
  title     = {Grokking: Generalization Beyond Overfitting on Small Algorithmic Datasets},
  author    = {Power, Alethea and Burda, Yuri and Edwards, Harri and Babuschkin, Igor and Misra, Vedant},
  booktitle = {ICLR MATH-AI Workshop},
  year      = {2021}
}

@inproceedings{mcallester1999pac,
  title     = {{PAC-Bayesian} Model Averaging},
  author    = {McAllester, David A.},
  booktitle = {Proceedings of the Twelfth Annual Conference on Computational Learning Theory},
  year      = {1999}
}

@inproceedings{zhou2019nonvacuous,
  title     = {Non-Vacuous Generalization Bounds at the {ImageNet} Scale: A {PAC-Bayesian} Compression Approach},
  author    = {Zhou, Wenda and Veitch, Victor and Austern, Morgane and Adams, Ryan P. and Orbanz, Peter},
  booktitle = {International Conference on Learning Representations},
  year      = {2019}
}

@inproceedings{lotfi2022pacbayes,
  title     = {{PAC-Bayes} Compression Bounds So Tight That They Can Explain Generalization},
  author    = {Lotfi, Sanae and Finzi, Marc and Kapoor, Sanyam and Potapczynski, Andres and Goldblum, Micah and Wilson, Andrew Gordon},
  booktitle = {Advances in Neural Information Processing Systems},
  year      = {2022}
}

@inproceedings{mallinar2022benign,
  title     = {Benign, Tempered, or Catastrophic: Toward a Refined Taxonomy of Overfitting},
  author    = {Mallinar, Neil and Simon, James and Abedsoltan, Amirhesam and Pandit, Parthe and Belkin, Misha and Nakkiran, Preetum},
  booktitle = {Advances in Neural Information Processing Systems},
  year      = {2022}
}

@inproceedings{parhi2025upper,
  author    = {Parhi, Rahul and Adcock, Ben},
  title     = {Upper Bounds on Averaged Sampling Numbers for General Model Classes},
  booktitle = {2025 International Conference on Sampling Theory and Applications (SampTA)},
  year      = {2025},
  pages     = {1--5},
  doi       = {10.1109/SampTA64769.2025.11133520},
  url       = {https://sparsity.ucsd.edu/publications/parhi2025upper.pdf}
}

@inproceedings{VardiYS22,
  author    = {Gal Vardi and
               Gilad Yehudai and
               Ohad Shamir},
  title     = {Width is Less Important than Depth in ReLU Neural Networks},
  booktitle = {Proceedings of Thirty Fifth Annual Conference on Learning Theory},
  year      = {2022},
}

@article{schmidt2020nonparametric,
  author  = {Schmidt-Hieber, Johannes},
  title   = {Nonparametric Regression Using Deep Neural Networks with {ReLU} Activation Function},
  journal = {The Annals of Statistics},
  volume  = {48},
  number  = {4},
  year    = {2020},
  doi     = {10.1214/19-AOS1875}
}

@Article{ohn2019smooth,
AUTHOR = {Ohn, Ilsang and Kim, Yongdai},
TITLE = {Smooth Function Approximation by Deep Neural Networks with General Activation Functions},
JOURNAL = {Entropy},
VOLUME = {21},
YEAR = {2019},
NUMBER = {7},
ARTICLE-NUMBER = {627}
}

@inproceedings{bubeck2020network,
 author = {Bubeck, Sebastien and Eldan, Ronen and Lee, Yin Tat and Mikulincer, Dan},
 booktitle = {Advances in Neural Information Processing Systems},
 title = {Network size and size of the weights in memorization with two-layers neural networks},
 year = {2020}
}

@inproceedings{Yun2018SmallRN,
  title={Small ReLU networks are powerful memorizers: a tight analysis of memorization capacity},
  author={Chulhee Yun and Suvrit Sra and Ali Jadbabaie},
  booktitle={Neural Information Processing Systems},
  year={2019}
}

@article{klusowski2018approximation,
  author    = {Klusowski, Jason M. and Barron, Andrew R.},
  title     = {Approximation by Combinations of {ReLU} and Squared {ReLU} Ridge Functions with $\ell^1$ and $\ell^0$ Controls},
  journal   = {IEEE Transactions on Information Theory},
  volume    = {64},
  number    = {12},
  pages     = {7649--7656},
  year      = {2018},
  publisher = {IEEE}
}

@incollection{Bengio+chapter2007,
author = {Bengio, Yoshua and LeCun, Yann},
booktitle = {Large Scale Kernel Machines},
publisher = {MIT Press},
title = {Scaling Learning Algorithms Towards {AI}},
year = {2007}
}

@article{Hinton06,
author = {Hinton, Geoffrey E. and Osindero, Simon and Teh, Yee Whye},
journal = {Neural Computation},
pages = {1527--1554},
title = {A Fast Learning Algorithm for Deep Belief Nets},
volume = {18},
year = {2006}
}

@book{goodfellow2016deep,
title={Deep learning},
author={Goodfellow, Ian and Bengio, Yoshua and Courville, Aaron and Bengio, Yoshua},
volume={1},
year={2016},
publisher={MIT Press}
}
\nocite{*}

\appendix
\section{Extended related work}\label{apdx:related work}
Substantial effort has gone into investigating \textbf{architecture-induced} versus \textbf{optimizer-induced bias} \citep{pmlr-v235-buzaglo24a, qiao2024stable, alexander2025neural, liang2026generalization}. We focus on related work in the former category, which are most relevant to this paper. 

\noindent\textbf{Simplicity Bias} \cite{vallePerez2019deep, depalma2019random} provide early evidence showing that uniform distributions over parameters push forward onto highly non-uniform distributions over implemented functions; related empirical work suggests that this architecture-induced function prior captures a significant component of the inductive bias observed under SGD \citep{mingard2021sgd, mingard2025occam}. \cite{berchenko2024simplicity} develops this perspective theoretically, by showing that random constructions can exhibit simplicity bias  in overparameterized models. Unlike most of these works, we specifically consider uniformity \emph{conditioned on interpolation}, which helps us derive rigorous generalization guarantees from architectural properties. At the same time, recent work has emphasized limits of a purely architectural explanation:  \cite{peleg2024bias} find settings in which increasing width benefits SGD but not random interpolators, while \cite{alexander2025neural} show that gradient methods can outperform ``dumb sampling'' in matrix factorization problems.

\noindent\textbf{Typical interpolating networks} \cite{pmlr-v235-buzaglo24a} shows that randomly sampled DNNs also generalize in the case of noiseless data from a realizable ground truth. \cite{harel2024provable} shows that this analysis can be extended to noisy binary classification, but considers only DNNs with binary weights. \cite{dziugaite2025size} reformulates the proof technique of \cite{pmlr-v235-buzaglo24a}, and considers DNNs drawn from loss-based Boltzmann distributions rather than only interpolating DNNs. To our knowledge, prior work does not cover deep networks with continuous weights, or label noise in this setting.

\noindent\textbf{PAC-Bayes} PAC-Bayes bounds analyze the risk of a hypothesis $h \sim Q_n$ sampled from a data-dependent distribution. Then, for any data-independent measure $P$, the worst-case complexity term in standard generalization bounds can be replaced by $\mathrm{KL}(Q_n || P)$, to get generalization bounds \citep{mcallester1999pac, alquier2024user, clerico2025generalisation}. Although our setup can be formalized within the PAC-Bayes framework,  we do not use PAC-Bayes results, and instead perform a direct analysis of the risk. Our primary technical contributions are orthogonal to PAC-Bayes theory, and relate specifically to generalization-promoting aspects of deep ReLU architectures.  In practice, it has been observed that using PAC-Bayes bounds with priors $P$ that favor \textbf{compressibility} leads to strong generalization bounds even at modern scales \citep{zhou2019nonvacuous, lotfi2022pacbayes, lotfi2024nonvacuous}, although these are numerical certificates and not analytical rates.  The ``deep lifting'' procedure shows that large volumes of interpolating parameters ``behave like'' parameters for smaller DNNs (i.e. are compressible), possibly shedding light on the empirical success of these approaches.

\section{Definition of $U^{W, D, M}_{\tau, n}$}\label{apdx sec:uniform dist def}
We define the uniform distribution over $S^{W,D,M}_{\tau,n}$, from which our estimate $\hat\theta$ will be drawn, as in \eqref{body eq:interpolation definitions}. Since $\mathcal C_M(W,D)\subset\mathbb R^{N_{W,D}}$, we inherit Lebesgue measure from this ambient parameter space. For any measurable parameter set $A$, when $S^{W,D,M}_{\tau,n}$ has positive volume, define
\[
U_{\tau,n}^{W,D,M}(A)
:=\operatorname{Unif}(S_{\tau,n}^{W,D,M})(A)
:=\frac{\lambda_{N_{W,D}}(A\cap S_{\tau,n}^{W,D,M})}
{\lambda_{N_{W,D}}(S_{\tau,n}^{W,D,M})}.
\]
When $S^{W,D,M}_{\tau,n}$ has zero volume, including when it is empty, define $U^{W,D,M}_{\tau,n}$ to be the point mass at the zero parameter vector.

\section{Memorization gadget}

This section introduces the memorization gadget that memorizes $m$ corruptions $\delta_1 \dots \delta_m \in \R^S$. Though there is prior work on memorization construction within DNNs \citep{zhang2017understanding, bubeck2020network, Yun2018SmallRN}, the following construction suits our particular needs well. 

\begin{lemma}[Real-valued memorization gadget]\label{lem:real valued memorization}
Let $x_1,\dots,x_n\in \R^p$, and suppose that the corrupted points are indexed by
$[m]$. Assume $m\in\{0,\ldots,n\}$, integer $D\geq1$, and $|x_i[1]|\leq X$ for $i\in[n]$. Let the correction vectors satisfy $\|\delta_i\|_\infty\leq\Gamma$ for $i\in[m]$, where $X>0$ and $\Gamma\geq0$. Fix
$\ell>0$, and for each $i\in[m]$ choose $a_i\in \R$ such that
\[
x_i[1]\in I_i:=[a_i,a_i+\ell].
\]
Define the enlarged interval
\[
\tilde I_i:=[a_i-\ell,a_i+2\ell].
\]
Let $B:=\left\lceil \frac mD\right\rceil$.
Then there exists a width $w_{\mathrm{mem}}:=2+2s+4B$ and depth-$D$ ReLU network $\theta_{\mathrm{mem}}\in \mathcal A(w_{\mathrm{mem}},D)$ such that
\[
g^{w_{\mathrm{mem}},D}_{\theta_{\mathrm{mem}}}(x)
=
0
\]
whenever $x[1]\notin \bigcup_{i=1}^m \tilde I_i$.

Moreover, if 
$x_j[1]\notin \tilde I_i$
$\text{for every } i\in[m] \text{ and every } j\neq i$, then
\[
g^{w_{\mathrm{mem}},D}_{\theta_{\mathrm{mem}}}(x_i)=\delta_i
\qquad \text{for } i\in[m],
\]
and
\[
g^{w_{\mathrm{mem}},D}_{\theta_{\mathrm{mem}}}(x_j)=0
\qquad \text{for } j\in\{m+1,\dots,n\}.
\]

Moreover, $\theta_\mathrm{mem}$ can be taken to lie within $\mathcal C_{M}(w_\mathrm{mem}, D)$ provided that
\[M \geq M_{X, \Gamma, \ell} := \max \{1, X + 2\ell, \frac{\Gamma}{\ell}\}.\]
\end{lemma}

\begin{proof}
The points $x_1,\dots,x_m$ are the corrupted data points. For each such point,
$\delta_i\in\R^s$ is the correction term that we wish to add at $x_i$.

For each $i\in[m]$, define the hinge points
$\tau_i^1=a_i-\ell$, $\tau_i^2=a_i$, $\tau_i^3=a_i+\ell$, and
$\tau_i^4=a_i+2\ell$. Define
\[
\psi_i(z)
=
\sigma\left({z-\tau_i^1}\right)
-
\sigma\left({z-\tau_i^2}\right)
-
\sigma\left({z-\tau_i^3}\right)
+
\sigma\left({z-\tau_i^4}\right).
\]

Thus $\psi_i$ is a trapezoidal function that takes value $\ell$ on
$I_i=[a_i,a_i+\ell]$ and is zero outside
$\tilde I_i=[a_i-\ell,a_i+2\ell]$.

Divide the $m$ corrupted points evenly among layers, into groups
$C_1,\dots,C_D$ of size at most $B=\lceil m/D\rceil$. We identify each $C_k$
with a subset of $[m]$, but abuse notation by saying ``$x_i\in C_k$'' whenever
$i\in C_k$.

Let $w_{\mathrm{mem}}=2+2s+4B$. We now construct $\theta_\mathrm{mem}$ by describing how it evolves its internal activations with affine operations. 

For internal activations in
$\R^{w_{\mathrm{mem}}}$, we divide coordinates into channels
$[Z^+,Z^-,P,N,S]$, where $Z^+,Z^-\in\R$, $P,N\in\R^s$, and
$S\in\R^{4B}$. The two coordinates $Z^+$ and $Z^-$ will carry the first input
coordinate as $Z^+=\sigma(x[1])$ and $Z^-=\sigma(-x[1])$, so
$Z^+-Z^-=x[1]$. The channels $P$ and $N$ will carry positive and negative
``corrections''. Write $\delta_i=\delta_i^+-\delta_i^-$, where
$\delta_i^+,\delta_i^-\in\R^s_{\geq 0}$ are the coordinatewise positive and
negative parts of $\delta_i$. The channel will $S$ carry hinge points for groups of corrupted points. .

For a group $C_k$, let $\tau_{C_k}\in\R^{4B}$ denote the vector of hinge points
associated to the points in $C_k$, padded with zero hinge locations and zero correction coefficients if $|C_k|<B$. We maintain
by induction that, after layer $k$, the activation
$[Z_k^+,Z_k^-,P_k,N_k,S_k]$ satisfies $Z_k^+-Z_k^-=x[1]$,
\[
P_k[j]=\sum_{u=1}^{k-1}\sum_{i\in C_u}\frac{\delta_i^+}{\ell}[j]\psi_i(x[1]),
\qquad
N_k[j]=\sum_{u=1}^{k-1}\sum_{i\in C_u}\frac{\delta_i^-[j]}{\ell}\psi_i(x[1]),
\]
and $S_k$ contains the hinge features
$\sigma((x[1]-\tau_i^r))$ for every $i\in C_k$ and $r\in[4]$. We interpret the empty sum at $k = 1$ as $0$. 

\textbf{Base case.}
At layer $1$, set $Z_1^+=\sigma(x[1])$, $Z_1^-=\sigma(-x[1])$, and
$P_1=N_1=0$. Also set
\[
\tilde S_1=\mathbf 1_{4B}x[1]-\tau_{C_1},
\qquad
S_1=\sigma(\tilde S_1),
\]
where $\mathbf{1}_{4B} \in  \{1\}^{4B}$ is the vector of $1$s. Thus $S_1$ contains the hinge features needed to form $\psi_i$ for
every $i\in C_1$, and the induction hypotheses hold at layer $1$.

\textbf{Induction.}
Now let $k\geq 2$ and assume the induction hypotheses hold at layer $k-1$.
Copy the first-coordinate channels by setting $\tilde Z_k^+=Z_{k-1}^+$
and $\tilde Z_k^-=Z_{k-1}^-$. Since these coordinates are nonnegative,
ReLU leaves them unchanged, so $Z_k^+-Z_k^-=x[1]$.

For each $i\in C_{k-1}$, let
\[
\Psi_i(S_{k-1})
:=
S_{k-1}^{i,1}-S_{k-1}^{i,2}-S_{k-1}^{i,3}+S_{k-1}^{i,4}.
\]
By induction and the definition of the hinge features,
$\Psi_i(S_{k-1})=\psi_i(x[1])$. For every $j\in[s]$, set
\[
\tilde P_k[j]
=
P_{k-1}[j]
+
\sum_{i\in C_{k-1}}\frac{\delta_i^+[j]}{\ell}\Psi_i(S_{k-1}),
\]
and
\[
\tilde N_k[j]
=
N_{k-1}[j]
+
\sum_{i\in C_{k-1}}\frac{\delta_i^-[j]}{\ell}\Psi_i(S_{k-1}).
\]
Equivalently, $\tilde P_k[j]=P_{k-1}[j]+\sum_{i\in C_{k-1}}
\frac{\delta_i^+[j]}{\ell}\psi_i(x[1])$, and similarly for $n_k$. These quantities are
nonnegative, so $P_k=\tilde P_k$ and $N_k=\tilde N_k$ after ReLU.

Finally, prepare the hinge features for the next group by setting
\[
\tilde S_k
=
\mathbf 1_{4B}(Z_{k-1}^+-Z_{k-1}^-)-\tau_{C_k},
\qquad
S_k=\sigma(\tilde S_k).
\]
Since $Z_{k-1}^+-Z_{k-1}^-=x[1]$, the channel $S_k$ contains the normalized
hinge features needed to form $\psi_i$ for every $i\in C_k$. This completes
the induction.

\textbf{Readout.}
At layer $D$, the channels $P_D$ and $N_D$ contain the accumulated corrections
from $C_1,\dots,C_{D-1}$, while $S_D$ contains the hinge features
for $C_D$. Define the readout coordinatewise by
\[
g^{w_{\mathrm{mem}},D}_{\theta_{\mathrm{mem}}}(x)[j]
=
P_D[j]-N_D[j]
+
\sum_{i\in C_D}
\left(
\frac{\delta_i^+[j]}{\ell}-\frac{\delta_i^-[j]}{\ell}
\right)
\Psi_i(S_D).
\]
Since $\Psi_i(S_D)=\psi_i(x[1])$ for every $i\in C_D$, this gives
\[
g^{w_{\mathrm{mem}},D}_{\theta_{\mathrm{mem}}}(x)
=
\sum_{i=1}^m \delta_i\frac{1}{\ell}\psi_i(x[1]).
\]

Now if $x[1]\notin\bigcup_{i=1}^m\tilde I_i$, then every $\psi_i(x[1])$
vanishes, so $g^{w_{\mathrm{mem}},D}_{\theta_{\mathrm{mem}}}(x)=0$.

Finally assume $x_j[1]\notin\tilde I_i$ for every $i\in[m]$ and every
$j\neq i$. For a corrupted point $x_i$, we have $x_i[1]\in I_i$, so
$\psi_i(x_i[1])=\ell$, while $\psi_r(x_i[1])=0$ for every $r\neq i$. Hence
$g^{w_{\mathrm{mem}},D}_{\theta_{\mathrm{mem}}}(x_i)=\delta_i$. If
$j\in\{m+1,\dots,n\}$, then $x_j[1]\notin\tilde I_i$ for every $i\in[m]$,
so every $\psi_i(x_j[1])$ vanishes and
$g^{w_{\mathrm{mem}},D}_{\theta_{\mathrm{mem}}}(x_j)=0$.

Since $a_i\in[x_i[1]-\ell,x_i[1]]$, all four hinge locations have magnitude at most $X+2\ell$. The other nonzero weights have magnitudes $1$ or at most $\Gamma/\ell$; unused slots have zero coefficients. Thus all parameters are bounded by $M_{X,\Gamma,\ell}$.
\end{proof}

We now use the $\Lambda$-marginal boundedness assumption (Assumption \ref{body ass:lambda marginal boundedness}) to show that the memorization gadget succeeds with high probability over the draw $x_1 \dots x_n$. 

\begin{lemma}\label{lem:probabilistic memorization}
Assume that $x_1,\dots,x_n\sim \mu$ iid, and that $\mu$ is
$\Lambda$-marginally bounded in its first coordinate. Assume $n\geq2$, $\Lambda>0$, $|x_i[1]|\leq X$ almost surely, $\Gamma\geq0$, and integer $D\geq1$. Fix a deterministic budget $m\in\{0,\ldots,n\}$, $\eta\in(0,1)$, and choose
\[
0<\ell \leq \frac{\eta}{3\Lambda n(n-1)}, \qquad  M
\geq
M_{X,\Gamma,\ell}
:=
\max\left\{
1,\,
X+2\ell,\,
\frac{\Gamma}{\ell}
\right\}.
\]
Define $w_{\mathrm{mem}}:=2+2s+4\left\lceil\frac{m}{D}\right\rceil$.

Then, with probability at least $1-\eta$ over $x_1,\dots,x_n$, the following holds simultaneously for every subset $I\subseteq[n]$ with $|I|\leq m$ and every collection of correction vectors
\[
\{\delta_i:i\in I\},
\qquad
\|\delta_i\|_\infty\leq\Gamma.
\]
There exists $\theta_{\mathrm{mem}}\in\mathcal C_M(w_{\mathrm{mem}},D)$ such that
\[
g_{\theta_{\mathrm{mem}}}^{w_{\mathrm{mem}},D}(x_i)
=
\delta_i
\qquad
\text{for every }i\in I,
\]
and
\[
g_{\theta_{\mathrm{mem}}}^{w_{\mathrm{mem}},D}(x_j)
=
0
\qquad
\text{for every }j\notin I.
\]
\end{lemma}

\begin{proof}
For every $i\in[n]$, set
\[
a_i:=x_i[1],
\qquad
\tilde I_i:=[a_i-\ell,a_i+2\ell].
\]
Let $\mathcal B$ be the event that there exist $i\neq j$ such that
$x_j[1]\in\tilde I_i$. Conditional on $x_i$, the interval
$\tilde I_i$ has length $3\ell$, so marginal boundedness gives
\[
\Pr[x_j[1]\in \tilde I_i\mid x_i]\leq 3\Lambda \ell.
\]
Taking a union bound over the $n(n-1)$ ordered pairs gives
\[
\Pr[\mathcal B]\leq 3\Lambda\ell n(n-1)\leq \eta.
\]
On $\mathcal B^c$, fix any $I\subseteq[n]$ with $|I|\leq m$ and any collection of correction vectors satisfying the stated bound. Enumerate the points indexed by $I$ and apply Lemma \ref{lem:real valued memorization}. The localization condition holds for every selected point, and the required width is at most
\[
2+2s+4\left\lceil\frac{m}{D}\right\rceil.
\]
Unused memory slots are assigned zero correction coefficients. The resulting parameter lies in $\mathcal C_M(w_{\mathrm{mem}},D)$ by the assumed lower bound on $M$, and it has the claimed values on all sampled points. Since $\mathcal B^c$ does not depend on $I$ or on the correction vectors, the conclusion holds simultaneously for every such choice.
\end{proof}

\begin{lemma}\label{lem:corrupted point bound}
Let $\eta\in(0,1)$, $\epsilon\in[0,1]$, and $m\sim\mathrm{Binom}(n,\epsilon)$. Then
\[
m\leq 2\epsilon n+\log\frac1\eta
\]
with probability at least $1-\eta$.
\end{lemma}
\begin{proof}
Follows from the Chernoff bound.
\end{proof}

For $n\geq2$, $\Lambda>0$, and $\eta\in(0,1)$, we define $M_{X, \Gamma, \Lambda, n, \eta} := M_{X, \Gamma, \ell}$ with $\ell := \frac{\eta}{3\Lambda n(n - 1)}$. Then we have the following corollary of Lemma \ref{lem:probabilistic memorization}.


\begin{lemma}[Interpolation with memorization]
\label{lem:interpolation via memorization}
Fix $\eta\in(0,1)$. Assume $n\geq2$, $\Lambda>0$, $\Gamma\geq0$, and integer $D\geq1$. Assume $x_1,\dots,x_n\sim\mu_X$ iid, where $\mu_X$ is $\Lambda$-marginally bounded in its first coordinate and satisfies $x[1]\in[-X,X]$ almost surely. Fix a deterministic budget $m\in\{0,\ldots,n\}$, and let $I=I(x^n)\subseteq[n]$ be any subset with $|I|\leq m$. Let $\{\delta_i:i\in I\}$ be any, possibly data-dependent, collection of correction vectors satisfying $\|\delta_i\|_\infty\leq\Gamma$.

Let $M \geq M_{X, \Gamma, \Lambda, n, \eta}$ and let $\theta \in \mathcal C_M(w, D)$. Then, with probability at least $1-\eta$ over the draw $x_1 \dots x_n \sim \mu_X^{\otimes n}$, simultaneously for every such $I$ and collection of correction vectors, there exists $\theta^\dagger \in \mathcal C_M(w', D)$ such that
\[
g_{\theta^\dagger}^{w',D}(x_i)
=
g_\theta^{w,D}(x_i)+\delta_i
\qquad
\text{for every }i\in I,
\]
and
\[
g_{\theta^\dagger}^{w',D}(x_j)
=
g_\theta^{w,D}(x_j)
\qquad
\text{for every }j\notin I,
\]
where $w' := w + w_\mathrm{mem}$, according to the definition of $w_\mathrm{mem}$ in Lemma \ref{lem:probabilistic memorization}.
\end{lemma}

\begin{proof}
On the good event of Lemma \ref{lem:probabilistic memorization}, there exists $\theta_{\mathrm{mem}}\in\mathcal C_M(w_{\mathrm{mem}},D)$ such that $g_{\theta_{\mathrm{mem}}}^{w_{\mathrm{mem}},D}(x_i)=\delta_i$ for $i\in I$, and $g_{\theta_{\mathrm{mem}}}^{w_{\mathrm{mem}},D}(x_j)=0$ for $j\notin I$. Run $\theta$ and $\theta_{\mathrm{mem}}$ in parallel and add their outputs at the readout. This gives the claimed $\theta^\dagger\in\mathcal C_M(w',D)$.
\end{proof}

By Lemma \ref{lem:interpolation via memorization}, given $(n, m, \Lambda, \Gamma, \eta)$, there exists a memorization gadget $\theta_\mathrm{mem}$ with width $w_\mathrm{mem}$, that, with probability at least $1 - \eta$, fits the residuals for the $m$-of-$n$ corrupted data points. We will refer to this as the $(n, m, \Lambda, \Gamma, \eta)$-memorization gadget. We will refer to the induced interpolating parameter $\theta^\dagger$ from Lemma \ref{lem:interpolation via memorization} as the interpolating parameter induced by the $(n, m, \Lambda, \Gamma, \eta)$-memorization gadget.

\section{Structured Interpolators and Continuous Lifts}

\subsection{Structured Stability}

\begin{definition}[Structured stability]\label{def:structured stability}
Let $\theta\in\mathcal C_M(u,D)$. We say that $\theta$ has $(H,L,C)$ structured stability, with $H,L,C>0$, if it satisfies the following conditions.

(i) $\theta$'s hidden activations are bounded by $H$ on $[-X,X]^p$: $\|x_k^\theta(x)\|_\infty \leq H$ for every $x\in[-X,X]^p$ and $k\in[D]$, 

(ii) If
\[
\tilde\theta\in\mathcal C_M(u,D),
\qquad
\|\tilde\theta-\theta\|_\infty\leq\rho,
\qquad
\rho\leq\frac1C,
\]
then 
\[\|x_k^{\tilde \theta}\|_\infty \leq H + 1 \qquad \text{and} \qquad 
\sup_{x\in[-X,X]^p}
\left\|
 g_{\tilde\theta}^{u,D}(x)-g_\theta^{u,D}(x)
\right\|_\infty
\leq L\rho.
\]
\end{definition}

\subsection{Lifting a Structured Parameter}
Throughout this subsection assume $M\geq1$ and integer widths and depth $1\leq u<W$, $D\geq1$.

Let $\theta\in\mathcal C_M(u,D)$ have matrices $\{W_k^\theta\}$, readout $R^\theta$, and biases $\{b_k^\theta\}$. Let $\phi\in\mathcal C_M(W,D)$ have matrices $\{W_k^\phi\}$, readout $R^\phi$, and biases $\{b_k^\phi\}$. For $a\in[-M,M]$ and $0<\rho\leq1$, define the inward interval
\[
I_\rho(a)
:=
\begin{cases}
[a,a+\rho],&a\leq0,\\
[a-\rho,a],&a>0.
\end{cases}
\]
For matrices and vectors, $I_\rho(A)$ denotes the Cartesian product of the corresponding intervals.

Write the parameters of $\phi$ in block form
\[
W_1^\phi=
\begin{pmatrix}
A_1\\ C_1
\end{pmatrix},
\qquad
W_k^\phi=
\begin{pmatrix}
A_k&B_k\\ C_k&D_k
\end{pmatrix}
\quad(k\geq2),
\]
\[
b_k^\phi=
\begin{pmatrix}
a_k\\ c_k
\end{pmatrix},
\qquad
R^\phi=(Q_1\;Q_2),
\]
where the first block has width $u$.

In the context of this section, we call $A_1 \dots A_D, a_1 \dots a_D, Q_1$ ``carried'' parameters, and the rest ``added'' parameters. Similarly, we say that the first $u$ entries of the activations are ``carried'' coordinates and the remainder are ``added'' coordinates. 

\begin{definition}[$(\rho,\rho')$-lift]\label{def:structured lift}
Let $\theta\in\mathcal C_M(u,D)$, let $W>u$, and let $0<\rho\leq1$ and $0<\rho'\leq M$. We say that $\phi$ is a $(\rho,\rho')$-lift of $\theta$ from width $u$ to width $W$ if:
\begin{enumerate}[label=\Roman*.]
\item $A_k\in I_\rho(W_k^\theta)$, $a_k\in I_\rho(b_k^\theta)$, and $Q_1\in I_\rho(R^\theta)$;
\item $B_k,D_k,Q_2$ are arbitrary in $[-M,M]$ entrywise;
\item $C_1,C_k\in[-\rho',\rho']$ entrywise, and every coordinate of $c_k$ lies in $[-M,-M+2\rho']$.
\end{enumerate}
We write $\mathscr L_{u\to W}^{\rho,\rho'}(\theta)$ for the set of all such lifts.
\end{definition}

For $u\geq1$ and $H>0$, define
\[
J_{u,H}:=\max\{pX+1,\;u(H+1)+1\}.
\]

\begin{lemma}[Added activations have no effect]
\label{lemma:continuous lift simulates subnet}
Let $\phi\in\mathscr L_{u\to W}^{\rho,\rho'}(\theta)$, and let $\tilde\theta\in\mathcal C_M(u,D)$ be the carried parameter formed from $A_1,\ldots,A_D,a_1,\ldots,a_D,Q_1$. Assume that the activations of $\tilde\theta$ are bounded by $H+1$. If
\[
\rho'\leq\frac{M}{J_{u,H}+1},
\]
then
\[
g_\phi^{W,D}(x)=g_{\tilde\theta}^{u,D}(x)
\qquad
\text{for every }x\in[-X,X]^p.
\]
\end{lemma}

\begin{proof}
Fix $x\in[-X,X]^p$. Write the layer-$k$ activation of $\phi$ as $x_k^\phi=(\kappa_k,\iota_k)$, where $\kappa_k\in\R^u$ denotes the carried coordinates and
$\iota_k\in\R^{W-u}$ the added coordinates. We show inductively that
\[
\kappa_k=x_k^{\tilde\theta},
\qquad
\iota_k=0
\]
for every $k\in[D]$.

At the first layer,
\[
W_1^\phi x+b_1^\phi
=
\begin{pmatrix}
A_1\\
C_1
\end{pmatrix}x
+
\begin{pmatrix}
a_1\\
c_1
\end{pmatrix}.
\]
Hence $\kappa_1=\sigma(A_1x+a_1)=x_1^{\tilde\theta}$.
For each added coordinate, using $\|x\|_\infty\leq X$ and the
definition of the lift,
\[
(C_1x+c_1)[j]
\leq
pX\rho'-M+2\rho'
\leq
(J_{u,H}+1)\rho'-M
\leq0.
\]
Thus $\iota_1=0$.

Now suppose that $x^{\phi}_{k - 1} = (\kappa_{k - 1}, \iota_{k - 1})$ where
\[
\kappa_{k-1}=x_{k-1}^{\tilde\theta},
\qquad
\iota_{k-1}=0.
\]
Then
\[
W_k^\phi x_{k-1}^\phi+b_k^\phi
=
\begin{pmatrix}
A_k&B_k\\
C_k&D_k
\end{pmatrix}
\begin{pmatrix}
x_{k-1}^{\tilde\theta}\\
0
\end{pmatrix}
+
\begin{pmatrix}
a_k\\
c_k
\end{pmatrix}.
\]
The carried preactivation is therefore $A_kx_{k-1}^{\tilde\theta}+a_k$,
so $\kappa_k=x_k^{\tilde\theta}$.

For every added coordinate, the assumed activation bound
$\|x_{k-1}^{\tilde\theta}\|_\infty\leq H+1$ gives
\[
(C_kx_{k-1}^{\tilde\theta}+c_k)[j]
\leq
u(H+1)\rho'-M+2\rho'
\leq
(J_{u,H}+1)\rho'-M
\leq0.
\]
Hence $\iota_k=0$, completing the induction.

Finally,
\[
x_D^\phi
=
\begin{pmatrix}
x_D^{\tilde\theta}\\
0
\end{pmatrix},
\]
and therefore, using $R^\phi=(Q_1\;Q_2)$,
\[
g_\phi^{W,D}(x)
=
(Q_1\;Q_2)
\begin{pmatrix}
x_D^{\tilde\theta}\\
0
\end{pmatrix}
=
Q_1x_D^{\tilde\theta}
=
g_{\tilde\theta}^{u,D}(x).
\]
\end{proof}

\begin{lemma}[Lift of a structured parameter]
\label{lem:lift behaves like structured parameter}
Suppose $\theta\in\mathcal C_M(u,D)$ has $(H,L,C)$ structured stability, where $M\geq1$ and $W>u$. Let $\zeta>0$, and suppose
\[
\rho
\leq
\min\left\{
1,
\frac1C,
\frac\zeta L
\right\},
\qquad
\rho'
\leq
\frac{M}{J_{u,H}+1}.
\]
Then every $\phi\in\mathscr L_{u\to W}^{\rho,\rho'}(\theta)$ satisfies
\[
\sup_{x\in[-X,X]^p}
\|g_\phi^{W,D}(x)-g_\theta^{u,D}(x)\|_\infty
\leq\zeta.
\]
\end{lemma}

\begin{proof}
The carried parameter $\tilde\theta$ satisfies $\|\tilde\theta-\theta\|_\infty\leq\rho$. Structured stability therefore gives
\[
\|x^{\tilde \theta}_k \|_\infty \leq H + 1\qquad \sup_x
\|g_{\tilde\theta}(x)-g_\theta(x)\|_\infty
\leq L\rho\leq\zeta.
\]
Due to the activation bound, we can apply Lemma \ref{lemma:continuous lift simulates subnet} to get $g_\phi(x)=g_{\tilde\theta}(x)$ for all $x \in [-X, X]^p$.
\end{proof}

For any $u\geq1$, let
\[
N_{u,D}:=up+(D-1)u^2+Du+us,
\]
\[
F_{u,W,D}:=(D-1)(W-u)W+s(W-u),
\]
\[
K_{u,W,D}:=(W-u)p+(D-1)(W-u)u+(W-u)D.
\]

$N_{u, D}$ is the number of parameters in a neural network of width $u$ and depth $D$, and is also the number of weights chosen from an interval of length $\rho$ in a $(\rho, \rho')$ lift from $u$ to $W$. $F_{u, W, D}$ is the number of parameters in a $(\rho, \rho')$-lift to from $u$ to $W$ that are freely chosen in $[-M, M]$. $K_{u, W, D}$ is the number of parameters in a $(\rho, \rho')$ lift chosen from an interval of length $\rho'$. 

Thus
\begin{equation}
N_{W,D}=N_{u,D}+F_{u,W,D}+K_{u,W,D}.
\label{eq:structured parameter count identity}
\end{equation}

\begin{lemma}[Volume of structured lifts]
\label{lemma:continuous denom bound}
Let $\theta\in\mathcal C_M(u,D)$, let $0<\rho\leq1$, and let $0<\rho'\leq M$. Then
\[
\lambda\left(\mathscr L_{u\to W}^{\rho,\rho'}(\theta)\right)
=
(2M)^{F_{u,W,D}}(2\rho')^{K_{u,W,D}}\rho^{N_{u,D}}.
\]
\end{lemma}

\begin{proof}
The lift is a Cartesian product of intervals: $F_{u,W,D}$ coordinates have length $2M$, $K_{u,W,D}$ coordinates have length $2\rho'$, and $N_{u,D}$ coordinates have length $\rho$.
\end{proof}

For fixed data, let
\[
B_t
:=
S_{\tau,n}^{W,D,M}
\cap
\{\theta\in\mathcal C_M(W,D):\Delta_\tau(\theta)>t\}.
\]

In the next Lemma, we assume without loss of generality that all labels except the first $m_0$ are from $f$. This is equivalent to assuming that the well-labeled points are chosen independently of the realization of $x_1 \dots x_n$. 

\begin{lemma}[Continuous numerator bound]
\label{lemma:continuous num_bound}
Let $t\in[0,1]$ and $x_1,\ldots,x_n\sim\mu_X$ i.i.d.
Assume $y_i=f(x_i)$ for $i=m_0+1,\ldots,n$, with no restriction on the first $m_0$ labels. Then
\[
\E_{x^n\sim\mu_X^{\otimes n}}[\lambda(B_t)]
\leq
(2M)^{N_{W,D}}e^{-t(n-m_0)}.
\]
\end{lemma}

\begin{proof}
By Tonelli's theorem,
\[
\E[\lambda(B_t)]
=
\int
\Pr[\phi\in B_t]
\,d\lambda(\phi).
\]
If $\Delta_\tau(\phi)>t$, then membership in $S_{\tau,n}^{W,D,M}$ requires $\tau$-accurate prediction on all $n-m_0$ uncorrupted observations. Thus
\[
\Pr[\phi\in B_t]
\leq
(1-t)^{n-m_0}
\leq
e^{-t(n-m_0)}.
\]
Integrating over $[-M,M]^{N_{W,D}}$ proves the result.
\end{proof}

Using a union bound over subsets of size $m_0$, we may extend the previous lemma to the case where the subset of well-labeled points depends arbitrarily on $x_1 \dots x_n$. 

\begin{lemma}[Adversarial continuous numerator bound]
\label{lemma:adaptive continuous num_bound}
Let $t\in[0,1]$ and $x_1,\ldots,x_n\sim\mu_X$ i.i.d.
Suppose that, after observing $x^n$, there is a possibly data-dependent subset $I=I(x^n)\subseteq[n]$ with $|I|\leq m_0$ such that $y_i=f(x_i)$ for every $i\notin I$, with no restriction on the labels $\{y_i:i\in I\}$. Then
\[
\E_{x^n\sim\mu_X^{\otimes n}}[\lambda(B_t)]
\leq
(2M)^{N_{W,D}}\binom{n}{m_0}e^{-t(n-m_0)}.
\]
\end{lemma}
\begin{proof}
For each fixed $J\subseteq[n]$ with $|J|=m_0$, define
\[
B_{t,J}
:=
\left\{
\phi\in\mathcal C_M(W,D):
\Delta_\tau(\phi)>t,\ 
\|g_\phi^{W,D}(x_i)-f(x_i)\|_\infty\leq\tau
\text{ for every }i\notin J
\right\}.
\]

Since $|I(x^n)|\leq m_0$,
\[
B_t
\subseteq
\bigcup_{\substack{J\subseteq[n]\\ |J|=m_0}} B_{t,J},
\]
and therefore
\[
\lambda(B_t)
\leq
\sum_{\substack{J\subseteq[n]\\ |J|=m_0}}
\lambda(B_{t,J}).
\]

Taking expectations and applying Lemma
\ref{lemma:continuous num_bound} to each fixed $J$,
after relabeling the observations, gives
\[
\begin{aligned}
\E[\lambda(B_t)]
&\leq
\sum_{\substack{J\subseteq[n]\\ |J|=m_0}}
\E[\lambda(B_{t,J})] \\
&\leq
\binom{n}{m_0}
(2M)^{N_{W,D}}e^{-t(n-m_0)}.
\end{aligned}
\]
\end{proof}

\subsection{Generalization from a Structured Interpolator}

\begin{theorem}[Generalization from a structured interpolator]
\label{thm:stable witness bound}
Let $x_1,\ldots,x_n\sim\mu_X$ i.i.d., $t\geq0$, and $0\leq\eta<1$.
Assume $y_i=f(x_i)$ for $i=m_0+1,\ldots,n$, with no restriction on the first $m_0$ labels. Fix $u<W$ and $D,M,H,L,C$ independently of the realized sample, with $M\geq1$, $\tau>0$, and $H,L,C>0$. Suppose there is a training-data event $\mathcal G$ with probability at least $1-\eta$ such that, on $\mathcal G$, there exists $\theta^\dagger\in\mathcal C_M(u,D)$ with $(H,L,C)$ structured stability and
\[
\max_{i\in[n]}
\|g_{\theta^\dagger}^{u,D}(x_i)-y_i\|_\infty
\leq\frac\tau2.
\]
Define
\[
\rho
:=
\min\left\{
1,
\frac1C,
\frac{\tau}{2L}
\right\},
\qquad
\rho'
:=
\frac{M}{J_{u,H}+1},
\]
\[
q_{\mathrm{eff}}
:=
\max\left\{
\frac{2M}{\rho},
\frac{M}{\rho'}
\right\}.
\] Then
\[
\Pr_{x^n,\hat\theta\sim U_{\tau,n}^{W,D,M}}
[\Delta_\tau(\hat\theta)>t]
\leq
\eta
+
\left(\frac{2M}{\rho}\right)^{N_{u,D}}
\left(\frac{M}{\rho'}\right)^{K_{u,W,D}}
e^{-t(n-m_0)}.
\]
In particular,
\[
\Pr[\Delta_\tau(\hat\theta)>t]
\leq
\eta
+
q_{\mathrm{eff}}^{N_{u,D}+K_{u,W,D}}
e^{-t(n-m_0)}.
\]
\end{theorem}

\begin{proof}
For $t\geq1$ the event is empty, so consider $0\leq t<1$.
On $\mathcal G$, Lemma \ref{lem:lift behaves like structured parameter}, with $\zeta=\tau/2$, gives
\[
\mathscr L_{u\to W}^{\rho,\rho'}(\theta^\dagger)\subseteq S_{\tau,n}^{W,D,M}.
\]
Lemma \ref{lemma:continuous denom bound} therefore gives
\[
\lambda(S_{\tau,n}^{W,D,M})\geq
V:=(2M)^{F_{u,W,D}}(2\rho')^{K_{u,W,D}}\rho^{N_{u,D}}>0
\qquad\text{on }\mathcal G.
\]
Split the probability before using a volume ratio. With $B_t$ as defined above,
\[
\begin{aligned}
\Pr[\Delta_\tau(\hat\theta)>t]
&\leq\eta+\Pr(\mathcal G)\,
\mathbb E\!\left[\left.
\frac{\lambda(B_t)}{\lambda(S_{\tau,n}^{W,D,M})}
\,\right|\,\mathcal G\right]\\
&\leq\eta+\frac{\mathbb E[\mathbf 1_{\mathcal G}\lambda(B_t)]}{V}\\
&\leq\eta+\frac{\mathbb E[\lambda(B_t)]}{V}.
\end{aligned}
\]
The conditional ratio is evaluated only on $\mathcal G$, where its denominator is positive. By Lemma \ref{lemma:continuous num_bound}, the last expression is at most
\[
\eta+\frac{(2M)^{N_{W,D}}e^{-t(n-m_0)}}
{(2M)^{F_{u,W,D}}(2\rho')^{K_{u,W,D}}\rho^{N_{u,D}}}.
\]
Using \eqref{eq:structured parameter count identity} yields the stated bound.

\end{proof}

\begin{corollary}[Structured interpolator rate]\label{cor:structured parameter rate}
In the setting of Theorem \ref{thm:stable witness bound}, assume
$0\leq\eta<\delta<1$ and $m_0\leq uD$. Then, for a constant $C_0$
depending only on $p$ and $s$, we have
\[
\Pr[\Delta_\tau(\hat\theta)>t_n]\leq\delta, \qquad 
t_n
:=
C_0\frac{uWD\log q_{\mathrm{eff}}+\log\frac1{\delta-\eta}}n
\]

Here 
\begin{equation}\label{eq:q eff from stability}
q_{\mathrm{eff}}\leq\max\left\{2M\max\{1,C,2L/\tau\},J_{u,H}+1\right\}.
\end{equation}

\end{corollary}

\begin{proof}
By definition,
\[
N_{u,D}+K_{u,W,D}
=
Wp+(D-1)Wu+DW+su
\leq
C_1uWD
\]
for a constant $C_1$ depending only on $p$ and $s$. If $t_n>1$,
the result is immediate. Otherwise,
\[
t_n(n-m_0)
\geq
nt_n-m_0.
\]
Since $m_0\leq uD$, $W\geq1$, and $q_{\mathrm{eff}}\geq2$, the term
$uWD\log q_{\mathrm{eff}}$ absorbs $m_0$. Taking $C_0$
sufficiently large makes the exponential term in Theorem
\ref{thm:stable witness bound} at most $\delta-\eta$.

The bound on $q_\mathrm{eff}$ follows from its definition in Theorem \ref{thm:stable witness bound}.
\end{proof}

\begin{corollary}[Adaptive structured interpolator rate]\label{cor:adaptive structured parameter rate}
In the setting of Theorem \ref{thm:stable witness bound}, with $M\geq1$, replace the assumption that the first $m_0$ labels are unrestricted by the assumption that the labels are induced by an $(m, \Gamma)$ adversarial corruption (Assumption \ref{body ass:adversarial corruptions}), and set $m_0:=m$. Define
\[
\bar q_{\mathrm{eff}}:=\max\{en,q_{\mathrm{eff}}\}.
\]
Assume $0\leq\eta<\delta<1$ and $m_0\leq uD$. Then, for a constant $C_0$ depending only on $p$ and $s$,
\[
t_n
:=
C_0\frac{uWD\log \bar q_{\mathrm{eff}}+\log\frac1{\delta-\eta}}n
\]
satisfies
\[
\Pr[\Delta_\tau(\hat\theta)>t_n]\leq\delta.
\]
\end{corollary}

\begin{proof}
The proof follows Theorem \ref{thm:stable witness bound} and Corollary \ref{cor:structured parameter rate}, using Lemma \ref{lemma:adaptive continuous num_bound} in place of Lemma \ref{lemma:continuous num_bound}. If $m_0=0$, the additional factor $\binom{n}{m_0}$ is one. If $m_0\geq1$, it is absorbed by the complexity term since
\[
\log\binom{n}{m_0}
\leq
m_0\log\frac{en}{m_0}
\leq
uD\log(en)
\leq
uWD\log\bar q_{\mathrm{eff}}.
\]
\end{proof}

By construction,
\[
q_{\mathrm{eff}}
\leq
\max\left\{
2M\max\left\{1,C,\frac{2L}{\tau}\right\},
J_{u,H}+1
\right\}.
\]
Thus the rate depends only logarithmically on the stability certificate. Every continuous interpolation theorem below follows by furnishing deterministic a high-probability $\tau/2$-interpolating parameter with $(H, L, C)$ structured stability, and then invoking Corollary \ref{cor:structured parameter rate}.

\subsection{Disturbance Stability and Existence of Structured Interpolators}\label{apdx subsec:existence of strucftured interpolators}

Structured Stability (Definition \ref{def:structured stability}) relates to the stability of a computation of a DNN under a perturbation to its weights.  We now draw a possibly surprising connection between Structured Stability and \emph{Disturbance Stability}, which  relates to perturbation of \emph{inputs} to the network. A \emph{disturbance} is a bounded adversarial perturbation injected to a DNN's computation at each layer, as a function of the current activations. For a parameter $\theta$, we write
\[
\theta=(W_1,\ldots,W_D,b_1,\ldots,b_D,R)\in\mathcal C_M(u,D)
\]
and, for  $x\in[-X,X]^p$:
\[
x_0^\theta(x):=x,
\qquad
x_k^\theta(x):=\sigma(W_kx_{k-1}^\theta(x)+b_k),
\quad k\in[D].
\]
Given functions
\[
\xi_k:[-X,X]^p\to\R^u,
\quad k\in[D],
\qquad
\xi_{D+1}:[-X,X]^p\to\R^s,
\]
we define the corresponding disturbed computation by
\[
x_0^{\theta,\xi}(x):=x,
\qquad 
x_k^{\theta,\xi}(x)
:=
\sigma\left(W_kx_{k-1}^{\theta,\xi}(x)+b_k+\xi_k(x)\right), \quad k\in[D],
\]
and
\[
g_\theta^\xi(x):=Rx_D^{\theta,\xi}(x)+\xi_{D+1}(x).
\]
We call this a \textbf{$\nu$-disturbed}  computation if
\[
\sup_{x\in[-X,X]^p}\|\xi_k(x)\|_\infty\leq\nu
\qquad
\text{for every }k\in[D+1].
\]

\begin{definition}[Disturbance Stability]\label{def:distrubance stability}
We say that $\theta$ has $(H,l,c)$ disturbance stability, with $H,l,c>0$, if its exact hidden
activations satisfy
\[
\sup_{x\in[-X,X]^p}\|x_k^\theta(x)\|_\infty\leq H
\qquad
\text{for every }k\in[D],
\]
and every $\nu$-disturbed computation with $\nu\leq1/c$ satisfies
\[
\sup_{x\in[-X,X]^p}\max_{k \in [D]}
\|x_k^{\theta,\xi}(x)-x_k^\theta(x)\|_\infty
\leq l\nu,
\quad
\|x_k^{\theta,\xi}(x)\|_\infty\leq H+1,
\quad
\sup_{x\in[-X,X]^p}
\|g_\theta^{\xi, u, D}(x)-g^{u, D}_\theta(x)\|_\infty
\leq l\nu.
\]
\end{definition}

Definition \ref{def:distrubance stability} is related to Adversarial Robustness \citep{madry2018towards}, but calls for an even stronger guarantee of robustness given possibly adaptive perturbations injected at each layer\footnote{The adversarial robustness literature focuses on learning adversarially robust DNNs. We deal exclusively with existence of robust DNNs. Our proof technique does not establish that $\hat{\theta}$ will have robustness.}. However, Disturbance Stability has useful compositional properties compared to Structured Stability, and we will be able to establish that it holds for parameters of interest, $\theta$. 

Notice also the extra requirement that intermediate activations remain similar under disturbances. Structured Stability does make any requirements on intermediate activations, but requiring it for Disturbance Stability facilitates constitutionality. 

We now show that Disturbance Stability implies Structured Stability. 

Define the quantity

\begin{equation}\label{eq:J_uH def}
J_{u,H}:=\max\{pX+1,\;u(H+1)+1\}.
\end{equation}

\begin{lemma}[Disturbance stability implies structured stability]
\label{lem:disturbance implies structured}
Suppose $\theta\in\mathcal C_M(u,D)$ has $(H,l,c)$ disturbance stability. Then
$\theta$ has $(H,L,C)$ structured stability with
\[
C=J_{u,H}c,
\qquad
L=J_{u,H}l.
\]
\end{lemma}

\begin{proof}
Let
\[
\tilde\theta
=
(\tilde W_1,\ldots,\tilde W_D,
\tilde b_1,\ldots,\tilde b_D,\tilde R)
\in\mathcal C_M(u,D)
\]
be a coordinatewise $\rho$-perturbation of $\theta$, so that
$\|\tilde\theta-\theta\|_\infty\leq\rho$. Write
\[
\tilde W_k=W_k+\Delta_k,
\qquad
\tilde b_k=b_k+\zeta_k,
\qquad
\tilde R=R+\Delta_R,
\]
where every entry of $\Delta_k$, $\zeta_k$, and $\Delta_R$ has magnitude at most $\rho$.

We define a disturbance
\[
\xi_1(x):=\Delta_1x+\zeta_1,
\qquad
\xi_k(x):=\Delta_kx_{k-1}^{\tilde\theta}(x)+\zeta_k
\quad (k\geq2),
\qquad
\xi_{D+1}(x):=\Delta_Rx_D^{\tilde\theta}(x),
\]
such that, by construction, 
\[
x_k^{\theta,\xi}(x)=x_k^{\tilde\theta}(x)
\quad (k\in[D]),
\qquad
g_\theta^\xi(x)=g_{\tilde\theta}(x).
\]

We now must show that if $\rho\leq\frac{1}{J_{u,H}c}$, then $(\xi_k)$ is a disturbance of size at most $J_{u,H}\rho\leq1/c$. So assume $\rho \leq \frac{1}{J_{u, H}c}.$

We begin by establishing the activation bound $\|x^{\tilde \theta}_k(x)\|_\infty \leq H + 1$. Define
\[
k_*
:=
\min\left(
\left\{
k\in[D]:
\sup_{x\in[-X,X]^p}
\|x_k^{\tilde\theta}(x)\|_\infty>H+1
\right\}
\cup\{D+1\}
\right),
\]
so that $\sup_x\|x_j^{\tilde\theta}(x)\|_\infty\leq H+1$
for every $j < k_*$.  At the first layer,
\[
\|\xi_1(x)\|_\infty
\leq
p\rho\|x\|_\infty+\rho
\leq
(pX+1)\rho
\leq
J_{u,H}\rho \leq \frac{1}{c}.
\]
For any $2\leq k\leq\min\{k_*,D\}$, we have $k-1<k_*$, and hence $\|x_{k-1}^{\tilde\theta}(x)\|_\infty\leq H+1$.
Therefore
\[
\|\xi_k(x)\|_\infty
\leq
u\rho\|x_{k-1}^{\tilde\theta}(x)\|_\infty+\rho
\leq
\bigl(u(H+1)+1\bigr)\rho
\leq
J_{u,H}\rho \leq \frac{1}{c}.
\]

Suppose that $k_*\leq D$. Define a complete disturbance sequence $\bar\xi$ by
\[
\bar\xi_k:=\xi_k \quad (k\leq k_*),
\qquad
\bar\xi_k:=0 \quad (k>k_*).
\]
Then $\bar\xi$ is a $J_{u,H}\rho$-disturbed computation. Since $J_{u,H}\rho\leq1/c$, disturbance stability gives
\[
\sup_x
\|x_{k_*}^{\theta,\bar\xi}(x)\|_\infty
\leq H+1
\]
But the computations of $(\theta,\bar\xi)$ and $\tilde\theta$ agree through layer $k_*$, so
\[
x_{k_*}^{\theta,\bar\xi}(x)
=
x_{k_*}^{\tilde\theta}(x),
\]
contradicting the assumption that $k_* \leq D$. Thus $k_*=D+1$, and our previous calculations tell us that 
\[
\sup_x\|x_k^{\tilde\theta}(x)\|_\infty\leq H+1, 
\qquad
\sup_x\|\xi_k(x)\|_\infty
\leq
J_{u,H}\rho \leq \frac{1}{c}
\qquad
\text{for every }k\in[D].
\]
Finally, at the output,
\[
\|\xi_{D+1}(x)\|_\infty
=
\|\Delta_Rx_D^{\tilde\theta}(x)\|_\infty
\leq
u\rho\|x_D^{\tilde\theta}(x)\|_\infty
\leq
u(H+1)\rho
\leq
J_{u,H}\rho \leq \frac{1}{c}.
\]

Hence the computation of $\tilde\theta$ is a $\nu$-disturbed computation of $\theta$ with $\nu:=J_{u,H}\rho\leq\frac1c$. Disturbance stability now gives
\[
\sup_x
\|g_{\tilde\theta}(x)-g_\theta(x)\|_\infty
=
\sup_x
\|g_\theta^\xi(x)-g_\theta(x)\|_\infty
\leq
l\nu
=
J_{u,H}l\rho.
\]
Since the exact activations of $\theta$ are already bounded by $H$, this proves structured stability with
\[
C=J_{u,H}c,
\qquad
L=J_{u,H}l.
\]
\end{proof}

In the previous proof, we did not need to use the closeness of the disturbed activations, since structured stability does not itself have any such requirement. 

\begin{lemma}[Activation bounds for bounded parameters and identity padding]
\label{lem:padded activation bound}
Let $\theta_0\in\mathcal C_M(w,d)$, and let $x_k^{\theta_0}(x)$ denote its layer-$k$
hidden activation. Define
\[
A_1:=M(pX+1),\qquad 
A_k:=M(wA_{k-1}+1).
\]
Then
\[
\sup_{x\in[-X,X]^p}\|x_k^{\theta_0}(x)\|_\infty\leq A_k
\qquad
\text{for every }k\in[d].
\]
If $\theta\in\mathcal C_M(w,D)$ is the $d$-to-$D$ identity padding of
$\theta_0$, then every hidden activation of $\theta$ is bounded by
\[
H_{\mathrm{pad}}:=\max\{1,A_d\}.
\]
Moreover,
\[
A_d\leq d(Mw)^d(Xp+2).
\]
\end{lemma}

\begin{proof}
At the first layer,
\[
x_1^{\theta_0}(x)=\sigma(W_1x+b_1).
\]
For every output coordinate $j$,
\[
\begin{aligned}
|(W_1x+b_1)[j]|
&\leq
\sum_{a=1}^p|W_1[j,a]||x[a]|+|b_1[j]|\\
&\leq
pMX+M\\
&=
M(pX+1)\\
&=
A_1.
\end{aligned}
\]
Since $|\sigma(z)|\leq|z|$, we obtain
\[
\|x_1^{\theta_0}(x)\|_\infty\leq A_1.
\]
Now suppose $\|x_{k-1}^{\theta_0}(x)\|_\infty\leq A_{k-1}$ for some $2\leq k\leq d$. For each coordinate $j$,
\[
\begin{aligned}
|(W_kx_{k-1}^{\theta_0}(x)+b_k)[j]|
&\leq
\sum_{a=1}^w|W_k[j,a]||x_{k-1}^{\theta_0}(x)[a]|+|b_k[j]|\\
&\leq
wMA_{k-1}+M\\
&=
M(wA_{k-1}+1)\\
&=
A_k.
\end{aligned}
\]
Applying ReLU gives
\[
\|x_k^{\theta_0}(x)\|_\infty\leq A_k \qquad (k \leq d).
\]

The layers after depth $d$ have identity weight
matrices and zero biases. Every hidden activation is coordinatewise
nonnegative, so for $k>d$,
\[
x_k^\theta(x)=\sigma(I_wx_{k-1}^\theta(x))=x_{k-1}^\theta(x).
\]
Thus
\[
x_k^\theta(x)=x_d^\theta(x)
\qquad
\text{for every }k\geq d,
\]
and every hidden activation is bounded by $A_d$, hence also by
$H_{\mathrm{pad}}$.

Finally, unrolling the recurrence gives
\[
A_d=(Mw)^{d-1}A_1+M\sum_{j=0}^{d-2}(Mw)^j.
\]
Since $M,w\geq1$,
\[
A_d
\leq
(Mw)^d(Xp+1)+d(Mw)^d
\leq
d(Mw)^d(Xp+2).
\]
\end{proof}

We now show that, for a completely generic parameter, disturbance stability parameters decay exponentially in depth. 

\begin{lemma}[Structured stability of $d$-deep parameters]
\label{lemma:structured stability bounded parameters}
Fix $X>0$, $M\geq1$, and $D\geq d$. Let
$\theta_0\in\mathcal C_M(w,d)$.

Define
\[
A_1:=M(pX+1),
\qquad
A_k:=M(wA_{k-1}+1),\qquad 
S_d:=\sum_{j=0}^{d-1}(Mw)^j.
\]
Then $\theta_0$ has $(H_0,l_0,c_0)$ disturbance stability with
\[
H_0:=A_d,
\qquad
c_0:=\max\{1,S_d\},
\qquad
l_0:=MwS_d+1,
\]
\end{lemma}

\begin{proof}

Consider the computation of $\theta_0$  under a disturbance $\xi = \{\xi_k\}$ of size $\nu \leq \frac{1}{c_0}$. Define 
\[
e_k
:=
\sup_{x\in[-X,X]^p}
\|x_k^{\theta_0,\xi}(x)-x_k^{\theta_0}(x)\|_\infty,
\qquad
e_0:=0.
\]
Then  $e_k\leq Mw e_{k-1}+\nu$,so that 
\[
e_k
\leq
\left(
\sum_{j=0}^{k-1}(Mw)^j
\right)\nu
\leq
S_d\nu
\qquad
\text{for every }k\in[d].
\]
By our choice of $\nu$, $e_k\leq1$ and so 
\[
\sup_x
\|x_k^{\theta_0,\xi}(x)\|_\infty
\leq
A_d+1
=
H_0+1.
\]
Since $S_d \leq l_0$, we also have that, for all $k \in [d]$
\[
\sup_x
\|x_k^{\theta_0,\xi}(x)-x_k^{\theta_0}(x)\|_\infty
\leq
l_0\nu
\]

Finally, at the readout,
\[
\begin{aligned}
\sup_x
\|g_{\theta_0}^\xi(x)-g_{\theta_0}(x)\|_\infty
&\leq
Mw e_d+\nu\leq 
(MwS_d+1)\nu= 
l_0\nu.
\end{aligned}
\]
Thus $\theta_0$ has $(H_0,l_0,c_0)$ disturbance stability.
\end{proof}

Although a generic $d$-deep parameter has stability that deteriorates as $d(Mw)^d$, if we pad that same parameter with identity layers so that it becomes an equivalent $D$-deep parameter, stability deteriorates only polynomially in $D$ and exponentially in $d$. This is what the next Lemma and its corollary show. 

\begin{lemma}[Disturbance stability under identity padding]
\label{lemma:disturbance stability padding H l c}
Fix $M\geq1$ and $D\geq d$. Let
$\theta_0\in\mathcal C_M(w,d)$ have $(H,l,c)$ disturbance stability,
and let $\theta\in\mathcal C_M(w,D)$ be its $d$-to-$D$ identity padding.
Define
\[
H_{\mathrm{pad}}:=H,
\qquad
c_{\mathrm{pad}}:=\max\left\{1,c,{l + D - d}\right\},
\qquad
l_{\mathrm{pad}}:=Mw(l + D - d + 1)
\]
Then $\theta$ has
$(H_{\mathrm{pad}},l_{\mathrm{pad}},c_{\mathrm{pad}})$ disturbance
stability. Consequently, it has
$(H_{\mathrm{pad}},L_{\mathrm{pad}},C_{\mathrm{pad}})$ structured
stability with
\[
C_{\mathrm{pad}}
=
J_{w,H_{\mathrm{pad}}}c_{\mathrm{pad}},
\qquad
L_{\mathrm{pad}}
=
J_{w,H_{\mathrm{pad}}}l_{\mathrm{pad}}.
\]
\end{lemma}

\begin{proof}
By construction,
\[
x_k^\theta(x)
=
x_d^{\theta_0}(x),
\qquad
k\geq d.
\]
Thus every activation of $\theta$ is bounded by $H\leq H_{\mathrm{pad}}$.

We now consider the computation of $\theta$ under a disturbance $\xi = \{\xi_k\}$ of size $\nu \leq \frac{1}{c_\mathrm{pad}}$.

For $k \leq d$, disturbance stability of $\theta_0$ and the definition of $\theta$  gives us 

\[
\sup_x
\|x_k^{\theta,\xi}(x)-x_k^\theta(x)\|_\infty
\leq l\nu,\qquad 
\sup_x\|x_k^{\theta,\xi}(x)\|_\infty
\leq H+1.
\]

Moreover, for $k > d$, we now show $\sup_x \|x_k^{\theta, \xi}(x) - x^{\theta}_k(x)\|_\infty \leq (l + k - d)$ by induction. The case $k = d$ is shown. For the inductive step, we have

\begin{align*}\sup_x \|x_k^{\theta, \xi}(x) - x^{\theta}_k(x)\|_\infty &= \sup_x \|\sigma(x^{\theta, \xi}_{k - 1}(x) + \xi_k(x)) - \sigma(x^{\theta}_{k - 1}(x))\|_\infty \\ 
&\leq \sup_x\|x^{\theta, \xi}_{k - 1}(x) + \xi_k(x) - x^\theta_{k - 1}(x)\|_\infty \\
&\leq \sup_x\|x^{\theta, \xi}_{k - 1}(x) - x^\theta_{k - 1}(x)\|_\infty + \nu \\ 
&\leq  (l + k - 1 - d)\nu + \nu = (l + k - d)\nu
\end{align*}

In particular, we have shown that for all $k \in [D]$, \[\sup_x \|x_k^{\theta, \xi}(x) - x^{\theta}_k(x)\|_\infty \leq (l + D - d)\nu \leq l_\mathrm{pad}\nu\]. 

For the activation bound assumption, we have, using the size bound on $\nu$ and  the above equation

\[\sup_x \|x_k^{\theta, \xi}(x)\|_\infty \leq \sup_x \|x_k^{\theta}(x)\|_\infty + \sup_x\|x_k^{\theta}(x) - x_k^{\theta, \xi}(x)\|_\infty \leq H  + 1\]

Finally, it remains to show that the readout of $\theta$ is stable under the disturbance:

\[\sup_x \|g_\theta^\xi(x) - g_\theta(x)\|_\infty = \sup_x \|R(x_D^{\theta, \xi}(x) - x^\theta_D)(x) + \xi_{D + 1}(x)\|_\infty \leq Mw(l + D - d + 1)\nu.\]
\end{proof}
For the noiseless result, a direct parameter-perturbation argument gives a sharper structured-stability bound by retaining the activation size at each layer.

\begin{lemma}[Direct structured stability under identity padding]
\label{lem:direct padded structured stability}
Let $X>0$, $M\geq1$, and $D\geq d\geq1$. Let
$\theta^\dagger\in\mathcal C_M(w,D)$ be the $d$-to-$D$ identity padding of
$\theta_0\in\mathcal C_M(w,d)$. Write $a:=Mw$, $b:=pX+1$, and
\[
A_1:=Mb,\qquad A_k:=aA_{k-1}+M\quad(2\leq k\leq d).
\]
Then $\theta^\dagger$ has $(H_{\mathrm{pad}},L_{\mathrm{pad}},C_{\mathrm{pad}})$ structured stability with
\[
\begin{aligned}
H_{\mathrm{pad}}&:=A_d,\\
C_{\mathrm{pad}}&:=3D(b+d)a^d,\\
L_{\mathrm{pad}}&:=(3aD+1)(b+d)a^d.
\end{aligned}
\]
In particular, all three parameters are bounded by
$(Mw)^d\mathrm{poly}(X,M,p,d,D,w)$.
\end{lemma}
\begin{proof}
Abbreviate the claimed parameters by $H,L,C$ and set $T:=D-d$.
The activation recurrence gives
\[
A_k\leq M(b+k-1)a^{k-1}.
\]
Since $a\geq1$, the sequence $A_k$ is nondecreasing. Thus all exact hidden activations are bounded by $H=A_d$, including the identity-padded layers.

Fix $\widetilde\theta\in\mathcal C_M(w,D)$ with
$\|\widetilde\theta-\theta^\dagger\|_\infty\leq\rho\leq C^{-1}$, and define
\[
e_k:=\sup_{x\in[-X,X]^p}
\|x_k^{\widetilde\theta}(x)-x_k^{\theta^\dagger}(x)\|_\infty.
\]
For the first layer, $e_1\leq b\rho$. For $2\leq k\leq d$, decompose the preactivation difference as
\[
\widetilde W_k(\widetilde x_{k-1}-x_{k-1})
+(\widetilde W_k-W_k)x_{k-1}
+(\widetilde b_k-b_k).
\]
The absolute row sums of $\widetilde W_k$ are at most $a$ because the perturbed parameter remains in $\mathcal C_M(w,D)$. Since ReLU is $1$-Lipschitz,
\[
e_k\leq ae_{k-1}+(wA_{k-1}+1)\rho
=ae_{k-1}+\frac{A_k}{M}\rho.
\]
Unrolling this recurrence yields, for $1\leq k\leq d$,
\[
e_k\leq B_k\rho,\qquad
B_k:=\left(kb+\frac{k(k-1)}2\right)a^{k-1}
\leq k(b+k)a^{k-1}.
\]

For each padded layer, $W_k=I_w$ and $b_k=0$, so the absolute row sums of $\widetilde W_k$ are at most $1+w\rho$. Therefore
\[
e_k\leq(1+w\rho)e_{k-1}+(wH+1)\rho\qquad(k>d).
\]
Since $C\geq wT$, we have $wT\rho\leq1$ and hence
$(1+w\rho)^T\leq\exp(wT\rho)\leq e<3$. For $0\leq t\leq T$, it follows that
\[
\begin{aligned}
e_{d+t}
&\leq(1+w\rho)^t\bigl(B_d+t(wH+1)\bigr)\rho\\
&\leq3\bigl(B_d+T(wH+1)\bigr)\rho.
\end{aligned}
\]
Using
\[
wH+1\leq(b+d)a^d,\qquad B_d\leq d(b+d)a^d,
\]
we obtain $e_k\leq C\rho\leq1$ for every hidden layer (also $B_k\leq B_d$ for $k\leq d$). Thus every perturbed hidden activation is bounded by $H+1$.

Finally, at the readout, the absolute row sums of $\widetilde R$ are at most $a$, and those of $\widetilde R-R$ are at most $w\rho$. Hence
\[
\begin{aligned}
\sup_x\|g_{\widetilde\theta}(x)-g_{\theta^\dagger}(x)\|_\infty
&\leq ae_D+wH\rho\\
&\leq(aC+wH)\rho\leq L\rho.
\end{aligned}
\]
This proves structured stability. The claimed dependence follows from
$H\leq(b+d)a^d$ and the explicit formulas for $C,L$.
\end{proof}

\begin{lemma}[Disturbance stability under parallel composition]
\label{lem:parallel disturbance stability}
Let $\theta_i\in\mathcal C_M(w_i,D)$ have $(H_i,l_i,c_i)$ disturbance
stability for $i\in\{1,2\}$, and suppose they have the same output
dimension. Let $\theta_1\oplus\theta_2$ denote the parameter obtained
by running them in parallel and summing their outputs. Then
$\theta_1\oplus\theta_2$ has
\[
\left(
\max\{H_1,H_2\},
\,l_1+l_2,
\,\max\{c_1,c_2\}
\right)
\]
disturbance stability.
\end{lemma}

\begin{proof}
A $\nu$-disturbed computation of the parallel network decomposes into
$\nu$-disturbed computations of the two subnetworks. If
$\nu\leq1/\max\{c_1,c_2\}$, both disturbance-stability guarantees
apply, so the combined hidden disturbed activations are bounded by $\max\{H_1,H_2\}+1$, and are $l _1 \vee  l_2$-close to their undisturbed counterparts. Since the outputs are summed, their errors are
at most $(l_1+l_2)\nu$; assign the output disturbance to either one
of the two subnetworks. The claim follows.
\end{proof}
\begin{lemma}[Structured stability after memorization]
\label{lemma:structured stability}
Let $\theta\in\mathcal C_M(w,D)$ have $(H_0,l_0,c_0)$ disturbance stability, and assume $M\geq M_{X,\Gamma,\ell}$.
Let $\theta_{\mathrm{mem}}$ be Lemma \ref{lem:real valued memorization}'s
$(n,m,\ell)$ memorization gadget, and define
\[
b:=\left\lceil\frac mD\right\rceil,
\qquad
w':=w+2+2s+4b.
\]
The memorization gadget has
$(H_{\mathrm{mem}},l_{\mathrm{mem}},c_{\mathrm{mem}})$ disturbance stability,
where
\[
H_{\mathrm{mem}}
:=
\max\{1,2X+2\ell,\Gamma m\},
\]
\[
c_z:=D,
\qquad
c_s:=2D+1,
\qquad
c_{p,n}:=
D\left(1+\frac{4b\Gamma}{\ell}c_s\right),
\]
\[
c_{\mathrm{mem}}
:=
\max\{1,c_z,c_s,c_{p,n}\},
\qquad
l_{\mathrm{mem}}
:=
2c_{p,n}+\frac{4b\Gamma}{\ell}c_s+1.
\]

Let $\theta^\dagger\in\mathcal C_M(w',D)$ be obtained by running
$\theta$ and $\theta_{\mathrm{mem}}$ in parallel and summing their outputs.
Then $\theta^\dagger$ has $(H,l,c)$ disturbance stability, where
\[
H
:=
\max\{H_0,H_{\mathrm{mem}}\}
=
\max\{H_0,1,2X+2\ell,\Gamma m\},
\]
\[
c
:=
\max\{c_0,c_{\mathrm{mem}}\}
=
\max\{1,c_0,c_z,c_s,c_{p,n}\},
\]
and
\[
l
:=
l_0+l_{\mathrm{mem}}
=
l_0+2c_{p,n}+\frac{4b\Gamma}{\ell}c_s+1.
\]
Consequently, $\theta^\dagger$ has $(H,L,C)$ structured stability with
\[
C=J_{w',H}c,
\qquad
L=J_{w',H}l.
\]

If $\theta$ is the $d$-to-$D$ identity padding of some
$\theta_0\in\mathcal C_M(w,d)$, then, for an absolute constant $C_0$,
\[
H,C,L
\leq
K\,
p w'dD^2
Xm^2\Gamma^2(1+\ell^{-1})
(Mw)^{2d+1}
\]
where $K$ is an absolute constant.
\end{lemma}

\begin{proof}
We first establish disturbance stability of $\theta_{\mathrm{mem}}$ alone.

Write the exact layer-$k$ hidden activation of the memorization gadget as
\[
x_k^{\theta_{\mathrm{mem}}}=(z_k,p_k,n_k,s_k),
\]
where $z_k=(Z_k^+,Z_k^-)\in\R^2$ is the first-coordinate copy channel,
$p_k,n_k\in\R^s$ are the positive and negative accumulation channels, and
$s_k\in\R^{4b}$ is the hinge-feature channel.

We first bound the exact activations. For the memorization gadget, the two copy
coordinates are defined as $Z_k^+=\sigma(x[1])$ and
$Z_k^-=\sigma(-x[1])$, so that
\[
\|z_k\|_\infty\leq X.
\]
Every hinge point associated with an active memory slot lies between
$-X-2\ell$ and $X+2\ell$. For unused slots, choose the padded hinge points in
this same interval and set their correction coefficients to zero. Since the
input coordinate lies in $[-X,X]$, every hinge feature is therefore bounded by
$2X+2\ell$, so
\[
\|s_k\|_\infty\leq2X+2\ell.
\]
Each normalized trapezoidal contribution $\ell^{-1}\psi_i(x[1])$ lies in
$[0,1]$. Each coordinate of $\delta_i^+$ and $\delta_i^-$ is at most
$\Gamma$, and at most $m$ corrections are accumulated. Consequently,
\[
\|p_k\|_\infty,\|n_k\|_\infty\leq\Gamma m.
\]
Thus the exact hidden activations of the memorization gadget are bounded by
\[
H_{\mathrm{mem}}
=
\max\{1,2X+2\ell,\Gamma m\}.
\]

Now consider a $\nu$-disturbed computation of
$\theta_{\mathrm{mem}}$. Define
\[
e_k^z:=\sup_x
\|z_k^{\theta_{\mathrm{mem}},\xi}(x)-z_k(x)\|_\infty,
\qquad
e_k^s:=\sup_x
\|s_k^{\theta_{\mathrm{mem}},\xi}(x)-s_k(x)\|_\infty,
\]
\[
e_k^p:=\sup_x
\|p_k^{\theta_{\mathrm{mem}},\xi}(x)-p_k(x)\|_\infty,
\qquad
e_k^n:=\sup_x
\|n_k^{\theta_{\mathrm{mem}},\xi}(x)-n_k(x)\|_\infty,
\]
and
\[
e_k^{p,n}:=\max\{e_k^p,e_k^n\}.
\]
Set the corresponding layer-zero errors equal to zero.

\paragraph{Disturbed activation errors.}
We prove simultaneously by induction on $k$ that
\[
e_k^z\leq k\nu,
\qquad
e_k^s\leq(2k-1)\nu,
\]
and
\[
e_k^{p,n}
\leq
k\left(1+\frac{4b\Gamma}{\ell}c_s\right)\nu,
\]
where
\[
c_s:=2D+1.
\]

At the first layer, the exact and disturbed memory channels receive the same
input, so their differences arise only from the additive disturbance. Hence
\[
e_1^z,e_1^s,e_1^p,e_1^n\leq\nu.
\]
The claimed bounds therefore hold for $k=1$.

Now let $k\geq2$ and assume the bounds hold at layer $k-1$.

\textbf{Copy channel.}
The copy channels are propagated by identity weights, so, since ReLU is
$1$-Lipschitz,
\[
e_k^z
\leq
e_{k-1}^z+\nu
\leq
k\nu.
\]

\textbf{Hinge channel.}
Each hinge preactivation depends on
$Z_{k-1}^+-Z_{k-1}^-$ and a fixed bias. Thus
\[
e_k^s
\leq
2e_{k-1}^z+\nu
\leq
(2k-1)\nu.
\]

\textbf{Accumulation channels.}
Consider one coordinate of the positive accumulation channel. Its exact
preactivation has the form
\[
p_{k-1}[j]
+
\sum_{i\in C_{k-1}}
\frac{\delta_i^+[j]}{\ell}\Psi_i(s_{k-1}),
\]
where $|C_{k-1}|\leq b$ and
\[
\Psi_i(s)=s^{i,1}-s^{i,2}-s^{i,3}+s^{i,4}
\]
is the trapezoid built from the four hinge points corresponding to point $i$.
For every $i$,
\[
|\Psi_i(s_{k-1}^{\theta_{\mathrm{mem}},\xi})
-\Psi_i(s_{k-1})|
\leq
4e_{k-1}^s.
\]
Since $0\leq\delta_i^+[j]\leq\Gamma$ and there are at most $b$ active
terms,
\[
e_k^p
\leq
e_{k-1}^p
+
\frac{4b\Gamma}{\ell}e_{k-1}^s
+
\nu.
\]
The same bound holds for the negative accumulation channel, and hence
\[
e_k^{p,n}
\leq
e_{k-1}^{p,n}
+
\frac{4b\Gamma}{\ell}e_{k-1}^s
+
\nu.
\]
Since
\[
e_{k-1}^s
\leq
(2k-3)\nu
\leq
c_s\nu,
\]
the inductive hypothesis gives
\[
\begin{aligned}
e_k^{p,n}
&\leq
(k-1)\left(1+\frac{4b\Gamma}{\ell}c_s\right)\nu
+
\frac{4b\Gamma}{\ell}c_s\nu
+
\nu\\
&=
k\left(1+\frac{4b\Gamma}{\ell}c_s\right)\nu.
\end{aligned}
\]
This completes the induction.

\paragraph{Stability of the memorization gadget.}
Define
\[
c_z:=D,
\qquad
c_{p,n}:=
D\left(1+\frac{4b\Gamma}{\ell}c_s\right),
\qquad
c_{\mathrm{mem}}
:=
\max\{1,c_z,c_s,c_{p,n}\}.
\]
If $\nu\leq1/c_{\mathrm{mem}}$, then
\[
e_k^z,e_k^s,e_k^p,e_k^n\leq1.
\]
Combining these bounds with the exact activation bounds gives, for every
$k\in[D]$,
\[
\sup_x
\|x_k^{\theta_{\mathrm{mem}},\xi}(x)\|_\infty
\leq
H_{\mathrm{mem}}+1.
\]

Define
\[
l_{\mathrm{mem}}
:=
2c_{p,n}
+
\frac{4b\Gamma}{\ell}c_s
+
1.
\]
Since
\[
l_{\mathrm{mem}}\geq c_z,
\qquad
l_{\mathrm{mem}}\geq c_s,
\qquad
l_{\mathrm{mem}}\geq c_{p,n},
\]
the preceding bounds also give
\[
\sup_x\max_{k\in[D]}
\|x_k^{\theta_{\mathrm{mem}},\xi}(x)
-x_k^{\theta_{\mathrm{mem}}}(x)\|_\infty
\leq
l_{\mathrm{mem}}\nu.
\]

It remains to bound the gadget output. Evaluating the preceding bounds at
$k=D$, the $P$ and $N$ channels contribute at most
$2c_{p,n}\nu$, while the final hinge group contributes at most
$\frac{4b\Gamma}{\ell}c_s\nu$.
Including the output disturbance, we obtain
\[
\sup_x
\|g_{\theta_{\mathrm{mem}}}^\xi(x)
-g_{\theta_{\mathrm{mem}}}(x)\|_\infty
\leq
\left(
2c_{p,n}
+
\frac{4b\Gamma}{\ell}c_s
+
1
\right)\nu
=
l_{\mathrm{mem}}\nu.
\]
Thus $\theta_{\mathrm{mem}}$ has
$(H_{\mathrm{mem}},l_{\mathrm{mem}},c_{\mathrm{mem}})$ disturbance
stability.

Finally, applying Lemma \ref{lem:parallel disturbance stability} to the
$(H_0,l_0,c_0)$ disturbance stability of $\theta$ and the $(H_\mathrm{mem},l_\mathrm{mem},c_\mathrm{mem})$ disturbance stability of  $\theta_{\mathrm{mem}}$, we obtain disturbance stability of the parallelized parameter
$\theta^\dagger$ with
\[
H
=
\max\{H_0,H_{\mathrm{mem}}\}
=
\max\{H_0,1,2X+2\ell,\Gamma m\},
\]
\[
c
=
\max\{c_0,c_{\mathrm{mem}}\}
=
\max\{1,c_0,c_z,c_s,c_{p,n}\},
\]
and
\[
l
=
l_0+l_{\mathrm{mem}}
=
l_0+2c_{p,n}+\frac{4b\Gamma}{\ell}c_s+1.
\]
From Lemma \ref{lem:disturbance implies structured}, we finally get Structured Stability with 
\[
C=J_{w',H}c,
\qquad
L=J_{w',H}l.
\]

Substituting in the definitions and bounding coarsely gives 
\[
H,C,L
\lesssim
C\,
p w'dD^2
Xm^2\Gamma^2(1+\ell^{-1})
(Mw)^{2d+1}.
\]
\end{proof}
\subsection{Barron Functions}\label{apdx sec:Barron theory}

We use the ReLU Barron norm studied in \cite{e2022barron}. A function $f:B_p\to\R$ has Barron norm
\begin{equation}\label{eq:barron norm def}
\|f\|_{\mathcal B}
:=
\inf_\rho
\E_\rho\left[|a|(\|b\|_1+|c|)\right],
\end{equation}
where the infimum is over probability measures $\rho(da, db, dc)$ such that 
\[
f(x)=\E_\rho[a\sigma(b^\top x+c)]
\qquad\text{for every }x\in B_p.
\]

\begin{lemma}[Barron approximation]\label{lem:barron approximation}
Let $Q:=\|f\|_{\mathcal B}$. For every $r\geq1$, there are coefficients
$a_k\in\R$ and $\beta_k\in\R^{p+1}$ such that
\[
\sum_{k=1}^r|a_k|\leq Q,
\qquad
\|\beta_k\|_2=1
\quad\text{for every }k\in[r],
\]
and
\[
f_r(x):=\sum_{k=1}^r a_k\sigma(\beta_k^\top(x,1))
\]
satisfies
\[
\|f-f_r\|_{L^\infty(B_p)}
\leq
C_p Q r^{-\frac12-\frac{3}{2p}},
\]
where $C_p$ depends only on $p$.
\end{lemma}

\begin{proof}
If $Q=0$, then $f=0$ and we take all $a_k=0$. Hence assume $Q>0$. In
\eqref{eq:barron norm def}, assume for simplicity that the infimum is
attained\footnote{If it is not attained, the same conclusion follows by a limiting argument.},
so that there is a representing measure $\rho$ satisfying
\[
\E_\rho\left[|a|(\|b\|_1+|c|)\right]=Q.
\]

For $(b,c)\neq0$, define
\[
s(b,c):=\sqrt{\|b\|_2^2+c^2},
\qquad
v(b,c):=\frac{(b,c)}{s(b,c)}\in\mathbb S^p.
\]
When $(b,c)=0$, define $v(b,c)$ arbitrarily. By positive homogeneity of ReLU,
\[
a\sigma(b^\top x+c)
=
a\,s(b,c)\,
\sigma\left(v(b,c)^\top(x,1)\right).
\]

Define a finite signed measure $\mu$ on $\mathbb S^p$ as the pushforward under $(a, b, c) \mapsto v(b, c)$ of $as(b, c)d\rho(a, b, c)$:
\[
\mu(A)
:=
\int
\mathbf 1\{v(b,c)\in A\}\,
a\,s(b,c)\,
d\rho(a,b,c),
\qquad
A\subseteq\mathbb S^p.
\]

Then
\[
\begin{aligned}
f(x)
&=
\int a\sigma(b^\top x+c)\,d\rho(a,b,c)= 
\int_{\mathbb S^p}
\sigma\left(v^\top(x,1)\right)\,d\mu(v).
\end{aligned}
\]
and 
\[
\begin{aligned}
\|\mu\|\leq 
\int |a|\,s(b,c)\,d\rho(a,b,c) \leq
\int |a|(\|b\|_1+|c|)\,d\rho(a,b,c)\leq 
Q.
\end{aligned}
\]
Thus $f\in\mathcal F_\sigma(Q)$, in Equation (1.2) of 
\cite{yangzhou2025optimal}. Equivalently, $f/Q\in\mathcal F_\sigma(1)$. Lemma 2.3 of \cite{yangzhou2025optimal} now gives
$g_r\in\mathcal F_\sigma(r,1)$ such that
\[
\left\|\frac{f}{Q}-g_r\right\|_{L^\infty(B_p)}
\leq
C_p r^{-\frac12-\frac{3}{2p}}.
\]
Setting $f_r:=Qg_r$ gives
$f_r\in\mathcal F_\sigma(r,Q)$ and
\[
\|f-f_r\|_{L^\infty(B_p)}
\leq
C_p Q r^{-\frac12-\frac{3}{2p}}.
\]
By the definition of $\mathcal F_\sigma(r,Q)$, $f_r$ has the displayed
form with $\|\beta_k\|_2=1$ and
$\sum_{k=1}^r|a_k|\leq Q$.
\end{proof}

We now use depth to implement the function $f_r$ at constant width. The technique below has precedent in e.g. \cite{VardiYS22}, but the key fact is that introducing a copy channel for the input and building features from that channel prevents disturbances from accumulating exponentially. 

\begin{lemma}[Deep Barron witness]\label{lem:serial barron witness}
The function $f_r$ in Lemma \ref{lem:barron approximation} is realized
by a width-$(2p+3)$ and depth-$r$ ReLU parameter
\[
\theta_r\in\mathcal C_{Q\vee1}(2p+3,r).
\]
Moreover, $\theta_r$ has
$(H_{p,Q},l_{p,Q}r,c_{p,Q}r)$ disturbance stability, where
\[
H_{p,Q},\ l_{p,Q},\ c_{p,Q}
\leq C_p(1+Q)
\]
for a constant $C_p$ depending only on $p$.
\end{lemma}
\begin{proof}
We describe a deep network that uses additional copy channels for the input $x$, as well as channels for a running sum that simulates the wide computation. 

\paragraph{Defining the network} Write $\beta_k=(b_k,c_k)\in\R^p\times\R$, and decompose $a_k=a_k^+-a_k^-$ into its positive and negative parts.  By Lemma \ref{lem:barron approximation},
\[
\sum_{k=1}^r(a_k^++a_k^-)
=
\sum_{k=1}^r|a_k|
\leq Q,
\qquad
\|\beta_k\|_2=1.
\]

We notate the exact layer-$k$ hidden activation as $x_k^{\theta_r}=(z_k,p_k,n_k,h_k)$, where $z_k=(Z_k^+,Z_k^-)\in\R^{2p}$ carries the input, $p_k,n_k\in\R$ contain the positive and negative parts of the running sum, and $h_k\in\R$ prepares the next summand. 
At the first layer, set
\[
Z_1^+=\sigma(x),
\qquad
Z_1^-=\sigma(-x),
\qquad
p_1=n_1=0,
\qquad
h_1=\sigma(b_1^\top x+c_1).
\]
For $k=2,\ldots,r$, copy the $Z^+$ and $Z^-$ channels and set
\[
p_k
=
\sigma\left(p_{k-1}+a_{k-1}^+h_{k-1}\right),
\qquad
n_k
=
\sigma\left(n_{k-1}+a_{k-1}^-h_{k-1}\right),
\qquad 
h_k
=
\sigma\left(
b_k^\top Z_{k-1}^+
-
b_k^\top Z_{k-1}^-
+
c_k
\right).
\]
Since $Z_k^+-Z_k^-=x$, we have $h_k=\sigma(\beta_k^\top(x,1))$ at all $k \in [r]$. 
Because $p_k$ and $n_k$ are nonnegative by definition, ReLU acts
as the identity on those channels, and we have
\[
p_r=\sum_{k=1}^{r-1}a_k^+h_k,
\qquad
n_r=\sum_{k=1}^{r-1}a_k^-h_k.
\]
We define our readout as 
\begin{equation}\label{eq:barron serial readout}
p_r-n_r+a_rh_r,
\end{equation}
which gives $f_r(x)$ by construction.

\paragraph{Parameter magnitude} In the above construction, all parameters have magnitude at most $Q\vee1$: $\|\beta_k\|_2=1$ implies that every entry of $b_k$ and $c_k$ has
magnitude at most one, while
\[
|a_k|\leq Q \implies a_k^+,a_k^- \leq Q.
\]
Therefore $\theta_r\in\mathcal C_{Q\vee1}(2p+3,r)$.

\paragraph{Stability} 
We now verify that $\theta_r$ satisfies disturbance stability with parameters depending only on $p$ and (polynomially) on $Q$. Here, we verify stability for inputs, $x$, in the larger set $ [-1, 1]^p \supset B_p$.

First, activation bounds. Clearly $\|z_k\|_\infty \leq 1$ for all $k$.  Since
$\|\beta_k\|_2=1$, we have $|h_k| \leq \|(x, 1)\|_2   \leq \sqrt{p+1}$. Using $\sum_k|a_k|\leq Q$, the $p_k$ and $n_k$
channels are bounded by $Q\sqrt{p+1}$. Hence all exact hidden
activations are bounded by some
\[
H_{p,Q}\leq C_p(1+Q),
\]

where $C_p$ depends on $p$ (as $\sqrt{p})$. Let $e_k^z,e_k^p,e_k^n,e_k^h$ denote the maximum coordinatewise
errors in the corresponding blocks under any $\nu$-disturbed
computation, in analogy to the computations in previous sections. Also let $e^{p, n}_k = e^p_k \vee e^n_k$. 

The copied input channels clearly satisfy $e_k^z\leq k\nu$.

Since $\|b_k\|_1\leq\sqrt p$, we have (using ReLU's Lipschitzness)
\[
e_k^h
\leq
2\|b_k\|_1e_{k-1}^z+\nu
\leq
(2\sqrt p(k-1)+1)\nu
\leq C_p k\nu.
\]
For the accumulation channels,
\[
e_k^{p,n}
\leq
e_{k-1}^{p,n}+|a_{k-1}|e_{k-1}^h+\nu.
\]
Since $\sum_{k=1}^r|a_k|\leq Q$, it follows that
\[
e_k^{p,n}
\leq
C_p(1+Q)r\nu.
\]
Choose $c_{p,Q}=C_p(1+Q)$, independently of $r$, with $C_p$ sufficiently large. Then
\[
\nu\leq(c_{p,Q}r)^{-1}
\quad\Longrightarrow\quad
\max_{k\in[r]}\max\{e_k^z,e_k^h,e_k^p,e_k^n\}\leq1.
\]
In particular, every disturbed hidden activation is bounded by $H_{p,Q}+1$.

Finally, using \eqref{eq:barron serial readout} and including the output
disturbance,
\[
\begin{aligned}
\sup_x
\left|
g_{\theta_r}^{2p+3,r,\xi}(x)
-
g_{\theta_r}^{2p+3,r}(x)
\right|
&\leq
e_r^p+e_r^n+|a_r|e_r^h+\nu\\
&\leq
C_p(1+Q)r\nu.
\end{aligned}
\]
Thus, with $l_{p,Q}=c_{p,Q}=C_p(1+Q)$, the disturbance-stability parameters are $(H_{p,Q},l_{p,Q}r,c_{p,Q}r)$. 
\end{proof}

\section{Proofs of main results}\label{apdx sec:proofs of main results}

\subsection{Proofs for Section \ref{subsec:noiseless realizable}}


\subsubsection{Proof of Theorem \ref{body thm:continuous noiseless thm}}

Let $\theta\in\mathcal C_M(w,d)$ be the witness satisfying
$\sup_{x\in\supp(\mu_X)}\|g_\theta^{w,d}(x)-f(x)\|_\infty\leq\tau/2$.
Let $\theta^\dagger\in\mathcal C_M(w,D)$ be its $d$-to-$D$ identity padding.
Padding preserves the output, so $\theta^\dagger$ $\tau/2$-interpolates every realized sample.
By Lemma \ref{lem:direct padded structured stability}, this padded parameter has structured stability with
\[
\begin{aligned}
H_{\mathrm{pad}}&=A_d,\\
C_{\mathrm{pad}}&=3D(pX+1+d)(Mw)^d,\\
L_{\mathrm{pad}}&=(3MwD+1)(pX+1+d)(Mw)^d.
\end{aligned}
\]
In particular,
\[
H_{\mathrm{pad}},L_{\mathrm{pad}},C_{\mathrm{pad}}
\leq(Mw)^d\mathrm{poly}(X,M,p,d,D,w).
\]

Apply Corollary \ref{cor:structured parameter rate} with $u=w$, $m_0=0$, $\eta=0$, and the padded witness $\theta^\dagger$.
For completeness, use the exact definition in Theorem \ref{thm:stable witness bound}:
\[
q_{\mathrm{eff}}=
\max\left\{2M\max\{1,C_{\mathrm{pad}},2L_{\mathrm{pad}}/\tau\},
J_{w,H_{\mathrm{pad}}}+1\right\}.
\]
Write $a:=Mw$, $b:=pX+1$, and $v:=\max\{1,\tau^{-1}\}$.
Since $d\leq D$ and $a\geq1$,
\[
H_{\mathrm{pad}}\leq D(b+1)a^d,\qquad
1\leq C_{\mathrm{pad}}\leq L_{\mathrm{pad}}
\leq4aD^2(b+1)a^d.
\]
Thus the first term in the maximum defining $q_{\mathrm{eff}}$ is at most
$4ML_{\mathrm{pad}}v\leq16Ma^{d+1}D^2(b+1)v$.
The second term obeys
\[
\begin{aligned}
J_{w,H_{\mathrm{pad}}}+1
&=\max\{b,w(H_{\mathrm{pad}}+1)+1\}+1\\
&\leq b+wH_{\mathrm{pad}}+w+2\\
&\leq4aD(b+1)a^d,
\end{aligned}
\]
which is also at most the same upper bound. Consequently,
\[
q_{\mathrm{eff}}\leq16Ma^{d+1}D^2(b+1)v
\leq\bigl(16MwD(pX+2)\max\{1,\tau^{-1}\}\bigr)^{d+2}.
\]
Taking logarithms proves the displayed bound in the theorem. Substituting this into the structured-interpolator rate gives the claimed dependence on the witness depth.

\subsubsection{Generalization with stable witness (Strengthening of Theorem \ref{body thm:continuous noiseless thm})}
Theorem \ref{body thm:continuous noiseless thm} relies on a crude exponential-in-$d$ bound on the structured stability of its witness. Below, we give a generic version of the theorem, which follows immediately from Corollary \ref{cor:structured parameter rate} and allows us to bound $q_\mathrm{eff}$ in terms of the stability of the provided witness. As we saw in the Barron example, approximation theory can often provide witnesses that are deep, yet stable, which makes results like Corollary \ref{cor:structured parameter rate} and the following Theorem useful. 

Below,  recall the definition that
\[
J_{u,H}:=\max\{pX+1,\;u(H+1)+1\}.
\]
\begin{theorem}[Disturbance-stable interpolator rate]\label{thm:disturbance stable rate}
Assume $y_i=f(x_i)$ for $i=1,\ldots,n$. Fix $u<W$ and $D,M,H,l,c$ independently of the realized sample, with $M\geq1$, $\tau>0$, and $H,l,c>0$. Suppose there is a training-data event $\mathcal G$ with probability at least $1-\eta$ such that, on $\mathcal G$, there exists $\theta^\dagger\in\mathcal C_M(u,D)$ with $(H,l,c)$ disturbance stability and
\[
\max_{i\in[n]}
\|g_{\theta^\dagger}^{u,D}(x_i)-y_i\|_\infty
\leq\frac\tau2.
\]
Assume $0\leq\eta<\delta<1$.

Then
\[
\Pr[\Delta_\tau(\hat\theta)\geq t_n]\leq\delta; \qquad t_n:=C_0\frac{uWD\log q_{\mathrm{eff}}+\log\frac1{\delta-\eta}}n.
\]
Here, $C_0$ is a constant depending only (linearly) on $p$ and $s$ and 
\[q_{\mathrm{eff}}
\leq
\max\left\{
2M\max\left\{1,J_{u,H}c,\frac{2J_{u,H}l}{\tau}\right\},
J_{u,H}+1
\right\}.
\]
\end{theorem}

\begin{proof}
    
Take $\theta^\dagger \in \mathcal C_M(u, D)$ so that $\max_i \|g_{\theta^\dagger}^{u, D}(x_i) - y_i\|_\infty \leq \tau/2$. By Lemma \ref{lem:disturbance implies structured}, $\theta^\dagger$ has $(H,L,C)$ structured stability with $C=J_{u,H}c$ and $L=J_{u,H}l$. Corollary \ref{cor:structured parameter rate} therefore applies with $m_0=0$. Substituting these values of $C$ and $L$ into the definition of $q_{\mathrm{eff}}$ gives the statement. 
\end{proof}

\section{Proof of Theorem \ref{body thm:noisy informal}}
\label{apdx sec:proof of continuous noisy thm}

\begin{theorem}[Theorem \ref{body thm:noisy informal}, formal]\label{apdx thm:continuous noisy main thm}
Let $\tau>0$, $\delta\in(0,1)$, $\Gamma>0$, and $n\geq2$. Consider a dataset of $n$ points $\{(x_i, y_i)\}_{i = 1}^n$, where $x_1 \dots x_n \sim \mu_X^{\otimes n}$ and $\mu_X$ satisfies Assumption \ref{body ass:lambda marginal boundedness}. Moreover, assume labels $\{y_i\}$ are $(m, \Gamma)$ adversarial corruptions from a ground truth $f$, as in Assumption \ref{body ass:adversarial corruptions}.

Let $M \geq \max \{1, X + 2\ell_n, \frac{\Gamma}{\ell_n}\}$, with $\ell_n = \frac{\delta}{6\Lambda n(n - 1)}$. Assume that, for some $w$ and $d$, there exists $\theta \in \mathcal C_M(w, d)$ so that $\|f(x)-  g_{\theta}^{w, d}(x)\|_\infty \leq \frac{\tau}{2}$ for all $x \in [-X, X]^p$. Let $D \geq d$ and $W > w + 2 + 2s + 4\lceil \frac{m}{D}\rceil =: w'$.

Then there is a constant $C_0$ depending only on $p$ and $s$, so that for 
\[ \Pr_{x^n \sim \mu^{\otimes n}, \hat{\theta} \sim U^M_{\tau, n}}\left[\Delta_\tau(\hat{\theta}) \geq t_n\right] \leq \delta; \qquad t_n := C_0\frac{(wWD + Wm)\log q_{\mathrm{eff}, n} + \log\frac{1}{\delta}}{n}.\]

Here, $q_{\mathrm{eff}, n}$ is a value satisfying 
\[\log q_{\mathrm{eff}, n} \leq 2d\log \left(\mathrm{poly}(\Gamma, \Lambda, X, n, m, w, p, s, \ell_n^{-1})M\max\{1,\tau^{-1}\}D\right).\]
\end{theorem}

\begin{proof}

Set $\eta=\delta/2$. Take the assumed $\theta \in \mathcal C_M(w, d)$ and pad it from depth $d$ to depth $D$ with identity layers. Now apply Lemma \ref{lem:interpolation via memorization}, which tells us that on an event $\mathcal G$ (over the training data) with probability at least $1-\eta$, for any $I$ with $|I| = m$ and $\{\delta_i\}_{i =1 }^m \subset [-\Gamma, \Gamma]^s$ we have a parameter $\theta^\dagger\in\mathcal C_M(w',D)$ so that $g_\theta^{w', D}(x_i) = g_\theta^{w, D}(x_i)$ for $i \notin I$, and $g_\theta^{w', D}(x_i) = g_\theta^{w, D}(x_i) + \delta_i$ for $i \in I$. Here, 
\[
w'=w+2+2s+4\left\lceil\frac mD\right\rceil.
\]

By Lemma \ref{lemma:structured stability}, any such $\theta^\dagger$ has structured stability with  \[H,C,L
\lesssim 
p w'dD^2
Xm^2\Gamma^2(1+\ell^{-1})
(Mw)^{2d+1}.\]

Above, we clearly see that  $m\leq w'D$, so Corollary \ref{cor:adaptive structured parameter rate} applies with $u=w'$, $m_0=m$, and $\eta=\delta/2$. Since
\[
w'WD\leq C_1(wWD+Wm)
\]
for a constant $C_1$ depending only on $p$ and $s$, we have 

\[\Pr\left[\Delta_\tau(\hat\theta ) \geq \frac{C_0(wWD + Wm)\log \bar q_\mathrm{eff, n} + \log \frac{2}{\delta}}{n}\right] \leq \delta.\]

Here, $\log \bar q_{\mathrm{eff},n}=\max\{\log q_{\mathrm{eff}},1+\log n\}$; the theorem uses this enlarged scale as $q_{\mathrm{eff},n}$, and so $\log q_{\mathrm{eff}, n} \leq 2d\log \left(\mathrm{poly}(\Gamma, \Lambda, X, n, m, w, p, s, \ell_n^{-1})M\max\{1,\tau^{-1}\}D\right)$ due to Lemma \ref{lemma:structured stability}.
\end{proof}
\section{Proofs of Corollary \ref{body cor:tempered regression}, and Theorem  \ref{body thm:unrealizable thm informal}}

\subsection{Proof of Theorem \ref{body cor:tempered regression}}

In this section, we state formally and prove Theorem \ref{body thm:unrealizable thm informal}. In the following assumption, we formally introduce the corruption setting for Theorem \ref{body thm:unrealizable thm informal}. 

\begin{assumption}[Constant rate adversarial corruptions]\label{ass:epsilon corrupted adversarial}
For an $\epsilon \in (0, 1)$, assume $m \sim \mathrm{Binom}(n, \epsilon)$ independently of the realization of the data. Then we assume that the data undergoes an $(m, \Gamma)$ adversarial corruption.  
\end{assumption}

The above setting includes the following special case.

\begin{assumption}[$\epsilon$-corrupted data model]\label{ass:epsilon corrupted}
Under the $\epsilon$-corruption model, we assume that each label $y_i \in \R^s$ is corrupted by a bounded corruption $\delta_i \in [-\Gamma, \Gamma]^s$ with (independent) probability $\epsilon$. Concretely, for $\delta_1 \dots \delta_n$ possibly depending on $x_1 \dots x_n$, 
\[y_i = \begin{cases}
f(x_i) & \text{with probability } 1 - \epsilon \\ 
f(x_i) + \delta_i & \text{with probability } \epsilon
\end{cases}\]

We may assume without loss of generality that the corrupted points are $y_1 \dots y_m$ for $m \sim \mathrm{Binom}(n, \epsilon)$, because the algorithm we are studying is invariant under the reordering of data points.     
\end{assumption}

\begin{theorem}[Corollary \ref{body cor:tempered regression}, formal]\label{cor: continuous epsilon theorem}
Let $\epsilon, \delta \in (0, 1)$ and $\tau > 0$. Assume $\mu_X$ satisfies Assumption \ref{body ass:lambda marginal boundedness}, and that we observe labels according to Assumption \ref{ass:epsilon corrupted adversarial} for some ground truth $f$. Assume that there exists $\theta \in \mathcal C_M(w, d)$ so that $f = g_\theta^{w, d}$ on $[-X, X]^p$. 

Let $W > w + 2 + 2s + 4 B$ for some $B\in\mathbb N_{\geq1}$. Let $m_n = \min\{n,\lceil 2\epsilon n + \log \frac{2}{\delta}\rceil\}$ and consider
\[D_n := \max \left\{d, 2, \left\lceil\frac{m_n}{B} \right\rceil\right\}.\]

Assume $M \geq \max \{1, X + 2\ell_n, \frac{\Gamma}{\ell_n}\}$, with $\ell_n = \frac{\delta}{12\Lambda n(n - 1)}$.

Consider a parameter $\hat{\theta}$ sampled from the set of $\tau$-interpolating solutions of width $W$ and depth $D_n$:

\[\hat{\theta} \sim U^{W,D_n,M}_{\tau,n}.\]

Then for $n = \max\left\{2,\left\lceil \frac{\log(2/\delta)}{\epsilon} \vee \frac{Bd-\log(2/\delta)}{\epsilon}\right\rceil\right\} $, we have

\[\Pr_{x^n, \hat{\theta}, m}\left[\Delta_\tau(\hat{\theta}) > 
\tilde O\left(W\left(1+\frac{w}{B}\right)\epsilon 
\right)\right] \leq \delta\]

where $\tilde O(\cdot)$ hides only problem-dependent constants, and log factors of $\frac{1}{\epsilon}$ and $\log \frac{2}{\delta}$, and $B$.
\end{theorem}
\begin{proof}
Apply Lemma \ref{lem:corrupted point bound} with $\eta=\delta/2$. On the resulting event, the sample has at most $m_n$ corruptions. Apply Theorem \ref{apdx thm:continuous noisy main thm} with confidence parameter $\delta/2$, corruption budget $m_n$, and $D_n$. Since
\[
W>w+2+2s+4B
\geq
w+2+2s+4\left\lceil\frac{m_n}{D_n}\right\rceil,
\]
the width condition holds because $B$ is an integer. Write $z=\log(2/\delta)$. The stated choice of $n$ gives $\epsilon n\geq z$ and $\epsilon n\geq Bd-z$, hence $\epsilon n\geq Bd/2$. Consequently,
\[
m_n\leq 2\epsilon n+z+1\leq5\epsilon n,
\qquad
D_n=\max\{d,2,\lceil m_n/B\rceil\}\leq7\epsilon n/B.
\]
Here $1\leq2\epsilon n/B$ and $d\leq2\epsilon n/B$ control the ceiling and the minimum depth. 
Absorbing these constants into $C_0$,
\[
t_n
\leq
C_0W\left(1+\frac wB\right)\epsilon\log q_{\mathrm{eff},n}
+
C_0W\left(1+\frac wB\right)
\frac{\log(2/\delta)\log q_{\mathrm{eff},n}}n
+
\frac{\log(2/\delta)}n.
\]

for the value of $q_{\mathrm{eff}, n}$ of Theorem \ref{apdx thm:continuous noisy main thm}. 

At the stated value of $n$, this upper bound is 
\[
t_n \leq \tilde O\left(W\left(1+\frac wB\right)\epsilon\right).
\]
Union bounding the binomial event and the noisy theorem gives that the above holds with probability at least $1 - \delta$.
\end{proof}
\subsection{Proof of Theorem \ref{body thm:unrealizable thm informal}}

\begin{theorem}[Theorem \ref{body thm:unrealizable thm informal} formal]\label{apdx thm:unrealizable thm formal}
Let $\delta \in (0, 1)$ and $\tau > 0$. Suppose $x_1 \dots x_n \sim \mu_X$ i.i.d. and $y_i = f(x_i)$ for all $i \in [n]$. Assume $\mu_X$ satisfies Assumption \ref{body ass:lambda marginal boundedness}. 

Consider sequences $\epsilon_n\in(0,1]$, $w_n$, $d_n$, $D_n$, $A_n$ with the requirements that $\epsilon_n = \Omega(\frac{1}{n})$ and $D_n\geq d_n$. Suppose that, for all $n\in\mathbb N_{\geq2}$, there is $\theta_{0, n} \in \mathcal C_{A_n}(w_n, d_n)$ that has disturbance stability $(H_{0, n}, l_{0, n}, c_{0, n})$ and satisfies

\[
\mu_X\left(
\left\{
x:
\|g_{\theta_{0, n}}^{w_n,d_n}(x)-f(x)\|_\infty
>
\frac{\tau}{2}
\right\}
\right)
\le \epsilon_n, 
\qquad 
\sup_{x\in[-X,X]^p}
\|g_{\theta_{0, n}}^{w_n,d_n}(x)-f(x)\|_\infty
\le \Gamma.\]

Let $m_n:=\min\{n,\lceil2\epsilon_n n+\log(4/\delta)\rceil\}$.  Let $W_n > w_n + 2 + 2s + 4\lceil \frac{m_n}{D_n}\rceil$, and $M_n \geq \max \{1, A_n,  X + 2\ell_n, \frac{\Gamma}{\ell_n}\}$, where $\ell_n = \frac{\delta}{12\Lambda n(n - 1)}$.

Consider $\hat \theta \sim U^{W_n,D_n,M_n}_{\tau,n}$. Then

\[\Pr_{x^n, \hat{\theta} }\left[\Delta_\tau(\hat{\theta}) \geq
\tilde O\left(W_n\left(\frac{w_nD_n + \log \frac{1}{\delta}}{n} +\epsilon_n\right)\log q_\mathrm{eff, n}\right) \right] \leq \delta\]

where $q_\mathrm{eff} = \mathrm{poly}(n, H_{0, n}, c_{0, n}, l_{0, n}, D_n, w_n, d_n, M_n,X,   \Lambda, \Gamma, \delta^{-1}, \tau^{-1})$. 
\end{theorem}

\begin{proof}
Define the exceptional set
\[
E_n
:=
\left\{
x\in[-X,X]^p:
\|g_{\theta_{0,n}}^{w_n,d_n}(x)-f(x)\|_\infty>\frac{\tau}{2}
\right\}.
\]

By assumption, $\mu_X(E_n)\leq\epsilon_n$. Define $I_{E_n} = \{i \in [n] \ : \ x_i \in E_n\}$ to be the (data-dependent) set of indices corresponding to exceptional points. Lemma \ref{lem:corrupted point bound}, applied with $\epsilon=\mu_X(E_n)$ and $\eta=\delta/4$, gives
\[
|I_{E_n}|\leq m_n
\]
with probability at least $1-\delta/4$. On this event, for each $i\in I_{E_n}$ define
\[
\delta_i:=f(x_i)-g_{\theta_{0,n}}^{w_n,d_n}(x_i).
\]
The uniform approximation bound gives $\|\delta_i\|_\infty\leq\Gamma$.

 Pad $\theta_{0,n}$ from depth $d_n$ to depth $D_n$ to get $\theta_n$, which has disturbance stability with $H_{\mathrm{pad}, n}, l_{\mathrm{pad}, n}, c_{\mathrm{pad}, n}$ by Lemma \ref{lemma:disturbance stability padding H l c}) applied to $\theta_{0, n}$ and $(H_{0, n}, l_{0, n}, c_{0, n})$.  Given the lower bound on $M_n$, we may apply Lemma \ref{lem:interpolation via memorization} with failure probability $\delta/4$, budget $m_n$, subset $I_{E_n}$, and base parameter $\theta_n$. In particular, memorization gadget used uses $\ell =  \frac{\delta}{12 \Lambda n(n -1)}$.

By a union bound, on an event $\mathcal G_n$ with probability at least $1-\delta/2$, there exists
\[
\theta_n^\dagger\in\mathcal C_{M_n}(u_n,D_n),
\qquad
u_n
:=
w_n+2+2s+4\left\lceil\frac{m_n}{D_n}\right\rceil,
\]
such that
\[
\max_{i\in[n]}
\|g_{\theta_n^\dagger}^{u_n,D_n}(x_i)-f(x_i)\|_\infty
\leq\frac{\tau}{2}.
\]
($H_n, l_n, c_n$) disturbance stability stability for $\theta^\dagger_n$ follows from applying Lemma \ref{lemma:structured stability} with $\theta = \theta_{n}$ and its associated disturbance-stable parameters $(H_{\mathrm{pad}, n}, l_{\mathrm{pad}, n}, c_{\mathrm{pad}, n})$.  Substituting $\ell =  \frac{\delta}{12 \Lambda n(n -1)}$ into its expression tells us that $\theta^\dagger_n$ has structured stability with parameters
\begin{equation}\label{eq:unrealizable theorem theta dag stability}
H_n, C_n, L_n\lesssim  \mathrm{poly}(H_{0, n}, l_{0, n}, c_{0, n}, p, s, u_n, d_n, D_, M, \Gamma, \Lambda,  X, n, \delta^{-1}).\end{equation}

Since $W_n>u_n$, we may now apply Corollary \ref{cor:structured parameter rate} with $u=u_n$, $D=D_n$, $m_0=0$, and $\eta=\delta/2$. The numerator uses all $n$ observations because every observed label equals $f(x_i)$. We obtain
\[
\Pr_{x^n\sim\mu_X^{\otimes n},\,\hat\theta\sim U^{W_n,D_n,M_n}_{\tau,n}}
\left[
\Delta_\tau(\hat\theta)\geq t_n
\right]
\leq\delta,\qquad 
t_n
=
C_0\frac{u_nW_nD_n\log q_{\mathrm{eff},n}+\log\frac{2}{\delta}}{n}.
\]

where Theorem \ref{body thm:unrealizable thm informal}'s bound on $q_\mathrm{eff, n}$ follows by combining \eqref{eq:unrealizable theorem theta dag stability} with \eqref{eq:q eff from stability}.

We now further bound $t_n$. Notice that 
$D_n\left\lceil\frac{m_n}{D_n}\right\rceil
\leq m_n+D_n \leq m_n + w_nD_n$
 ($w_n\geq1$), so that, for some constant $C_1$ depending only on $s$:
\[
u_nD_n
\leq
C_1\bigl(w_nD_n+m_n\bigr).
\]
Therefore, we have a $p$ and $s$-dependent constant $C_0$ such that 
\begin{equation}\label{eq:intermediate tn unrealizable pf}
t_n
\leq
C_0W_n
\left(
\frac{w_nD_n+m_n}{n}
\right)
\log q_{\mathrm{eff},n}+
\frac{\log\frac{2}{\delta}}{n}.
\end{equation}

Consider the term $\frac{m_n}{n}\leq\frac{\lceil 2\epsilon_n n\rceil}{n}+\frac{\left\lceil\log\frac{4}{\delta}\right\rceil}{n}$. We have
\[
\frac{\lceil 2\epsilon_n n\rceil}{n}=
O\left(\max\left\{\epsilon_n,\frac1n\right\}\right) = O(\epsilon_n)
\]
since $\epsilon_n=\Omega(n^{-1})$. Thus, plugging back into \eqref{eq:intermediate tn unrealizable pf}
\[
\frac{m_n}{n}
=
O\left(
\epsilon_n+\frac{\log\frac{1}{\delta}}{n}
\right).
\]
This allows us to complete the proof by noting
\[
t_n
=
O\left(
W_n\left(
\frac{w_nD_n+\log\frac{1}{\delta}}{n}
+\epsilon_n
\right)
\log q_{\mathrm{eff},n}
\right).
\]
\end{proof}

\section{Classification}\label{apdx:classification}
In the classification setting, we assume we observe labeled data $\{(x_i, \bar y_i)\}_{i = 1}^n$. As before, we assume $x_i \in [-X, X]^p$ are random features distributed according to $\mu$. However, we now assume that $\bar y_i \in [s]$ is a discrete observation indicating membership in one of $s$ classes. As such, we now interpret the $s$ outputs of the neural network as scores over $s$ classes, with class predictions then given by returning the class with the highest score. We notate the corresponding decision rule as 

\[
c^{w,d}_\theta(x):=\operatorname*{arg\,max}_{j\in[s]}g_\theta^{w,d}(x)[j]
\]

with ties broken arbitrarily. In the classification setting, a parameter $\theta \in \mathcal C_M(w, d)$ \textbf{interpolates} if $c_\theta^{w,d }$ correctly classifies each data point. Given data $\{(x_i, \bar y_i)\}$, a weight-bound $M$, and a target width and depth $W$, $D$, we define

\[
S_n^{W, D, M}
:=
\left\{
\theta\in\mathcal C_M(W,D):
c_\theta^{W,D}(x_i)=\bar y_i
\text{ for all }i\in[n]
\right\},  \quad U^{W, D, M}_n := \mathrm{Unif}(S^{W, D, M}_n),
\]
where the technical details behind the definition of $U^{W, D, M}_n$ are as before (refer to Appendix \ref{apdx sec:uniform dist def}). 

In this section, we focus on the case where $\bar y_i$ reflect partially corrupted labels from a ground truth labeling function, $c:\mathcal X \rightarrow [s]$. We evaluate any estimate ${\theta} \in \mathcal C_M(W, D)$ in terms of its classification error with respect to $c$:

\[\Delta^{\mathrm{cls}}({\theta}; c) = \Pr_{x}\left[c_\theta^{W, D}(x) \neq c(x)\right].\]

Parallel to the regression setting, we give results for a noisy setting where $m$ data points can be corrupted arbitrarily. These can be converted into tempered overfitting-type results exactly as in the regression setting. 

\begin{assumption}[Adversarial noise, classification]\label{ass:adversarial noise, classification}
The labels $\{(x_i, \bar y_i)\}_{i = 1}^n$ are generated as follows: $\{x_i\}_{i = 1}^n \overset{iid}{\sim} \mu_{X}$, and $y_i := c(x_i)$. Then an adversary can change the class label for any $m$ points, resulting in the dataset $\{(x_i, \bar y_i)\}_{i = 1}^n$.
\end{assumption}

We also introduce a margin assumption on the witness: if we wish to measure error with respect to a classifier, $c$, then some $\theta \in \mathcal C_M(w, d)$ must induce $c^{w, d}_\theta = c$ with margin on the logits. 

\begin{assumption}[Realizability with a small witness at margin $\gamma$]\label{ass:realizability with margin}
Assume there exists $\theta \in \mathcal C_M(w, d)$ so that, letting $g := g_\theta^{w, d}$, we have

\[ g(x)[c(x)] > \max_{j\neq c(x)}g(x)[j]+\gamma\]
for every $x\in\supp (\mu)$. Furthermore, assume $\|g(x)\|_\infty \leq B/2$ for all $x \in \supp (\mu)$.  
\end{assumption}

As an example, it suffices for the Bayes probability to have margin for (almost) all $x$, and be realizable by $\theta \in \mathcal C_M(w, d)$. In binary classification, such a requirement is referred to as a \emph{Massart condition}. 

We now state the analog of Theorem \ref{body thm:noisy informal} for the classification setting. First, for simplicity we state an informal version that can be compared to the informal statement of Theorem \ref{body thm:noisy informal}. 

\begin{theorem}[Noisy classification setting, informal]
\label{body thm:classification noisy}
Let $\delta \in (0, 1)$ and $n\geq2$. Assume data is gathered according to Assumption \ref{ass:adversarial noise, classification}'s adversarial corruption model. Also suppose $\mu_X$ satisfies marginal boundedness (Assumption \ref{body ass:lambda marginal boundedness}) with parameter $\Lambda$.

 Assume that Assumption \ref{ass:realizability with margin} holds for $\theta \in \mathcal C_M(w, d)$ with margin $\gamma > 0$ and bound $B$. Let $M\geq \mathrm{poly}(n,\Lambda,B,X,\delta^{-1})$, where $\mathrm{poly}$ is a fixed polynomial stated in the formal version. 
 
 Let $D\geq d$ and $W>w + w_+$, where $w_+ \leq C\left(1 + \frac{m}{D}\right)$ for some constant $C$ depending on $s$. 

Then, for $\hat{\theta} \sim U^{W, D, M}_n$ ,  we have
\[
\Pr_{x^n,\,\hat\theta}
\left[
\Delta^{\mathrm{cls}}(\hat\theta)\geq C_0\frac{W(wD+m)\log q_{\mathrm{eff},n}}{n}
\right]
\leq\delta.
\]
Here, $q_{\mathrm{eff},n} = \tilde O(d)$.
\end{theorem}

\begin{lemma}[Classification numerator bound]
\label{lemma:classification continuous num bound}
Let $t\in[0,1]$. Let
\[
B_{t,n}
:=
S_n^{W, D, M}
\cap
\{\phi\in\mathcal C_M(W,D):\Delta^{\mathrm{cls}}(\phi)\geq t\}.
\]
Under the noisy classification setting with at most $m$ corrupted labels,
\[
\E_{x^n}[\lambda(B_{t,n})]
\leq
(2M)^{N_{W,D}}
\binom{n}{m}
e^{-t(n-m)}.
\]
\end{lemma}
\begin{proof}
The proof is identical to Lemma \ref{lemma:adaptive continuous num_bound},
replacing $\tau$-accurate prediction by correct classification.
\end{proof}

For the formal version of the theorem, we write the rate directly in terms of an arbitrary stability certificate for $\theta$, which we know to be at worst exponential-in-$d$, but which may be better.   

\begin{theorem}[Theorem \ref{body thm:classification noisy}, formal]
\label{apdx thm:classification noisy}
Let $\gamma > 0$, $\delta \in (0, 1)$. Let $n\geq2$ be the number of data points and $m$ be the
maximum number of corrupted data points. Assume the noisy classification
setting of Assumption \ref{ass:adversarial noise, classification}. Also
suppose Assumption \ref{body ass:lambda marginal boundedness} holds with
parameter $\Lambda$. Finally assume that Assumption \ref{ass:realizability with margin} holds with margin $\gamma$ and bound $B$, and further assume that the witness $\theta$ has disturbance stability with parameters $(H_0, l_0, c_0)$. 

Let $M\geq\max\{1,X+2\ell_n,B/\ell_n\}$, with
$\ell_n=\frac{\delta}{6\Lambda n(n-1)}$.  Let $D\geq d$ and
$W>w+2+2s+4\left\lceil\frac{m}{D}\right\rceil$.

Let $\hat \theta \sim U^{W, D, M}_n$.  Then there is a constant $C_0$ depending only on $p$ and $s$, so that
\[
\Pr_{x^n, \hat{\theta}}
\left[
\Delta^{\mathrm{cls}}(\hat\theta)\geq t_n
\right]
\leq\delta,\qquad
t_n
:=
C_0\frac{(wWD+Wm)\log q_{\mathrm{eff},n}
+\log\frac{1}{\delta}}{n},
\]
Here, $q_{\mathrm{eff},n}$ is a value satisfying
\[
\log q_{\mathrm{eff},n}
\leq
\log\left( \mathrm{poly}(H_{0}, l_{0}, c_{0}, p, s, w, d, D, M, B, \Lambda,  X, n, \delta^{-1})\max\{1,\gamma^{-1}\}
\right).
\]
\end{theorem}
\begin{proof}

Let $I\subseteq[n]$ denote the set of corrupted points, with $|I|\leq m$.
For each class $a\in[s]$, define $v^a\in\R^s$ by
\[
v^a[a]=\frac B2,
\qquad
v^a[j]=-\frac B2
\quad(j\neq a).
\]
For each corrupted point, set
\[
\delta_i^{\mathrm{cls}}:=v^{\bar y_i}-g_\theta^{w, d}(x_i).
\]
Then $\|\delta_i^{\mathrm{cls}}\|_\infty\leq B$, so the memorization gadget
applies with $\Gamma=B$. Let
\[
w':=w+2+2s+4\left\lceil\frac mD\right\rceil.
\]
Let $\mathcal G$ be the event that the
$(n,m,\Lambda,B,\delta/2)$-memorization gadget succeeds. Then
$\Pr(\mathcal G)\geq1-\delta/2$, and on $\mathcal G$ padding the witness and
appending the gadget gives $\theta^\dagger\in\mathcal C_M(w',D)$ such that
\[
g^{w', D}_{\theta^\dagger}(x_i)=v^{\bar y_i}
\quad(i\in I),
\qquad
g^{w', D}_{\theta^\dagger}(x_i)=g^{w, d}_\theta(x_i)
\quad(i\notin I).
\]
Thus $\theta^\dagger$ classifies every observed label with margin at least
$B$ on the corrupted points and at least $\gamma$ on the uncorrupted points.

Give the choice of $\ell_n$ from the Theorem statement, Lemma \ref{lemma:disturbance stability padding H l c} together with Lemma \ref{lemma:structured stability} tells us that $\theta^\dagger$ has structured stability with \begin{equation}\label{eq:classification H L C bound}H, C, L \lesssim  \mathrm{poly}(H_{0}, l_{0}, c_{0}, p, s, w , d, D, M, B, \Lambda,  X, n, \delta^{-1}).\end{equation}
Define
\begin{equation}\label{eq:rho defs class}
\rho_{\mathrm{cls}}
:=
\min\left\{
1,
\frac1{C},
\frac{\gamma}{3L}
\right\},
\qquad
\rho'_{\mathrm{cls}}
:=
\frac{M}{J_{w',H}+1},
\end{equation}
and
\begin{equation}\label{eq:tilde q eff n def}
\widetilde q_{\mathrm{eff}, n}
:=
\max\left\{
\frac{2M}{\rho_{\mathrm{cls}}},
\frac{M}{\rho'_{\mathrm{cls}}}
\right\}.
\end{equation}
By Lemma \ref{lem:lift behaves like structured parameter}, every
$(\rho_{\mathrm{cls}},\rho'_{\mathrm{cls}})$-lift lies within $\gamma/3$ of
$\theta^\dagger$. Since $B\geq\gamma$, every such lift classifies the sample
correctly. Therefore, on $\mathcal G$,
\[
\lambda(S^{W, D, M}_n)
\geq
(2M)^{F_{w',W,D}}
(2\rho'_{\mathrm{cls}})^{K_{w',W,D}}
\rho_{\mathrm{cls}}^{N_{w',D}}.
\]
Splitting off $\mathcal G^c$ first, as in Theorem \ref{thm:stable witness bound}, and combining this positive denominator bound on $\mathcal G$ with Lemma \ref{lemma:classification continuous num bound} gives
\begin{equation}\label{eq:classification detailed bound}
\Pr[\Delta^{\mathrm{cls}}(\hat\theta)\geq t]
\leq
\frac\delta2
+
\binom{n}{m}
\left(\widetilde q_{\mathrm{eff}, n}\right)^{
N_{w',D}+K_{w',W,D}}
e^{-t(n-m)}.
\end{equation}
Set
\[
q_{\mathrm{eff}, n}
:=
\max\left\{
en,\widetilde q_{\mathrm{eff}, n}
\right\}.
\]

$q_{\mathrm{eff}, n}$ satisfies the bound from the Theorem statement, as we may see by combining Equations \eqref{eq:classification H L C bound}, \eqref{eq:rho defs class} and \eqref{eq:tilde q eff n def}. By the same calculations as in Corollary \ref{cor:adaptive structured parameter rate}, Equation \eqref{eq:classification detailed bound} can be manipulated into the given bound (using $w'WD\lesssim (wWD+Wm)$).  
\end{proof}

To recover the informal version of the theorem statement, we can simply use the concrete $(Mw)^d$-type stability bound on the witness $\theta \in \mathcal C_M(w, d)$.

\subsection{Boltzmann distribution}

A more realistic model for the distribution of learned parameter vectors $\hat{\theta} \in \mathcal C(W, D)$ for fixed training data $x_1 \dots x_n$ is a loss-based Boltzmann distribution with temperature $\beta > 0$ \citep{mingard2021sgd}: 
\begin{equation}\label{eq:boltzmann def}b_n^{W, D, M}(\theta) \propto e^{-\beta \hat{L}_n(\theta)}\mathbf{1}\{\theta \in \mathcal C_M(W, D)\},\end{equation}

where $\hat{L}_n = \frac{1}{n}\sum_{i = 1}^n \ell (x_i, y_i,  \theta)$ is the empirical loss corresponding to a \emph{loss-function} $\ell$. 

Here, we focus on classification, where a common choice of $\ell$ is \emph{cross-entropy} (CE)-loss. To define the loss function, we first consider the logits induced by any $\theta \in \mathcal C(W, D)$:

\[\mathrm{logit}_\theta(x)[i] = \frac{\exp (g_\theta^{W, D}(x)[i])}{\sum_{j = 1}^s \exp (g_\theta^{W, D}(x)[j])}.\]

Then, when $\bar y_i \in [s]$ indicates membership in one of $s$ classes,  $\ell_{CE}$ is defined as 

\[\ell_{CE}(x_i, \bar y_i, \theta) := -\log \left(\mathrm{logit_\theta}(x_i)[\bar y_i]\right) = -g_\theta^{W, D}(x_i)[\bar y_i]  + \log \sum_{j = 1}^s \exp (g_\theta^{W, D}(x_i)[j]).\]

Now take 

\[\hat{L}_n(\theta) = \frac{1}{n}\sum_{i = 1}^n \ell_{CE}(x_i, y_i, \theta),\]

and consider the definition of $b^{W, D, M}$ in \eqref{eq:boltzmann def}. We may \emph{condition} $b^{W, D, M}$ on interpolation. Specifically, we take 

\begin{equation}\label{eq:conditioned boltzmann def}\pi^{W, D, M}_n(\theta) \propto b_n^{W, D, M}(\theta) \mathbf{1}\{\theta \in S^{W, D, M}_n\} \propto e^{-\beta\hat{L}_n(\theta)}\mathbf{1}\{\theta \in S^{W, D, M}_n\}.\end{equation}
This density is used when $S_n^{W,D,M}$ has positive ambient volume; otherwise, return the zero parameter vector, as for $U_n^{W,D,M}$.

The next result shows that $\hat{\theta}\sim \pi^{W, D, M}_n(\theta)$ generalizes under the same assumptions as Theorem \ref{apdx thm:classification noisy}. 

\begin{theorem}\label{apdx thm:class boltzmann}
Adopt the setting of Theorem \ref{apdx thm:classification noisy}, but suppose $\hat\theta \sim \pi^{W, D, M}_n$ rather than $\hat{\theta}\sim U^{W, D, M}_n$. Then there is a constant $C_0$ depending on $p$ and $s$ so that
\[
\Pr_{x^n, \hat{\theta}}
\left[
\Delta^{\mathrm{cls}}(\hat\theta)\geq t_n
\right]
\leq s^\beta \delta,\qquad
t_n
:=
C_0\frac{(wWD+Wm)\log q_{\mathrm{eff},n}
+\log\frac{1}{\delta}}{n},
\]
Here, $q_{\mathrm{eff},n}$ is a value satisfying
\[
\log q_{\mathrm{eff},n}
\leq
\log\left( \mathrm{poly}(H_{0}, l_{0}, c_{0}, p, s, w, d, D, M, B, \Lambda,  X, n, \delta^{-1})\max\{1,\gamma^{-1}\}
\right).
\]
\end{theorem}
\begin{proof}
\begin{equation}\label{eq:boltzmann change of measure}\Pr_{
\hat\theta \sim \pi^{W, D, M}_n, x_n
}\left[\Delta^\mathrm{cls}(\hat{\theta}) \geq t\right] \leq \E_{x^n}\E_{\hat{\theta}\sim {U}_n^{W, D, M}}\left[\underbrace{\frac{e^{-\beta \hat{L}_n(\hat\theta)}}{\E_{\theta \sim U_n^{W, D, M}}[e^{-\beta \hat{L}_n(\theta)}]}}_{r(\hat{\theta})}\mathbf1\{\Delta^{\mathrm{cls}}(\hat\theta)\geq t\}\right]\end{equation}

On zero-volume datasets the two laws coincide and $r=1$ at their common output. Otherwise, let $\theta \in S^{W, D, M}_n$ be an interpolating solution, in the sense that $c_\theta^{W,D}(x_i)=\bar y_i$ for all $i \in [n]$. Notice that this implies $\mathrm{logit}_\theta(x_i)[\bar y_i] \geq \frac{1}{s}$. Therefore $\ell_{CE}(x_i, y_i, {\theta}) \leq \log s$ for each $i \in [n]$, and hence $\hat{L}_n({\theta}) \leq \log s$. Moreover, we definitionally have $\hat{L}_n(\theta) \geq 0$ for any $\theta$. Hence
\[r(\hat{\theta}) \leq e^{\beta \log s} = s^\beta.\]

Since the indicator in Equation \eqref{eq:boltzmann change of measure} is positive, we have the upper bound 

\[\Pr_{\hat\theta \sim \pi^{W, D, M}_n, x_n
}\left[\Delta^\mathrm{cls}(\hat{\theta}) \geq t\right] \leq s^\beta \Pr_{x^n, \hat{\theta} \sim U^{W, D, M}_n}\left[\Delta^\mathrm{cls}(\hat{\theta}) \geq t\right].\]

But the quantity on the right hand side is now bounded by Theorem \ref{apdx thm:classification noisy}.
\end{proof}

\begin{remark}
Setting $\delta' = \delta/s^\beta$ and appealing to Theorem \ref{apdx thm:class boltzmann} says that, with probability $\delta$, we achieve risk $t_n = \tilde O\left(\frac{\beta(wWD + Wm)}{n}\right)$, since $1/\delta$ only appears in log terms.
\end{remark}

\paragraph{Why is the Boltzmann model important?}
To heuristically see why the Boltzmann distribution is a natural model for gradient-based optimizers, first ignore the interpolation constraint. In the interior of
\(\mathcal C_M(W,D)\), Langevin dynamics targeting the ambient law \(b_n^{W, D, M}\) takes the form
\[
\theta_{k+1}
=
\theta_k-\eta\beta\nabla_\theta L_n(\theta_k)+\sqrt{2\eta}\,\xi_k,
\qquad
\xi_k\sim N(0,I),
\]
which is a noised version of gradient descent. $\pi_n^{W, D, M}$ essentially considers conditioning on the set $S^{W, D, M}_n$, which (still heuristically) can be viewed as assuming the above process finds an interpolating solution (i.e. repeating the process, allowing it to mix each time). More rigorously, the connection between Boltzmann/Bayesian sampling and gradient-methods has a long history in the literature \citep{welling2011bayesian, mingard2021sgd, chen2022stationary}.

\section{Full examples}\label{apdx:examples}

\subsection{Examples for Section \ref{subsec:noiseless realizable}}

\paragraph{Example: H\"older functions}
Suppose $\supp(\mu_X) \subseteq [0, 1]^p$ and let $f: [0, 1]^p \rightarrow \R$ satisfy H\"older smoothness with index $\beta > 0$ and radius $K> 0$, meaning  it has bounded partial derivatives up to order $\lfloor \beta \rfloor$, and its $\lfloor \beta \rfloor$-order partial derivative are $\beta - \lfloor \beta\rfloor$ H\"older continuous \citep{schmidt2020nonparametric}. Then it is known from \cite{ohn2019smooth} that for some $L_0, N_0, M_0$ depending only on $p, \beta$ and $K$, there exists a ReLU network $\theta \in \mathcal C_{M_0}(w, d)$ with width $w = N_0 \tau^{-p/\beta}$, depth $d = L_0 \log \frac{1}{\tau}$ such that$|g_{\theta}^{w, d} - f(x)| \leq \tau$ for $x \in [0, 1]^p$. Thus, we have the following high probability guarantee whenever $W > w$ and $D \geq d$:

\[\Delta_{\tau}(\hat\theta)\leq\tilde  O_{p,\beta,K}\left(\frac{WD\tau^{-p/\beta}}{n}\right).\]

If we choose $W \asymp w$ and $D \asymp d$, we get $\Delta_{\tau}(\hat\theta)\leq\tilde  O_{p,\beta,K}\left(\frac{\tau^{-2p/\beta}}{n}\right)$. If $f$ is bounded by $F$, following Remark \ref{body remark:L1 conversion} and tuning $\tau$ leads to the high probability $L_1$ bound:

\[\left\|\hat{f} - f\right\|_{L^1(\mu_X)} = \tilde O\left(n^{-\frac{\beta}{\beta + 2p}}\right)\]

For reference, the minimax rate is $O(n^{-\beta/p})$, usually achieved by specialized algorithms that are aware of $\beta$ (unlike our agnostic sampling rule), and requiring additional regularity assumptions on $\mu_X$ not made by us \citep{kohler2013optimal}. Without these regularity assumptions, there is an $\Omega\left(\frac{1}{n}\right)$ lower bound even for infinitely smooth $f$ \citep{kohler2013optimal}; the exponent in our rate approaches this barrier as $\beta$ grows with $p$ fixed.

\paragraph{Example: Barron functions}

Now assume $\supp(\mu_X) \subseteq B_p$, where $B_p$ is the Euclidean unit ball in $p$ dimensions, so that we may take $X=1$.  ReLU-Barron functions $f: B_p \rightarrow \R$ are functions realizable as ``infinite-width'' ReLU neural networks:
\begin{equation}\label{apdx eq:barron func def}f(x) = \E_{a, b, c \sim \rho}\left[a \sigma(b^Tx + c)\right].\end{equation}

The ReLU-Barron norm $\|f\|_\mathcal{B}$ is defined as $\inf_{\rho}\E\left[|a|(\|b\|_1 + |c|)\right]$, where the infimum is taken over measures $\rho$ that realize \eqref{apdx eq:barron func def}. Let $Q:=\|f\|_{\mathcal B}$. In Appendix \ref{apdx sec:Barron theory}, we use the results of  \cite{yangzhou2025optimal} to show that for every $r\geq1$ there is a function $f_r$ satisfying
\[
\|f-f_r\|_{L^\infty(B_p)}
\leq
C_p Q r^{-\frac{p+3}{2p}},
\]
and that $f_r$ can be represented exactly by a ReLU network $\theta_r \in \mathcal C_{2Q\vee1}(2p + 3, r)$ with $(H_{p,Q},L_{p,Q}r,C_{p,Q}r)$-structured stability (to be defined in Section \ref{sec:proof sketch}). Crucially, the stability is linear in $r$ rather than exponential. 

Thus, taking $d_\tau\asymp (Q/\tau)^{\frac{2p}{p+3}}$ gives a width-$(2p+3)$, depth-$d_\tau$ structured-stable parameter whose realization lies within $\tau/2$ of $f$ uniformly on $B_p$. Taking $D\asymp d_\tau$, Theorem \ref{thm:disturbance stable rate} --- a strengthening of Theorem \ref{body thm:continuous noiseless thm} ---  gives the following high probability guarantee whenever $W>2p+3$:
\[
\Delta_\tau(\hat\theta)
\leq
\tilde O\left(
\frac{W}{n}\left(\frac{Q}{\tau}\right)^{\frac{2p}{p+3}}
\right),
\]
where $\tilde O(\cdot)$ hides polynomial dependence on $p$. Since $\|f(x)\|_\infty\leq Q$, under the procedure of Remark \ref{body remark:L1 conversion}, we obtain that $\tau \asymp Q (\frac{W}{n})^{\frac{p + 3}{4p + 6}}$ yields
\[
\left\|\hat f - f\right\|_{L_2(\mu_X)}^2=
\tilde O\left(
Q^2\left(\frac{W}{n}\right)^{\frac{p+3}{2p + 3}}
\right).
\]
For fixed $p,Q$ and $W>2p+3$, this rate matches, up to logarithmic factors, the squared-$L_2$ minimax lower bound for Gaussian-noise regression \citep[Remark 4.3]{yangzhou2025optimal} and the best known upper-bound exponent for noiseless i.i.d. sampling \citep[Equation (39), $k=1$]{parhi2025upper}. 

In particular, the $L_2$-norm exponent tends to $1/4$ as $p\to\infty$, showing that the curse of dimensionality does not apply for Barron classes. 

\subsection{Half space example for Section \ref{subsec:unrealizable}}
\paragraph{Example: Learning half spaces} Consider the space of functions indicating $p$-dimensional half spaces:
\[
\mathrm{Halfspaces}(R)
=
\left\{
f_{a,b}(x)=\mathbf{1}\{a^Tx-b>0\}
:
a\in\mathbb S^{p-1},\ |b|\leq R
\right\}.
\]
Assume that our data is distributed i.i.d. according to $\mu_X$, and that any linear marginal of $\mu_X$ has bounded measure: i.e. there exists $\Lambda$ so that for any $a\in\mathbb S^{p-1}$, it holds that $\Pr_{x\sim\mu_X}\left[a^Tx\in I\right]\leq\Lambda |I|\footnotemark$. This assumption is satisfied by e.g. any uniform distribution on a bounded positive volume set, and vectors with i.i.d. entries from bounded-density measures

Take any $f\in\mathrm{Halfspaces}(R)$ with $f(x)=\mathbf{1}\{a^Tx-b>0\}$.
Because $f$ is discontinuous and ReLU networks are continuous, the $\tau$-realizability condition of Theorem \ref{body thm:continuous noiseless thm} is unsatisfiable for any $\tau<1$. However, we may define an approximant $f_h(x)
:=
\frac{1}{2h}
\left(
\sigma(a^Tx-b+h)-\sigma(a^Tx-b-h)
\right)$
that agrees with $f$ wherever $|a^Tx-b|\geq h$. Notice that $f_h$ is realized by some
$\theta_h\in\mathcal C_{A_h}(2,1)$, where $A_h
=
\max\left\{
1,\frac{1}{2h},R+h
\right\}$.
We also have $\sup_x |f(x)-f_h(x)|\leq\frac12$ so that the bounded difference hypothesis is satisfied with $\Gamma=\frac12$. The measure of the disagreement set between $f_h$ and $f$ is given by $\Pr_{x\sim\mu_X}
\left[
a^Tx\in(b-h,b+h)
\right]
\leq
2\Lambda h$. Taking $h_n=\frac{1}{2\Lambda n}$ allows us to apply Theorem~\ref{body thm:unrealizable thm informal} with $\epsilon_n=\frac1n$, $w_n=2$, $d_n=1$, $A_n=A_{h_n}$, and $D_n=D\geq1$.

Set $m_\delta:=\lceil2+\log(4/\delta)\rceil$ and require
\[W>6+4\left\lceil\frac{m_\delta}{D}\right\rceil.\]

We choose $M_n$ polynomial in $n$ and the problem parameters and large enough to satisfy the lower bound in the theorem statement. Then, for $\hat{\theta}\sim U^{W,D,M_n}_{\tau,n}$,
we have
\[
\Pr_{x^n,\hat{\theta}}
\left[
\Delta_\tau(\hat{\theta})
\geq
\tilde O\left(
\left(\frac{WD}{n}\right)
\log q_{\mathrm{eff},n}
\right)
\right]\leq\delta,
\]
where $q_{\mathrm{eff},n}$ is polynomial in $n,D,\Lambda,\delta^{-1},R,\tau^{-1}$ and the fixed problem parameters. Since $f$ takes values in $[0,1]$, the convention of Remark \ref{body remark:L1 conversion} gives $\|\hat f_n-f\|_{L_1(\mu_X)}\leq\tau+\Delta_\tau(\hat\theta)$. Thus, if $\tau\asymp1/n$, with probability at least $1-\delta$,
\[\|\hat f_n-f\|_{L_1(\mu_X)}=\tilde O\left(\frac{WD}{n}\right).\]

As an extension, results can be derived for learning intersections of halfspaces using additional witness depth (i.e. polytopes), which can be used to approximate general convex bodies with continuous boundaries.





\end{document}